%% file: main.tex
\documentclass[a4paper,12pt]{report}

\usepackage[left=2.5cm,right=2.5cm,top=2.5cm,bottom=2.5cm]{geometry}

\usepackage[afrikaans,english]{babel}
\usepackage{microtype}
\usepackage{setspace}
\usepackage{lmodern}
\usepackage{siunitx}

\usepackage{url}
\usepackage{xcolor}
\usepackage{combelow}
\usepackage{pifont}
\newcommand{\cmark}{\ding{51}}%
\newcommand{\chrupala}{Chrupa\l a}

\usepackage{tikz}
\usepackage{mathtools}
\usepackage{fontawesome}
\usetikzlibrary{backgrounds}
\usetikzlibrary{positioning}
\usetikzlibrary{fit}

\usepackage{amsmath}
\usepackage{amssymb}
\usepackage{namedtensor}
\usepackage{stmaryrd}
\DeclareMathOperator*{\mean}{mean}

\DeclareMathOperator*{\argmax}{argmax}
\ndef{\temp}{T}
\ndef{\emb}{K}
\ndef{\vocab}{W}

\newcommand{\ab}{\mathbf{a}}  
\newcommand{\ib}{\mathbf{i}}  
\newcommand{\tb}{\mathbf{t}}  
\newcommand{\hb}{\mathbf{h}}  
\newcommand{\yb}{\mathbf{y}}  
\newcommand{\Hb}{\mathbf{H}}  
\newcommand{\yv}{y^\mathrm{vis}} 
\newcommand{\yt}{y^\mathrm{bow}} 
\newcommand{\ybv}{\mathbf{y}^{\mathrm{vis}}}
\newcommand{\dnna}{\mathrm{DNN}_a} 
\newcommand{\dnni}{\mathrm{DNN}_i} 
\newcommand{\enc}{\mathrm{Enc}} 
\newcommand{\pool}{\mathrm{Pool}} %
\newcommand{\clf}{\mathrm{Clf}}

\newcommand{\yoruba}{Yor\`ub\'a\xspace}
\newcommand{\oneata}{{Onea\cb{t}\u{a}}\xspace}
\newcommand{\saraclar}{{Sara\c{c}lar}\xspace}
\usepackage{xspace}
\newcommand{\etal}{{\it et al.}\xspace}
\usepackage[raggedright,sf,bf]{titlesec}
\usepackage[margin=\the\parindent,small,bf,sf]{caption}
\titlelabel{\thetitle.\ }
\titleformat{\chapter}[display]{\huge\bfseries\sffamily}{\chaptertitlename\ \thechapter}{15pt}{\Huge \raggedright}
\titlespacing*{\chapter}{0pt}{0pt}{40pt}  
\makeatletter
\let\originall@chapter\l@chapter
\def\l@chapter#1#2{\originall@chapter{{\sffamily #1}}{#2}}
\makeatother

\definecolor{hermancolor}{HTML}{FF6600}
\definecolor{dancolor}{HTML}{9A00FF}
\definecolor{kayodecolor}{HTML}{00BFFF}

\newcommand{\kayode}[1]{\textcolor{kayodecolor}{#1}}

\usepackage{booktabs}
\usepackage{tabularx}
\usepackage{multirow}
\usepackage{siunitx}
\newcolumntype{C}{>{\centering\arraybackslash}X}
\newcolumntype{L}{>{\raggedright\arraybackslash}X}
\newcolumntype{R}{>{\raggedleft\arraybackslash}X}

\newcommand{\ii}[1]{{\footnotesize \textcolor{gray}{#1}}}

\newcommand{\config}[1]{\texttt{#1}}
\newcommand{\mylabel}[1]{{\color{gray} \footnotesize \sf #1}}
\let \savenumberline \numberline
\def \numberline#1{\savenumberline{#1.}}

\usepackage{graphicx}
\usepackage{pdfpages}
\usepackage{subcaption}
\usepackage{booktabs}
\usepackage{tabularx}
\usepackage{multirow}
 \usepackage{xspace}

\newcolumntype{C}{>{\centering\arraybackslash}X}
\newcolumntype{L}{>{\raggedright\arraybackslash}X}

\usepackage{fancyhdr}
\fancypagestyle{plain}{\fancyhead{}
                       
                       \fancyfoot[C]{\thepage}}
                       
\usepackage[prependcaption,textsize=scriptsize]{todonotes}
\reversemarginpar

\usepackage{algorithm}  

\usepackage{hyperref}
\hypersetup{colorlinks=true,linktoc=all,citecolor=black,linkcolor=black}
\usepackage[nottoc]{tocbibind}

\usepackage{algpseudocode}  
 
\usepackage{cite}  
\usepackage{etoolbox}
\makeatletter
\patchcmd{\ttlh@hang}{\parindent\z@}{\parindent\z@\leavevmode}{}{}
\patchcmd{\ttlh@hang}{\noindent}{}{}{}
\makeatother

\usepackage{tcolorbox}

\begin{document}

\input{phd_title_page}
\pagenumbering{roman}
\input{acknowledgements}
\input{declaration}
\input{abstract}
\tableofcontents
\listoffigures
\listoftables
\newpage
\pagenumbering{arabic}

\input{introduction}
\input{related_work}
\input{vgs}
\input{localisation_methods}
\input{yfacc}
\input{conclusion}
\bibliography{mybib}
\end{document}

%% file: phd_title_page.tex
\graphicspath{{frontmatter/fig/}}

\begin{titlepage}
	\begin{center}
		
		~\vspace{4.5em}
		
		{\sffamily \bfseries \huge Visually Grounded Keyword Detection and Localisation for Low-Resource Languages\par}
		
		\vspace{7em}
		
		{\large {\Large by \\ Kayode Kolawole Olaleye}}
		
		\vspace{8em}
		
		{\large Dissertation presented for the degree of Doctor of Philosophy (Electronic Engineering) in the Faculty of Engineering at Stellenbosch University \par}
		
		\vfill
		
		{\large {Supervisor}: Prof. Herman Kamper}
		
		\vspace{10em}
		
		{\Large March 2023}
	\end{center}
\end{titlepage}

%% file: acknowledgements.tex
\chapter*{Acknowledgements}
\makeatletter\@mkboth{}{Acknowledgements}\makeatother

I would like to express my deepest gratitude to my supervisor, Assoc. Prof. Herman Kamper, for his unparalleled support and guidance throughout my PhD journey. The funding he helped secure allowed me to focus solely on my studies, and his mentorship and calming presence during difficult times were invaluable. 

I am deeply grateful to Dr. Dan \oneata for his invaluable role as my unofficial co-supervisor and for his invaluable contributions to my research. His guidance and support have been instrumental in the success of my dissertation and I am truly grateful for the time and effort he has invested in my work.

I would like to thank DAAD, Google, and MIH Media Lab, for funding my studies, and many thanks to Prof. Bruce Basset, Benji Meltzer at Aerobotics, Prof. Olabode Koriko, Prof. Babatope Omolofe, and my supervisor Herman Kamper for their support in my funding applications. I also want to thank Prof. Marc Deisenroth for the opportunity to participate in the machine learning summer school in London at the start of my PhD journey, which was an incredible learning experience.

I am also grateful to all the past and present members of the LSL research group, Benjamin van Niekerk, Leanne Nortje, Christiaan Jacobs, Lisa van Staden, Matthew Baas, Werner van der Merwe, and Kevin Eloff, for their friendship, support, and invaluable contributions at different stages of my research. A special thanks to Benjamin for his helpful input for some of the experiments, and for the (therapeutic) chess games that helped me manage burnout; Leanne, for her words of encouragement and delicious cakes; Christiaan, for his unfailing friendship and words of encouragement; and Matthew for helping with that hard-drive issue. 

I would also like to express my gratitude to my examiners, Prof. Willie Brink,  Prof. Murat \saraclar, and Assoc. Prof. Richard Klein, as well as Prof. Herman Engelbrecht, the convener for my oral defence, for creating a relaxed and comfortable atmosphere during the defence. I would also like to acknowledge the entire members of the MIH Media Lab for providing a cosy atmosphere for conducting my research.

I am forever grateful to my parents, Pastor and Mrs. Olaleye, and my sisters Tolani Olaleye and Teniola Akinsanya for their unwavering support and encouragement. I also want to thank my wife Olubunmi Olaleye for her love and support.

Finally, I want to give thanks to Abba Father for being my source and sustenance throughout this journey.

%% file: declaration.tex
\graphicspath{{figs/}}
\newpage
\thispagestyle{plain}
\addcontentsline{toc}{chapter}{Declaration}
\makeatletter\@mkboth{}{Declaration}\makeatother

\centerline{\includegraphics[width=12cm]{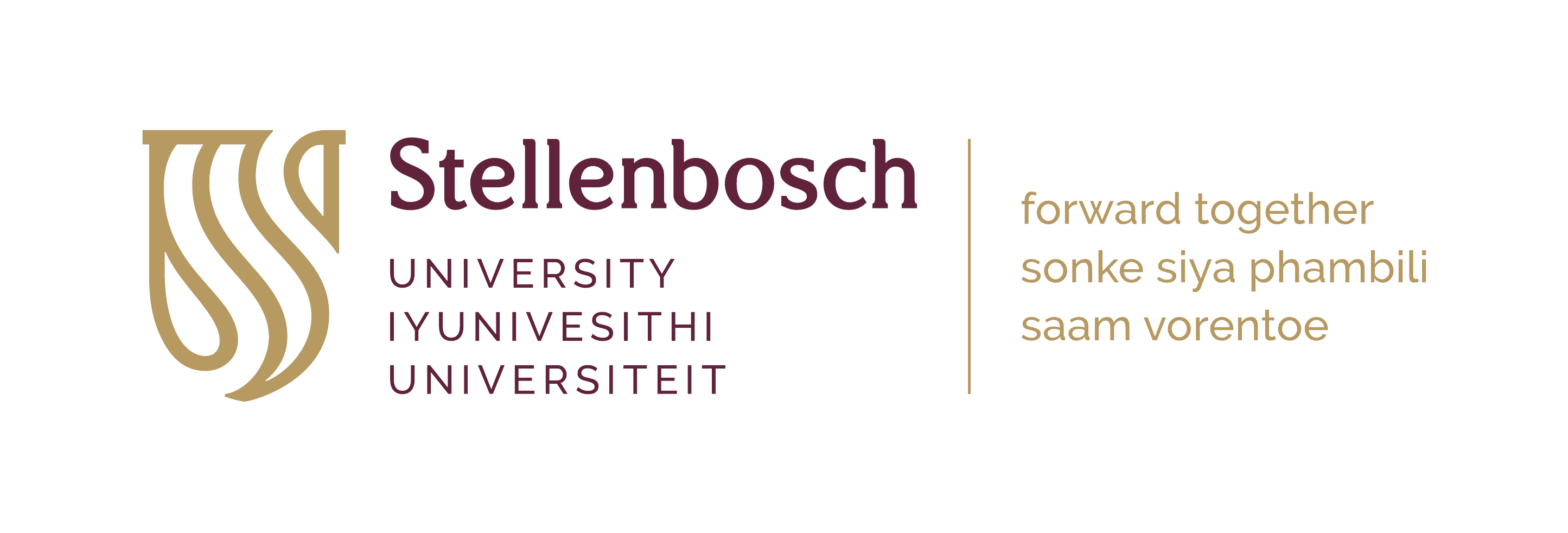}}
\vspace*{-10pt}

\section*{\centering Plagiaatverklaring / \textit{Plagiarism Declaration}}

\vspace*{1pt}

\begin{enumerate}
    \setlength\itemsep{0em}
    \item Plagiaat is die oorneem en gebruik van die idees, materiaal en ander intellektuele eiendom van ander persone asof dit jou eie werk is.\\
    \textit{Plagiarism is the use of ideas, material and other intellectual property of another's work
        and to present is as my own.}
    
    \item Ek erken dat die pleeg van plagiaat 'n strafbare oortreding is aangesien dit 'n vorm van diefstal is.\\
    \textit{I agree that plagiarism is a punishable offence because it constitutes theft.}
    
    \item Ek verstaan ook dat direkte vertalings plagiaat is. \\
    \textit{I also understand that direct translations are plagiarism.}
    
    \item Dienooreenkomstig is alle aanhalings en bydraes vanuit enige bron (ingesluit die internet) volledig verwys (erken). Ek erken dat die woordelikse aanhaal van teks sonder aanhalingstekens (selfs al word die bron volledig erken) plagiaat is. \\
    \textit{Accordingly all quotations and contributions from any source whatsoever (including the internet) have been cited fully. I understand that the reproduction of text without quotation marks (even when the source is cited) is plagiarism}
    
    \item Ek verklaar dat die werk in hierdie skryfstuk vervat, behalwe waar anders aangedui, my eie oorspronklike werk is en dat ek dit nie vantevore in die geheel of gedeeltelik ingehandig het vir bepunting in hierdie module/werkstuk of 'n ander module/werkstuk~nie. \\
    \textit{I declare that the work contained in this assignment, except where otherwise stated, is my original work and that I have not previously (in its entirety or in part) submitted it for grading in this module/assignment or another module/assignment.}
\end{enumerate}

\vfill


\noindent\begin{tabularx}{1.0\linewidth}{|L|L|}
    \hline
     {Studentenommer / \textit{Student number}} & 
     {Handtekening / \textit{Signature}} \\
    {\textbf{21524319}} &  {\includegraphics[width=0.20\columnwidth]{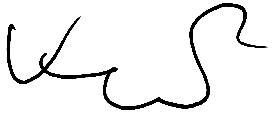}} \\
    \hline
     {Voorletters en van / \textit{Initials and surname}} & 
     {Datum / \textit{Date}} \\
    {\textbf{K.K. Olaleye}} & {\textbf{\today}} \\
    \hline
\end{tabularx}

\vspace{15pt}

%% file: abstract.tex
\chapter*{Abstract}
\addcontentsline{toc}{chapter}{Abstract}
\makeatletter\@mkboth{}{Abstract}\makeatother

\subsubsection*{English}
Visually grounded speech (VGS) models are trained on images paired with unlabelled spoken captions. Such models could be used to build speech systems in settings where it is impossible to get transcribed data, e.g.\ for documenting unwritten languages.
We investigate keyword localisation in speech---finding where in an utterance a given written keyword occurs---using VGS models trained in a real low-resource setting. Existing VGS studies fall short in two areas. Firstly, previous work has shown that VGS models can be used for tasks such as cross-modal retrieval, keyword detection and keyword spotting, but keyword localisation has not been explored. Secondly, most previous VGS studies use datasets where images are paired with speech in English (or another well-resourced language). English is therefore often used as a proxy for a low-resource language, making it difficult to accurately assess their performance in a real low-resource setting. 

Based on this, we address the following two overarching research questions: (i) Is keyword localisation possible with VGS models? (ii) In a real low-resource setting, can we do visually grounded keyword localisation cross-lingually?

To address the first question, we augment and extend existing VGS models with the ability to not only detect, but also localise written keywords. For this research question, we constrain ourselves to the artificial low-resource setting where English VGS data is used, allowing us to compare and directly extend previous work.
We use as starting point an existing methodology for training VGS models to detect keywords in speech: training images are tagged with soft textual labels using an existing offline image tagger, and these tags are then used as targets to train a speech network. I.e., the model receives a noisy target for whether words occur in an utterance, but not where or in which order.
We extend this model using four localisation methods. Input masking masks the input signal at different locations and measures the difference in the output unit for a particular keyword. Attention localisation requires an attention layer that pools features over the temporal axis; we use the attention weights as localisation scores. Grad-CAM is a saliency-based method that can be applied to any convolutional neural network to determine which parts of the network input most contribute to a particular output decision. The score aggregation method uses a particular type of pooling so that the output score can be regarded as an aggregation of
local scores; these can be used to select the most likely temporal location for a query
keyword.
In an oracle localisation test (where the model is told that a keyword is present in an utterance and then asked where it occurs), the masked-based localisation method achieves an accuracy of 57.0\%, outperforming all the other approaches, with the attention-based method coming second with 46.0\%.

To tackle the second research question (cross-lingual keyword localisation in a real low-resource setting), we start by collecting and releasing a new VGS dataset. The \yoruba Flickr Audio Caption Corpus (YFACC) dataset contains spoken captions for 6k Flickr images produced by a single speaker in \yoruba: a real low-resource language spoken in Nigeria. Using this data, we consider the problem of cross-lingual keyword detection and localisation: given an English text query, we detect whether the query occurs in \yoruba speech, and if it is detected, we localise where in the utterance the query occurs. To build this VGS system, images are automatically tagged with English visual labels serving as targets for an attention-based model that takes \yoruba speech as input. Then we apply the attention-based localisation method to do cross-lingual keyword detection and localisation for the first time in a real low-resource setting.  
The
cross-lingual model obtains a precision of $16.0$\% in actual keyword localisation which involves first detecting whether a keyword occurs before doing localisation. Although this result
is modest when viewed in isolation, this is a model trained without any parallel English-\yoruba data or any transcriptions.
We find that the performance can be improved by initialising the cross-lingual model from a model pretrained on the English image--speech dataset, giving a result of 22.8\%.

In answering the two main research questions, we make the following concrete contributions:
(1)~We propose a new VGS model for keyword detection and keyword spotting
using attention, and carry out a thorough comparison to existing VGS-based
methods. (2)~VGS models are extended with four localisation methods. (3)~We present a detailed quantitative and qualitative analysis revealing the
limits of the models above, showcasing their success and failure modes. We observe good localisation matches for some of the 67 keywords in the system's vocabulary (\textit{black}, \textit{pool}, \textit{soccer}, \textit{tree}),
while others are confused with semantically related words: \textit{ocean} $\to$ \textit{surfer}; \textit{ball} $\to$ \textit{soccer}; \textit{swimming} $\to$ \textit{pool}. 
(4)~We release a new multimodal, multilingual dataset which enables VGS
modelling in a real low-resource setting, resembling a language documentation scenario. The
dataset extends the Flickr8k image--text dataset to include \yoruba spoken captions.
(5)~We introduce a system for cross-lingual keyword detection and
keyword localisation in a real low-resource setting using our new \yoruba speech--image dataset. (6)~We provide a comprehensive analysis of the cross-lingual VGS model. We observe that there are keywords with good performance, such as \textit{brown} (\textit{b\'ur\'a\`un}; $100.0$\% precision), \textit{bike} (\textit{k\d{\`e}k\d{\'e}}; $94.1$\%) and \textit{grass} (\textit{kor\'iko} $90.9$\%). But there are many others on which the model struggles due to poor visual grounding and confusion between semantically related concepts.

In summary, we show that VGS models can be used for a limited form of keyword localisation in a real low-resource setting. We hope that our new dataset and new findings will stimulate more research in the use of VGS models for real low-resource languages.

\selectlanguage{yoruba}

\subsubsection*{\yoruba}

A m\'a \'n\d{s}e \`id\'anil\d{\'e}k\d{\`o}\d{\'o} f\'un \`aw\d{o}n \d{\`e}r\d{o} \`aw\`o\d{s}e \d{\`o}r\d{\`o} \`af\d{e}nus\d{o} t\'i a f’\`aw\`or\'an dar\'i n\'ipa l\'ilo \`aw\d{o}n \`aw\`or\'an t\'i a so p\d{\`o} p\d{\`e}l\'u \`aw\d{o}n \`ap\`ej\'uw\`e\'e \`af\d{e}nus\d{o} w\d{o}n t\'i k\`o n\'i k\'ik\d{o}. Ir\'u \`aw\d{o}n \d{\`e}r\d{o} b\d{\'e}\d{\`e} j\d{\'e} k\'i \'o \d{s}e\'e\d{s}e l\'ati k\d{\'o} \d{\`e}r\d{o} \d{\`o}r\d{\`o} \`af\d{e}nus\d{o} n\'i ibi t\'i k\`o s\'i \d{\`o}p\d{\`o}l\d{o}p\d{\`o} d\'at\`a (\d{o}r\d{o} \`af\d{e}nus\d{o} ati \`af\d{o}w\d{\'o}k\d{o} w\d{o}n), n\'i \`ap\d{e}r\d{e}, f\'un \`id\'ab\`ob\`o \`aw\d{o}n \`ed\`e t\'i k\`o n\'i k\'ik\d{o}. A \d{s}e \`iw\'ad\`i\'i b\'i a \d{s}e l\`e t\d{\'o}ka s\'i ibi t\'i \d{\`o}r\d{\`o} \`af\d{o}w\d{\'o}k\d{o} kan w\`a n\'in\'u \`aw\d{o}n \d{\`o}r\d{\`o} \`af\d{e}nus\d{o} n\'ipa l\'ilo \`aw\d{o}n \d{\`e}r\d{o} \`aw\`o\d{s}e \d{\`o}r\d{\`o} \`af\d{e}nus\d{o} t\'i a f’\`aw\`or\'an dar\'i t\'i a k\d{\'o} n\'ibi t\'i k\`o s\'i \d{\`o}p\d{\`o}l\d{o}p\d{\`o} ohun \`el\`o \d{\`e}k\d{\'o}. \`Al\'eb\`u m\'ej\`i w\`a l\'ara \`aw\d{o}n \`iw\'ad\`i\'i l\'ori \`aw\`o\d{s}e \`aw\d{o}n \d{\`o}r\d{\`o} \`af\d{e}nus\d{o} t\'i a f’\`aw\`or\'an dar\'i t\'i \'o w\`a. N\'i \`ak\d{\'o}k\d{\'o}, k\`o s\'i n\'in\'u \`aw\d{o}n \`iw\'ad\`i\'i lori \`aw\`o\d{s}e \`aw\d{o}n \d{\`o}r\d{\`o} \`af\d{e}nus\d{o} t\'i a f’\`aw\`or\'an dar\'i t\'i \'o l\'epa b\'i a \d{s}e l\`e l\`o w\d{\'o}n l\'ati t\d{\'o}ka s\'i ibi t\'i \d{\`o}r\d{\`o} \`af\d{o}w\d{\'o}k\d{o} kan w\`a n\'in\'u \`aw\d{o}n \d{\`o}r\d{\`o} \`af\d{e}nus\d{o}; b\'i \'o ti l\d{\`e} ti j\d{\'e} w\'ip\'e a ti l\`o w\d{\'o}n f\'un \`iw\'ad\`i\'i b\'oy\'a \d{\`o}r\d{\`o} \`af\d{o}w\d{\'o}k\d{o} kan w\`a n\'in\'u \`aw\d{o}n \d{\`o}r\d{\`o} \`af\d{e}nus\d{o}. N\'i \d{\`e}\d{\`e}kej\`i, \d{\`o}p\d{\`o}l\d{o}p\d{\`o} n\'in\'u \`aw\d{o}n \`iw\'ad\`i\'i l\'or\'i \`aw\`o\d{s}e \`aw\d{o}n \d{\`o}r\d{\`o} \`af\d{e}nus\d{o} t\'i a f’\`aw\`or\'an dar\'i lo \`aw\d{o}n \`aw\`or\'an t\'i a so p\d{\`o} p\d{\`e}l\'u \`aw\d{o}n \d{\`o}r\d{\`o} \`af\d{e}nus\d{o} n\'i \`ed\`e g\d{\`e}\d{\'e}s\`i (\`ab\'i \`aw\d{o}n \`ed\`e t\'i \'o n\'i d\'at\`a p\'up\d{\`o}). A m\'a \'ns\'ab\`a lo \`ed\`e g\d{\`e}\d{\'e}s\`i g\d{\'e}g\d{\'e} b\'i a\d{s}oj\'u \`ed\`e t\'i k\`o n\'i \d{\`o}p\d{\`o}l\d{o}p\d{\`o} d\'at\`a, n\'itor\'i \`ey\'i, \'o nira l\'ati l\`e w\d{o}n b\'i \d{\`e}r\d{o} \`aw\`o\d{s}e \d{\`o}r\d{\`o} \`af\d{e}nus\d{o} t\'i a f’\`aw\`or\'an dar\'i \d{s}e \d{s}e d\'aad\'aa s\'i n\'i ibi t\'i k\`o s\'i \d{\`o}p\d{\`o}l\d{o}p\d{\`o} d\'at\`a.

El\'ey\`i\'i l\'o m\'u wa koj\'u \`aw\d{o}n \`ib\'e\`er\`e m\'ej\`i w\d{\`o}ny\'i: (i) \d{S}\'e \'o \d{s}e\'e\d{s}e l\'ati lo \d{\`e}r\d{o} \`aw\`o\d{s}e \`aw\d{o}n \d{\`o}r\d{\`o} \`af\d{e}nus\d{o} t\'i a f’\`aw\`or\'an dar\'i f\'un \`it\d{\'o}kas\'i ibi t\'i \d{\`o}r\d{\`o} \`af\d{o}w\d{\'o}k\d{o} w\`a n\'in\'u \`aw\d{o}n \d{\`o}r\d{\`o} \`af\d{e}nus\d{o}? (ii) N\'i ibi t\'i k\`o s\'i \d{\`o}p\d{\`o}l\d{o}p\d{\`o} d\'at\`a, \d{s}\'e \'o \d{s}e\'e\d{s}\'ee l\'ati lo \d{\`e}r\d{o} \`aw\`o\d{s}e \`aw\d{o}n \d{\`o}r\d{\`o} \`af\d{e}nus\d{o} t\'i a f’\`aw\`or\'an dar\'i f\'un \`it\d{\'o}kas\'i \d{\`o}r\d{\`o} \`af\d{o}w\d{\'o}k\d{o} n\'ipa l\'ilo \`ed\`e m\'ej\`i pap\d{\`o}?

L\'ati koj\'u \`ib\'e\`er\`e \`ak\d{\'o}k\d{\'o}, a r\'o \d{\`e}r\d{o} \`aw\`o\d{s}e \d{\`o}r\d{\`o} \`af\d{e}nus\d{o} t\'i a f’\`aw\`or\'an dar\'i p\d{\`e}l\'u agb\'ara l\'ati l\`e \d{s}e \`it\d{\'o}kas\'i ibi t\'i \d{\`o}r\d{\`o} \`af\d{o}w\d{\'o}k\d{o} w\`a n\'in\'u \`aw\d{o}n \d{\`o}r\d{\`o} \`af\d{e}nus\d{o}. A \d{s}e \`iw\'ad\`i\'i wa n\'i ibi t\'i a ti lo \`ed\`e t\'i \'o n\'i \d{\`o}p\d{\`o}l\d{o}p\d{\`o} d\'at\`a g\d{\'e}g\d{\'e} b\'i a\d{s}oj\'u f\'un \`ed\`e t\'i k\`o n\'i \d{\`o}p\d{\`o}l\d{o}p\d{\`o} d\'at\`a. El\'ey\`i\'i gb\`a w\'a l\'ay\`e l\'ati \d{s}e \`afiw\'e t\`a\`ar\`a p\d{\`e}l\'u i\d{s}\'e t\'i \'o \d{s}\'a\'aj\'u i\d{s}\'e wa. L\'ati b\d{\`e}r\d{\`e} \`iw\'ad\`i\'i wa, a lo \`il\`an\`a f\'un k\'ik\d{\'o} \d{\`e}r\d{o} \`aw\`o\d{s}e \d{\`o}r\d{\`o} \`af\d{e}nus\d{o} t\'i a f’\`aw\`or\'an dar\'i t\'i \'o w\`a t\d{\'e}l\d{\`e}. \`Il\`an\`a y\`i\'i d\'a l\'or\'i k\'ik\d{\'o} \d{\`e}r\d{o} \`aw\`o\d{s}e \d{\`o}r\d{\`o} \`af\d{e}nus\d{o} t\'i a f’\`aw\`or\'an dar\'i p\d{\`e}l\'u \`aw\d{o}n \`aw\`or\'an \`ati t\'a\`ag\`i t\'i k\`o g\'unr\'eg\'e w\d{o}n, n\'ipa l\'ilo \d{\`e}r\d{o} t\'i a fi \'n t\'a\`ag\`i \`aw\`or\'an. T\'a\`ag\`i w\d{\`o}ny\'i ni w\d{\'o}n l\`o l\'ati fi k\d{\'o} \d{\`e}r\d{o} \d{\`o}r\d{\`o} \`af\d{e}nus\d{o}.  Ir\'u \d{\`e}r\d{o} \`aw\`o\d{s}e \`aw\d{o}n \d{\`o}r\d{\`o} \`af\d{e}nus\d{o} ti a f’\`aw\`or\'an dar\'i t\'i a \d{s}e b\'ay\`i\'i k\d{\'o} ni a r\'o p\d{\`e}l\'u agb\'ara n\'ipa l\'ilo \`il\`an\`a f\'un \`it\d{\'o}kas\'i \d{\`o}r\d{\`o} \`af\d{o}w\d{\'o}k\d{o} n\'in\'u \`aw\d{o}n \d{\`o}r\d{\`o} \`af\d{e}nus\d{o}. \`Il\`an\`a \`ak\d{\'o}k\d{\'o} (Input masking) d\'a l\'or\'i b\'ibo \`aw\d{o}n ib\`ikan n\'in\'u \`aw\d{o}n \d{\`o}r\d{\`o} \`af\d{e}nus\d{o} m\d{\'o}l\d{\`e} l\'ati w\d{o}n ipa t\'i \'o ma n\'i l\'or\'i \`id\'am\d{\`o} \d{\`o}r\d{\`o} \`af\d{o}w\d{\'o}k\d{o} kan. \`Il\`an\`a \d{\`e}\d{\`e}kej\`i (Attention) d\'a l\'or\'i \d{s}\'i\d{s}e \`ak\'iy\`es\'i ibi t\'i \'o \d{s}e p\`at\`ak\`i j\`u l\'ara \`aw\d{o}n \d{\`o}r\d{\`o} \`af\d{e}nus\d{o} l\'ati \d{s}e \`id\'am\d{\`o} \d{\`o}r\d{\`o} \`af\d{o}w\d{\'o}k\d{o} kan. \`Il\`an\`a \d{\`e}\d{e}k\d{e}ta (Grad-CAM) d\'a l\'or\'i \d{s}\'i\d{s}e \`iw\'ad\`i\'i ibo l\'ara \d{\`o}r\d{\`o} \`af\d{e}nus\d{o} ni \'o \d{s}e p\`at\`ak\`i j\`u f\'un \d{\`o}r\d{\`o} \`af\d{o}w\d{\'o}k\d{o} t\'i a f\d{\'e} d\'am\d{\`o}. \`Il\`an\`a \d{\`e}\d{\`e}k\d{e}rin (Score aggregation) d\'a l\'or\'i \`ik\'oj\d{o}p\d{\`o} \`aw\d{o}n \`es\`i l\'ati l\`e t\d{\'o}ka s\'i ibi t\'i \d{\`o}r\d{\`o} \`af\d{o}w\d{\'o}k\d{o} w\`a n\'in\'u \`aw\d{o}n \d{\`o}r\d{\`o} \`af\d{e}nus\d{o}. \`Il\`an\`a \`ak\d{\'o}k\d{\'o} gb\'egb\'a or\'ok\`e p\d{\`e}l\'u \`i\d{s}ed\'ed\'e 57\% l\'or\'i i\d{s}\'e t\'it\d{o}kas\'i ibi t\'i \d{\`o}r\d{\`o} \`af\d{o}w\d{\'o}k\d{o} w\`a l\'a\`i b\`ik\'it\`a f\'un \`id\'am\d{\`o} \d{\`o}r\d{\`o} b\d{\'e}\d{\`e}. \`Il\`an\`a \d{\`e}\d{\`e}kej\`i n\'i \`i\d{s}ed\'ed\'e 46.0\%.

L\'ati koj\'u \`ib\'e\`er\`e \d{\`e}\d{\`e}kej\`i (n\'i ibi t\'i k\`o s\'i \d{\`o}p\d{\`o}l\d{o}p\d{\`o} d\'at\`a, \d{s}\'e \'o \d{s}e\'e\d{s}\'ee l\'ati lo \d{\`e}r\d{o} \`aw\`o\d{s}e \d{\`o}r\d{\`o} \`af\d{e}nus\d{o} t\'i a f’\`aw\`or\'an dar\'i f\'un \`it\d{\'o}kas\'i \d{\`o}r\d{\`o} \`af\d{o}w\d{\'o}k\d{o} n\'ipa l\'ilo \`ed\`e m\'ej\`i pap\d{\`o}?), a b\d{\`e}r\d{\`e} l\'ati \d{s}a d\'at\`a \`aw\`o\d{s}e \`aw\d{o}n \d{\`o}r\d{\`o} \`af\d{e}nus\d{o} t\'i a f’\`aw\`or\'an dar\'i tuntun j\d{o}. D\'at\`a y\`i\'i, t\'i a p\`e n\'i \yoruba Flickr Audio Caption Corpus (YFACC) n\'i \d{\`o}r\d{\`o} \`af\d{e}nus\d{o} t\'i \'o \d{s}\`ap\`ej\'uw\`e\'e 6000 n\'in\'u \`aw\d{o}n \`aw\`or\'an Flickr t\'i en\`ikan\d{s}o\d{s}o k\'o j\d{o} p\d{o} n\'i \`ed\`e \yoruba. A lo YFACC l\'ati k\d{\'o} \d{\`e}r\d{o} \`aw\`o\d{s}e \d{\`o}r\d{\`o} \`af\d{e}nus\d{o} ti a f’\`aw\`or\'an dar\'i l\'ati \d{s}e i\d{s}\d{\'e}\d{e} d\'id\'am\d{\`o} \`ati t\'it\d{\'o}kas\'i \d{\`o}r\d{\`o} \`af\d{o}w\d{\'o}k\d{o} n\'i \`ed\`e g\d{\`e}\d{\'e}s\`i n\'in\'u \`aw\d{o}n \d{\`o}r\d{\`o} \`af\d{e}nus\d{o} n\'i \`ed\`e \yoruba. L\'ati k\d{\'o} \d{\`e}r\d{o} \`aw\`o\d{s}e \d{\`o}r\d{\`o} \`af\d{e}nus\d{o} t\'i a f’\`aw\`or\'an dar\'i ir\'ub\d{\`e}, a t\'a\`ag\`i \`aw\d{o}n \`aw\`or\'an, l\'a\`if\d{o}w\d{\'o}y\'i, p\d{\`e}l\'u t\'a\`ag\`i n\'i \`ed\`e g\d{\`e}\d{\'e}s\`i. \`Ey\'i ni a s\`i fi dar\'i k\'ik\d{\'o} \`il\`an\`a \d{\`e}\d{\`e}kej\`i (Attention) f\'un t\'it\d{\'o}kas\'i \d{\`o}r\d{\`o} \`af\d{o}w\d{\'o}k\d{o} ninu \`aw\d{o}n \d{\`o}r\d{\`o} \`af\d{e}nus\d{o} f\'un i\d{s}\d{\'e}\d{e} d\'id\'am\d{\`o} \`ati t\'it\d{\'o}kas\'i \d{\`o}r\d{\`o} \`af\d{o}w\d{\'o}k\d{o} g\d{\`e}\d{\'e}s\`i n\'in\'u \`aw\d{o}n \d{\`o}r\d{\`o} \`af\d{e}nus\d{o} \yoruba n\'i \`igb\`a a\`k\d{\'o}k\d{\'o} n\'i ibi t\'i k\`o s\'i \d{\`o}p\d{\`o}l\d{o}p\d{\`o} d\'at\`a. \`Il\`an\`a y\`i\'i ni \`i\d{s}ed\'ed\'e al\'a\`it\`as\'e 16.0\% l\'or\'i  i\d{s}\d{\'e}\d{e} t\'it\d{\'o}kas\'i \`af\d{o}w\d{\'o}k\d{o} n\'i \`ed\`e g\d{\`e}\d{\'e}s\`i n\'in\'u \d{\`o}r\d{\`o} \`af\d{e}nus\d{o} n\'i \`ed\`e \yoruba l\d{\'e}y\`in \d{s}\'i\d{s}e \`id\'am\d{\`o} \d{\`o}r\d{\`o} \`af\d{o}w\d{\'o}k\d{o} n\'a\`a. B\'i\'otil\d{\`e}j\d{e}pe \`il\`an\`a n\'a\`a \d{s}e d\'ed\'e n\'i \`iw\d{\`o}nba t\'i a b\'a w\`o\'o n\'i \`iy\`as\d{\'o}t\d{\'o}, \`Il\`an\`a n\'a\`a k\`o n\'i \`a\`ay\`e s\'i \`ik\d{\'e}k\d{\`o}\d{\'o} t\'o d\'an m\d{\'o}nr\'an. A ri n\'in\'u \`iw\'ad\`i\'i wa p\'e \`i\d{s}ed\'ed\'e n\'a\`a g\`unk\`e s\'i 22.8\% t\'i a b\'a fi \d{\`e}r\d{o} m\'ir\`an t\'i \'o k\d{\'e}\d{\`e}k\d{\'o} l\'ara \`aw\d{o}n \`aw\`or\'an \`ati \`ap\`ej\'uw\`e\'e \`af\d{e}nus\d{o} w\d{o}n n\'i \`ed\`e g\d{\`e}\d{\'e}s\`i \d{s}e \`ip\'il\d{e}\d{s}\d{\`e} r\d{\`e}.

N\'ipa k\'ikoj\'u \`ib\'e\`er\`e \`iw\'ad\`i\'i wa, a \d{s}e \`aw\d{o}n \`afik\'un s\'in\'u \`im\d{\`o} w\d{\`o}ny\'i: (1)  A d\'ab\`a\'a \`il\`an\`a tuntun f\'un b\'i a \d{s}e l\`e lo \d{\`e}r\d{o} \`aw\`o\d{s}e \d{\`o}r\d{\`o} \`af\d{e}nus\d{o} t\'i a f’\`aw\`or\'an dar\'i f\'un \`id\'am\d{\`o} \`ati \`it\d{\'o}kas\'i \d{\`o}r\d{\`o} \`af\d{o}w\d{\'o}k\d{o}; a s\`i \d{s}e \`ifiw\'era p\'ip\'e \`aw\`o\d{s}e n\'a\`a p\d{\`e}l\'u \`aw\d{o}n \`il\`an\`a t\'i \'o ti w\`a t\d{\'e}l\d{\`e}. (2) A r\'o \d{\`e}r\d{o} \`aw\`o\d{s}e \`aw\d{o}n \d{\`o}r\d{\`o} \`af\d{e}nus\d{o} t\'i a f’\`aw\`or\'an dar\'i l'agb\'ara p\d{\`e}l\'u \`il\`an\`a m\d{\'e}rin f\'un \`it\d{\'o}kas\'i \d{\`o}r\d{\`o} \`af\d{o}w\d{\'o}k\d{o}. (3) A \d{s}e \d{\`e}k\'unr\d{\'e}r\d{\'e} \`it\'upal\d{\`e} l\'ati \d{s}\`afih\`an \`aw\d{o}n ibi t\'i \`aw\d{o}n \d{\`e}r\d{o} \`aw\`o\d{s}e n\'a\`a ti \d{s}\`a\d{s}ey\d{o}r\'i \`ati ibi t\'i w\d{\'o}n ti k\`un\`a. A \d{s}\`ak\'iy\`es\'i p\'e l\'ara \`aw\d{o}n \d{\`o}r\d{\`o} 67 t\'i \'o w\`a n\'in\'u fokab\'ul\'ar\`i t\'i a l\`o (black, pool, soccer, tree) ni \`it\d{\'o}kas\'i w\d{o}n \d{s}ed\'ed\'e, n\'igb\`at\'i \`it\d{\'o}kas\'i \`aw\d{o}n \d{\`o}r\d{\`o} \`iy\'ok\`u k\`un\`a: \textit{ocean} $\to$ \textit{surfer}; \textit{ball} $\to$ \textit{soccer}; \textit{swimming} $\to$ \textit{pool}. (4) A \d{s}e \`at\'us\'il\`e d\'at\`a t\'i \'o w\'ul\`o f\'un \d{s}\'i\d{s}e \`id\'anil\d{\'e}k\d{\`o}\d{\'o} f\'un \`aw\d{o}n \d{\`e}r\d{o} \`aw\`o\d{s}e \`aw\d{o}n \d{\`o}r\d{\`o} \`af\d{e}nus\d{o} t\'i a f’\`aw\`or\'an dar\'i (5) A \d{s}e \`agb\'ekal\d{\`e} \d{\`e}r\d{o} f\'un d\'id\'am\d{\`o} \`ati t\'it\d{\'o}kas\'i \d{\`o}r\d{\`o} \`af\d{o}w\d{\'o}k\d{o} n\'i \`ed\`e g\d{\`e}\d{\'e}s\`i n\'in\'u \d{\`o}r\d{\`o} \`af\d{e}nus\d{o} n\'i \`ed\`e \yoruba n\'i ibi t\'i k\`o s\'i d\'at\`a p\'up\d{\`o} n\'ipa l\'ilo YFACC. (6) A \d{s}e on\'ir\'uur\'u \`it\'upal\d{\`e} \d{\`e}r\d{o} \`aw\`o\d{s}e \d{\`o}r\d{\`o} \`af\d{e}nus\d{o} t\'i a f’\`aw\`or\'an dar\'i el\'ed\`e m\'ej\`i. A \d{s}\`ak\'iy\`es\'i p\'e l\'ara \`aw\d{o}n \d{\`o}r\d{\`o} \`af\d{o}w\d{\'o}k\d{o} ni \`i\d{s}ed\'ed\'e t\'i \'o d\'ara, b\'i \`ap\d{e}r\d{e}, \textit{brown} (\textit{b\'ur\'a\`un}; \`i\d{s}ed\'ed\'e al\'a\`it\`as\'e 100\%), \textit{bike} (\textit{k\d{\`e}k\d{\'e}}; 94.1\%) \`ati \textit{grass} (\textit{kor\'iko} 90.9\%). \d{S}\`ugb\d{\'o}n \d{\`e}r\d{o} n\'a\`a k\`o \d{s}e d\'ad\'aa l\'or\'i \d{\`o}p\d{\`o}l\d{o}p\d{\`o}  \`aw\d{o}n \d{\`o}r\d{\`o} m\'ir\`an.

N\'i \`ak\'oj\d{o}p\d{\`o}, a fih\`an p\'e \`aw\d{o}n \d{\`e}r\d{o} \`aw\`o\d{s}e \d{\`o}r\d{\`o} \`af\d{e}nus\d{o} t\'i a f’\`aw\`or\'an dar\'i \d{s}e\'e l\`o f\'un \`it\d{\'o}kas\'i \d{\`o}r\d{\`o} \`af\d{o}w\d{\'o}k\d{o} n\'i ibi t\'i k\`o s\'i \d{\`o}p\d{\`o}l\d{o}p\d{\`o} d\'at\`a. A n\'ir\`et\'i p\'e d\'at\`a tuntun wa \`ati \`aw\d{o}n \`aw\'ar\'i tuntun t\'i a \d{s}e y\'o\`o \d{s}e \`iw\'ur\'i f\'un \`iw\'ad\`i\'i s'i\'i ni l\'ilo \`aw\d{o}n \d{\`e}r\d{o} \`aw\`o\d{s}e \d{\`o}r\d{\`o} \`af\d{e}nus\d{o} t\'i a f’\`aw\`or\'an dar\'i f\'un \`aw\d{o}n \`ed\`e t\'i k\`o n\'i \d{\`o}p\d{\`o}l\d{o}p\d{\`o} d\'at\`a.

\selectlanguage{english}

%% file: introduction.tex
\chapter{Introduction}
\label{chap:introduction}
	Automatic speech recognition (ASR)
	enables human-computer interaction and improves accessibility by transcribing audio media.
	However,
	 accurate ASR systems are available in only a fraction of the world's languages because such systems require a vast amount of transcribed speech data~\cite{besacier2014}. 
	As a result, there has been growing interest in speech processing systems that, instead of using exact transcriptions, can learn from weakly labelled data~\cite{duong2016, palaz2016, settle2017, weiss2017}.
	
	One paradigm is that of visually grounded speech (VGS) modelling. A VGS model is a specific form of multimodal system that learns from two modalities: images and their spoken captions~\cite{driesen2010, synnaeve2014b, harwath2015, harwath2016,  harwath2017, harwath2018a, harwath2018b, eloff2019, harwath2019a, harwath2019b}.
	Since paired speech and images are signals available during early language acquisition, such co-occurring audio-visual inputs are ideal for modelling infants' language learning~\cite{bomba1983, pinker1994, eimas+quinn94, roy2003, boves2007, chrupala2016, okko2019}.
	VGS models are also suitable for
	teaching new words to robots~\cite{meng2013, nortje2020}.
	The appeal for the VGS setup further stems from the fact that
	images and utterances describing them could arguably be easier to collect than speech and transcriptions,
	especially when developing systems for low-resource languages~\cite{de1998} or languages with no written form~\cite{scharenborg2018, lupke2010, bird2020, scharenborg2020, wang2021a}. 
	
	Previous work has shown that VGS models can be used for tasks such as cross-modal retrieval, keyword detection and keyword spotting. This research aims
	 to investigate keyword localisation in speech---finding where in an utterance a keyword occurs---using VGS models trained in a real low-resource linguistic environment. This task and problem setting have not been explored in previous studies.
	
In the remainder of this chapter, we
 provide an introduction to this dissertation by first presenting the motivation for the study (Section~\ref{sec:motivation}). We then present the work's main research questions (Section~\ref{sec:research_questions}), research methodologies (Section~\ref{sec:methodology}), and research contributions~ (Section~\ref{sec:contributions}). Finally, we give an overview of the structure of the dissertation (Section~\ref{sec:structure_overview}).

\section{Motivation}
\label{sec:motivation}
We start this section by presenting two scenarios to illustrate settings in which the findings of this dissertation could be useful. We then give background on VGS modelling, focussing on methodologies and datasets used in previous work. 

\subsection{Application scenarios}

To illustrate how VGS-based modelling could be used in real low-resource settings, consider the following two illustrative examples.
\paragraph{Language documentation.}
Imagine a linguist is tasked with documenting a low-resource language (most of the world's roughly 7k languages are considered low-resource). She collects a data bank of utterances from native speakers in the language of interest, and would like to learn how to say general concepts (such as ``child", ``water", ``red") in that new language. She considered using machine learning tools to
aid with this process. However, developing automated models requires labelled data, for example audio and transcriptions. These resources are scarce or even completely unavailable in some settings, as is the case with languages that do not have a written form. In this situation, it is reasonable to assume that she can collect images paired with their spoken description to train a machine learning system. Such a system could, for instance, help her to detect and locate specific keywords in the collected data bank.
\paragraph{Delivering humanitarian aid.} Imagine a foreign doctor providing medical support for natural disaster victims. She needs to be able to communicate with the victims in order to know the treatment to administer. Unfortunately, these victims only speak their own language, which she does not understand, and human translators are scarce. In this scenario, VGS technology capable of detecting and localising keywords will be beneficial
for rapidly scanning victim’s utterances for the occurrence of keywords of interest. Ideally, she should be able to provide the keywords in her native language, and the system should then spot occurrences of the keywords in the foreign speech (a task referred to as cross-lingual keyword spotting~\cite{kamper2018}). Such a system might be far from perfect. However, when faced with the alternative of having no translation system for an unknown language in an emergency, the imperfect system could be of great use~\cite{besacier2014}.
\subsection{Background}
\label{subsec:background}
The above examples illustrate some practical settings in which VGS models can be employed. Below we briefly review the main existing work on VGS modelling. A full review of the literature is given in Chapter~\ref{chap:related_work}. But here the goal is to provide the background to create the context for our research questions and contributions (described in the sections that follow).

\begin{figure}[t]
	\centering
	\includegraphics[width=0.99\columnwidth]{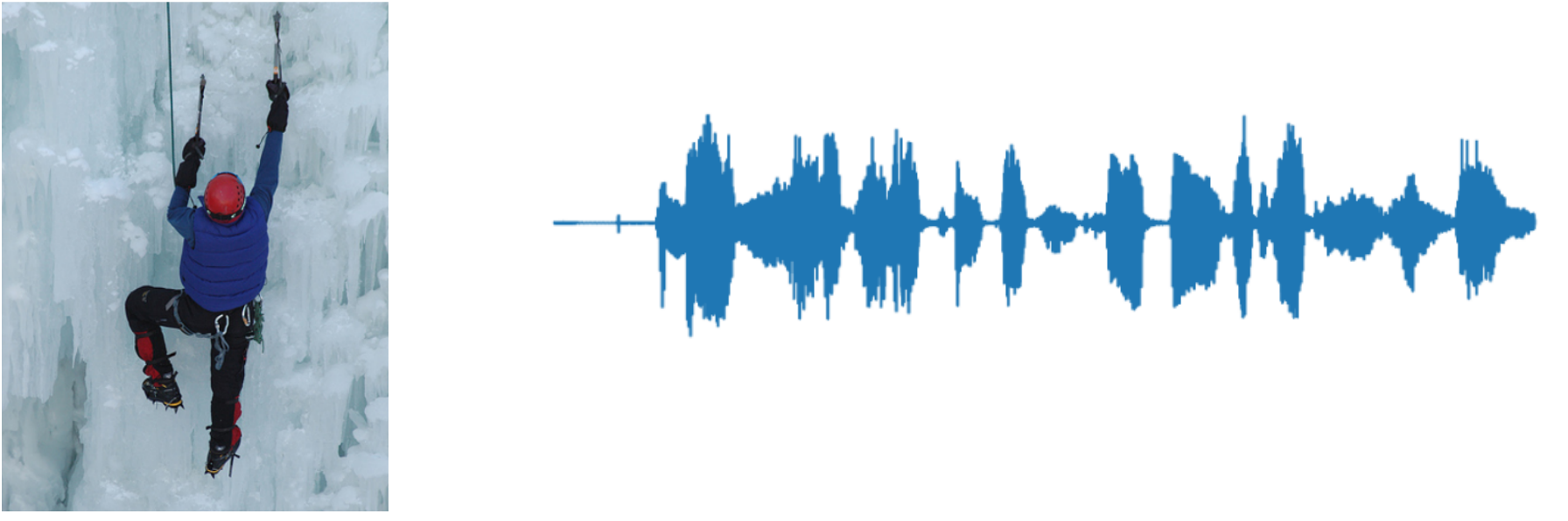}
	\caption{A single training data item in the Flickr Audio Caption Corpus~\cite{harwath2015} dataset consisting of a still image (left) and its spoken caption (right). Both the image and utterance are unlabelled.}
	\label{fig:vgs_dataset_intro}
\end{figure}
\begin{figure}[!t]
	\centering
	\includegraphics[width=0.95\columnwidth]{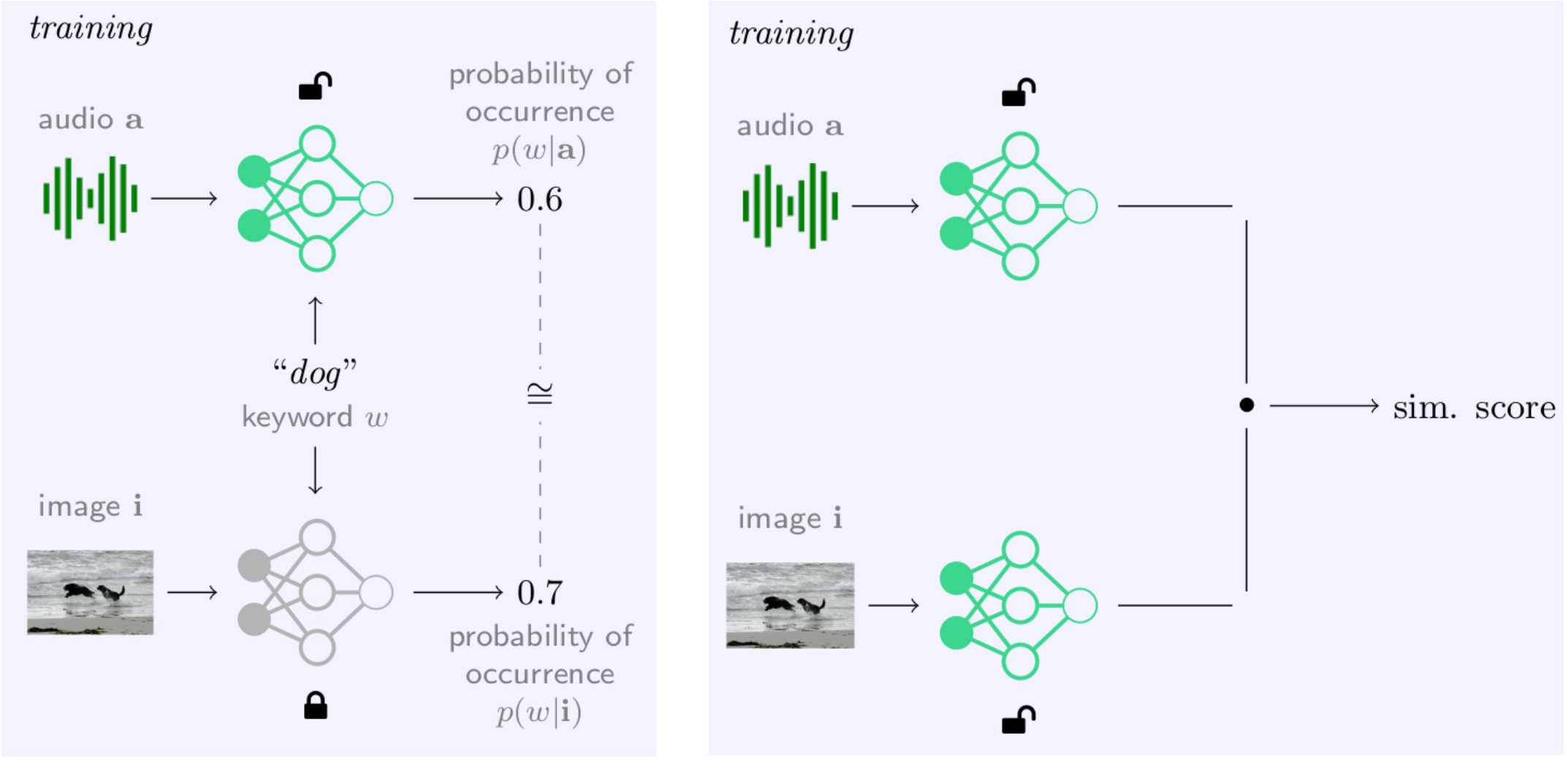}
	\caption{Left: An illustration of the framework for training a VGS model for keyword detection using the approach of Kamper~\etal~\cite{kamper2017a}. Right:  An illustration of the framework for training a VGS model for cross-modal retrieval using the approach of Harwath~\etal~\cite{harwath2016}.
	}
	\label{fig:training_kamper_harwath}
\end{figure}

VGS models learn from images paired with unlabelled utterances, as illustrated in Figure~\ref{fig:vgs_dataset_intro}. 
In this visually supervised setting, both the utterance and the paired image are unlabelled. Some form of self-supervision is therefore required to train a VGS system. One approach is to train a model that projects images and speech into a joint embedding space~\cite{synnaeve2014b, harwath2015, harwath2016, chrupala2017, peng2021}. Such models are typically trained to optimise a ranking-based criterion~\cite{karpathy2014a} such
that images and captions that belong together are more similar in the embedding space than mismatched image--caption pairs. This is illustrated on the right of Figure~\ref{fig:training_kamper_harwath}. Such a system can be used for cross-modal retrieval: given a large collection of images, the task is to retrieve the image that matches a given spoken caption (or the other way around, where an image query is given and the task is to find the correct spoken utterance from a large audio bank).

\begin{figure}[!t]
	\centering
	\includegraphics[width=0.95\columnwidth]{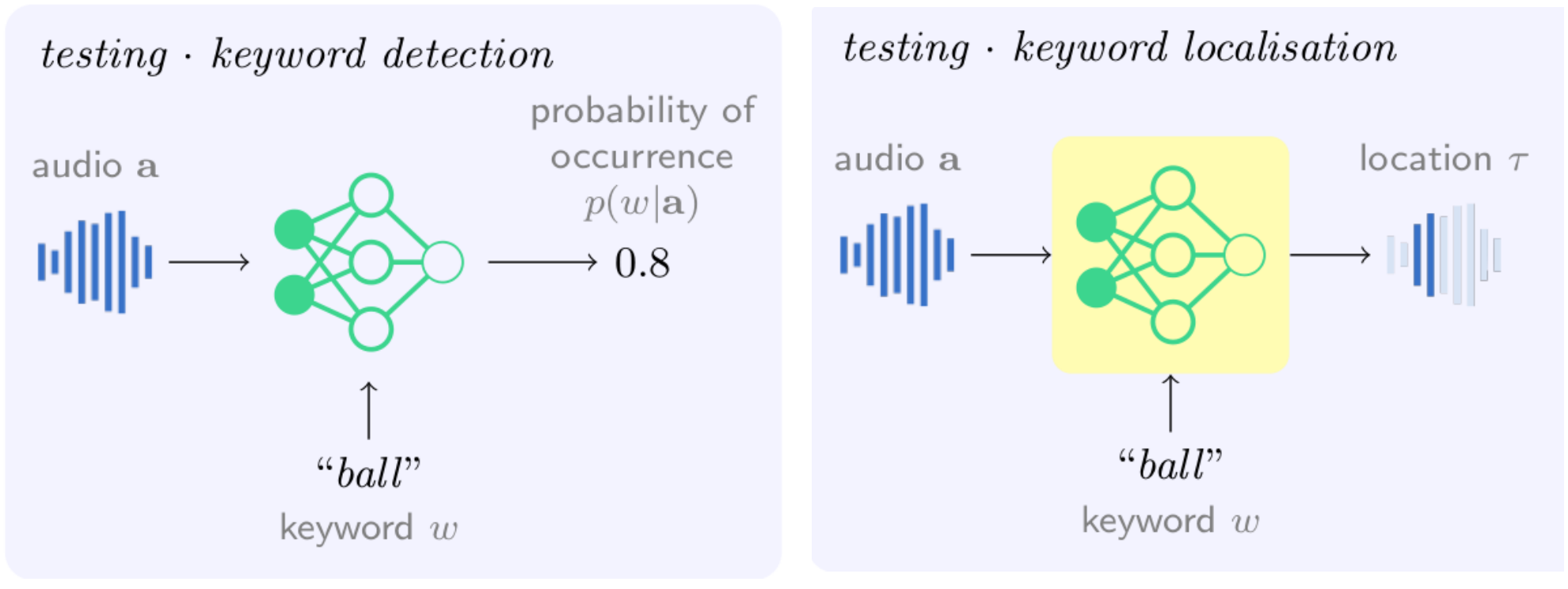}
	\caption{Left: At test time, a VGS network can be used to identify whether a certain keyword occurs in any given utterance (keyword detection). Right: One of the primary goals of this dissertation is to consider whether a VGS model can also be used to locate a keyword in an input utterance (keyword localisation).
	}
	\label{fig:eval_detection_localisation}
\end{figure}

While models that can define a joint image-speech space (or models that can move between spaces) are useful in themselves because they allow for cross-modal retrieval, they do not provide a mapping of speech to textual labels.
This was addressed by Kamper~\etal~\cite{kamper2017a, kamper2019b}, which we use as the starting point of our VGS models throughout this dissertation (full details are given in Section~\ref{subsec:cross-modal-distillation}). Their approach, illustrated on the left of Figure~\ref{fig:training_kamper_harwath}, involves the use of an offline image tagger to obtain soft text labels, which then serves as noisy labels for training a convolutional neural network. At test time, the trained speech model is then used to map input utterances to keywords (without knowing the order or the count of the keywords). 
Without seeing any transcriptions, the authors showed that such VGS models can do keyword detection at a relatively high precision~\cite{kamper2017a}. As illustrated on the left of Figure~\ref{fig:eval_detection_localisation}, keyword detection involves predicting whether a query keyword appear anywhere in an utterance. At test time the trained model receives only an utterance (no image) and a query keyword, and produces a probability for the occurrence of the keyword in the utterance.

These keyword-based VGS models (which can be seen as models that perform a coarse form of speech recognition), were also used to perform the task of keyword spotting. Keyword spotting
is related but different from keyword detection. In keyword spotting, the goal is to rank a collection
of utterances according to how likely they are to contain a given written keyword query. This is different from keyword detection where the task is to determine whether or not the query occurs in a single utterance.
Again, Kamper~\etal~\cite{kamper2017a} showed that a VGS model, trained without
any transcription, can perform a limited degree of keyword spotting. They showed that
these models can find not only exact matches, but also semantic matches~\cite{kamper2019b, kamper2019a}. 
For example, given the query ``children", the system would retrieve utterances such as ``young boys playing soccer in the park".

Apart from the tasks of retrieval, keyword detection and keyword spotting, more recent studies
have also used VGS methodologies in a range of other tasks and in other modelling approaches. For example, some studies proposed multilingual variants where VGS systems are trained with two or more spoken languages~\cite{harwath2018b, azuh2019, ohishi2020a, ohishi2020b}, allowing systems to perform cross-lingual retrieval, i.e. retrieving content in one language when presented with a query in the other. Other recent work has considered using videos instead of still images~\cite{ngiam2011, huang2013, mroueh2015, afouras2018, petridis2018}. In this dissertation, as mentioned in the preceding paragraphs, we limit ourselves to the setting where we have still images and their spoken captions.

Despite VGS modelling being used to tackle the tasks mentioned above, none of these previous studies have considered the task of keyword localisation.
Keyword localisation involves finding where in a speech utterance a given query keyword occurs, as illustrated on the right of Figure~\ref{fig:eval_detection_localisation}. One of the primary goals of our work is to address this shortcoming.

We now turn to another major shortcoming of work on VGS modelling, which we also aim to address in this work. Almost all of the studies mentioned above are based on English. To illustrate this, we briefly review the most relevant VGS datasets. Notable among these is the Flickr Audio Captions Corpus (FACC)~\cite{harwath2015}. FACC is a multimodal dataset consisting of the same images as in the Flickr8k text--image dataset~\cite{rashtchian2010,hodosh2015}. 40k spoken captions of the images were recorded by crowd workers on Amazon's Mechanical Turk from the five textual captions available for each of the 8k images contained in the Flickr8k dataset. Temporal alignments were also obtained for words uttered in the spoken captions using a forced-alignment system. The release of the FACC dataset has subsequently enabled several VGS tasks, such as image--speech alignment \cite{harwath2015}, semantic speech retrieval \cite{kamper2019b}, image-to-speech synthesis \cite{hasegawa2017} and speech-to-image generation \cite{li2020}. Other bigger speech-image datasets that have been developed are based on the MIT Places image dataset~\cite{zhou2014b}, which contains about 7M labelled pictures of scenes. This image dataset has been augmented with 400k spoken captions in the English language~\cite{harwath2016}, 100k spoken captions in the Hindi language~\cite{harwath2018b}, and 100k spoken captions in the Japanese language~\cite{ohishi2020a}. All of these languages, however, can be considered well-resourced. In summary, there is therefore currently no VGS dataset which allows investigations in a truly low-resource setting.

Despite this shortcoming in VGS work, many of the VGS studies state as a primary motivation the application of VGS models in low-resource settings. Most of the studies therefore use English as a proxy for a low-resource language. As an example, one very relevant study~\cite{kamper2018} simulated a low-resource environment, treating English as the
low-resource language and German as the high-resource language. Specifically, the authors used a visual tagger producing German text labels as targets for the speech network, i.e. the same methodology as in Figure~\ref{fig:training_kamper_harwath} (left) is followed, but now instead of producing English tags for an image, German tags are produced (Figure~\ref{fig:training_kamper_harwath} left-bottom). The result is a speech model (Figure~\ref{fig:training_kamper_harwath} left-top) that takes in English speech and produces German text labels. More concretely, for the task of cross-lingual keyword spotting, their model can be given a written German keyword (for example, ``Hunde", the German word for ``dogs") and asked to retrieve English speech utterances containing that keyword (for example, ``two dogs playing outside near the water"). Previous studies on cross-lingual keyword spotting adopted
a direct approach of using text transcriptions for training the system instead of using visual grounding~\cite{sheridan1997}. This was therefore a milestone study in that it showed that cross-lingual keyword spotting (and also detection) can be performed using a VGS model---an approach that could be used in low-resource settings. However, English was still used to simulate the low-resource language, and thus it is unclear whether the approach would work in a realistic low-resource scenario. (As a side-note,~\cite{kamper2018} also only considered keyword detection and spotting; keyword localisation was not considered.)

From the above, it is clear that even the studies that attempted to explicitly consider
low-resource environments, did so in an artificial manner. We address this by collecting a new image--speech dataset in a real low-resource language, and then illustrate (for the first time) how cross-lingual keyword detection and localisation can be performed with a VGS model trained on a real low-resource language.
\section{Research questions}
\label{sec:research_questions}
From the background given above, we identify the following concrete research questions that we address in this dissertation. 
	\paragraph{Research question 1.} \textit{Is keyword localisation possible with VGS models?}
	As established in Section~\ref{subsec:background}, while VGS models have been previously used for keyword spotting and detection, they have not been used for keyword localisation. As a reminder, keyword localisation is the task of finding where in an utterance a given query keyword occurs, as illustrated on the right of Figure~\ref{fig:eval_detection_localisation}. Because most of the previous work on keyword spotting and detection has been done in an artificial low-resource setting using English VGS data, we will also first consider this research question in this artificial setting. This allows us to compare to and directly extend previous work. But then, in the next research question, we consider a real low-resource scenario.
	
	\paragraph{Research question 2.} \textit{Can we do visually grounded keyword localisation cross-lingually in a real low-resource setting?} 
	The analyses of the VGS systems carried out in an artificial setting where English is treated as a low-resource language will help to address the question of whether keyword localisation is possible with VGS models. But it will not address the question of how VGS systems will perform in a real low-resource setting. For a low-resource spoken language, the question is whether we can build a VGS system that can locate words cross-lingually. As a reminder, the task of cross-lingual keyword spotting involves retrieving utterances in
	one language containing a given keyword in another language~\cite{kamper2018}. The question is therefore, firstly, whether this can be done with a VGS model in a real low-resource language, and then, secondly, whether a model can also be used for cross-lingual keyword localisation (a task not considered in any study before).
	
\section{Methodology}
\label{sec:methodology}
Throughout this dissertation, we use the approach of Kamper~\etal~\cite{kamper2017a} mentioned in Section~\ref{subsec:background}, as a starting point. As a reminder, their approach involves training a VGS model for keyword detection and spotting using soft label tags extracted from images with an offline image tagger. We specifically augment their approach with the ability to also localise keywords.

Since no location information is available at training time, we have to use the intrinsic ability of the VGS models to localise~\cite{oquab2015}.
Specifically, we propose four different localisation methods to adapt the models for keyword localisation. One method employs the popular explainability method  \textit{Grad-CAM}~\cite{selvaraju2017}, which was originally developed in the context of images. This method can be used within any convolutional neural network architecture. The second and third methods use the fact that for certain architectures, the internal activations can be interpreted as localisation scores. Concretely, the second approach (\textit{score aggregation}) assumes that the final pooling layer is a simple transformation (for instance, average or max pooling) so that the output score can be regarded as an aggregation of local scores. These can be used to select the most likely temporal location for the query keyword. The third approach (\textit{attention}) assumes that the network involves an attention layer that pools features over the temporal axis; we use the attention weights as localisation scores. Our attention framework is heavily inspired by~\cite{tamer2020}. The fourth approach (\textit{input masking}) involves masking the input signal at different locations and measuring the response score predicted by the trained model on the partial inputs; large variations in the output suggest the presence of a keyword. Input masking is similar to the first approach because it can be applied to any classification model after training, irrespective of architecture. 

The methodology described up to this point helps with tackling the first research question. To tackle the second research question, we use as starting point the approach of Kamper and Roth~\cite{kamper2018}, briefly described in Section~\ref{subsec:background} and described in detail in Section~\ref{sec:previous_work_chapter5}. As a reminder, the idea behind their approach is to obtain soft labels for images in one (well-resourced) language and then pair these soft tags with speech from another (low-resource) language (they simulated this setting artificially using two well-resourced languages).
To realistically assess the cross-lingual keyword localisation performance of a VGS system in a real low-resource setting, we collect a new dataset from a real low-resource language, \yoruba. The \yoruba language is one of the three official languages in Nigeria, and even though it is spoken natively by more than 44M speakers, there are only a handful of high-quality datasets available in this language~\cite{vanniekerk2014a,gutkin2020,meyer2022,ogayo2022}. We then tag images associated with corresponding \yoruba speech with English labels using an offline image tagger, allowing us to train a VGS model to perform cross-lingual keyword detection with an English keyword and \yoruba speech. Then we apply the localisation methods, developed for the first research question, for the first time in a real low-resource setting to do cross-lingual keyword detection and localisation.
\section{Contributions}
\label{sec:contributions}
This dissertation makes the following specific contributions to the body of knowledge on VGS modelling.
\kayode{}

\paragraph{Research contribution 1.} A new VGS model for keyword detection and keyword spotting using attention, and a thorough comparison of this approach to existing VGS-based methods.
\paragraph{Research contribution 2.} An extension of existing VGS models for keyword localisation by proposing four localisation methods.
\paragraph{Research contribution 3.} A detailed quantitative and qualitative analysis revealing the limits of the models above, showcasing their success and failure modes. This is carried out in a setting where English VGS data is artificially considered as a low-resource language.
\paragraph{Research contribution 4.} A new multimodal, multilingual dataset which enables VGS modelling in a real low-resource setting resembling language documentation setting. The dataset extends the Flickr8k image--text dataset~\cite{rashtchian2010,hodosh2015} to \yoruba with three modalities: \yoruba translations of 6k of its captions; corresponding spoken recordings of these translations by a single speaker; and temporal alignments of 67 \yoruba keywords for a subset of 500 of the captions.
\paragraph{Research contribution 5.} A baseline system for cross-lingual keyword detection and keyword localisation
in a real low-resource setting. The baseline system is trained  on 5k \yoruba utterances paired with English soft text labels, producing a speech network that detects and localise an English keyword in \yoruba speech. We also show that this system can be greatly improved by first pretraining it on the larger English speech--image dataset.
\paragraph{Research contribution 6.} Multi-faceted analyses of the cross-lingual VGS models above. The analyses help to better understand how performance on the keyword localisation task varies across the 67 keywords in the system's vocabulary, and further help to interpret the failure modes of the cross-lingual model.
\subsection{Publications}
Below is the list of published contributions that resulted from the work presented in this dissertation.\\

\begin{tcolorbox}[width=\linewidth, colback=white!95!black, boxrule=0.5pt]
	\small
	\textit{Research paper 1: (Workshop)} \\
	K. Olaleye, B. van Niekerk, and H. Kamper, ``Towards localisation of keywords in speech using weak supervision,'' \textit{NeurIPS Self-Supervised Learning for Speech and Audio Processing Workshop}, 2020. \\
	
	\textit{Research paper 2: (Conference)} \\
	K. Olaleye, and H. Kamper, ``Attention-based keyword localisation in speech using visual grounding,'' \textit{Interspeech}, 2021. \\
	
	\textit{Research paper 3: (Journal)}\\
	K. Olaleye, D. Oneață, and H. Kamper, ``Keyword localisation in untranscribed speech using visually grounded speech models,'' \textit{Journal of Selected Topics in Signal Processing}, 2022.\\
	
	\textit{Research paper 4: (Conference)}\\
	K. Olaleye, D. Oneață, and H. Kamper, ``YFACC: A \yoruba speech--image dataset for cross-lingual keyword localisation through visual grounding,'' \textit{IEEE Spoken Language Technology (SLT)}, 2023.
	
\end{tcolorbox}

\subsection{Code repositories}

\begin{tcolorbox}[width=\linewidth, colback=white!95!black, boxrule=0.5pt]
	\small
	\textit{Research paper 1: Towards localisation of keywords in speech using weak supervision} \\
	\href{https://github.com/kayodeolaleye/keyword_localisation_speech/tree/main/SAS}{Code URL}\\
	
	\textit{Research paper 2: Attention-based keyword localisation in speech using visual grounding} \\
	\href{https://github.com/kayodeolaleye/keyword_localisation_speech/tree/main/Interspeech}{Code URL} \\
	
	\textit{Research paper 3: Keyword localisation in untranscribed speech using visually grounded speech models}\\
	\href{https://github.com/kayodeolaleye/keyword_localisation_speech/tree/main/Interspeech_no_Trimming}{Code URL}\\
	
	\textit{Research paper 4: YFACC: A \yoruba speech--image dataset for cross-lingual keyword localisation through visual grounding}\\
	\href{https://github.com/kayodeolaleye/keyword_localisation_speech/tree/main/Cross_lingual_localisation}{Code URL 1}, \href{https://github.com/kayodeolaleye/keyword_localisation_speech/tree/main/English_5000}{Code URL 2}\\ \href{https://kamperh.com/yfacc}{Data URL}
	
\end{tcolorbox}

\section{Dissertation outline}
\label{sec:structure_overview}
The purpose of this section is to present a roadmap of what to expect in terms of the structure of this dissertation. We provide a brief summary of each chapter's purpose and content. 
\paragraph{Chapter~\ref{chap:related_work}: Related work.}
In this chapter, we review in depth previous studies related to the topic of this dissertation. We first review the various datasets that have been developed and curated to support multimodal learning---focussing primarily on three modalities: text, visual, and audio. We review different methods for exploiting complementary information from multiple modalities. We then review studies that focussed on analysing and explaining some of the multimodal systems, particularly VGS.
Furthermore, we survey other types of multimodal learning (both within and outside the speech domain) that predate VGS namely audio-visual speech recognition, which involves training an ASR systems by leveraging visual information available during speech production, and multimodal learning from images and text. Finally we review studies on keyword spotting and localisation (with and without visual supervision) which are the core tasks performed in this dissertation.

\paragraph{Chapter~\ref{chap:problem_formulation_setup}: Visually grounded keyword detection and spotting.}
In this chapter we present VGS models for keyword detection and spotting.
The purpose of this chapter is to partially address our first research question. In particular, we explain the keyword detection models that we will extend in the rest of the dissertation for the task of keyword localisation. We describe one of the VGS datasets used throughout the dissertation: the English FACC corpus mentioned in Section~\ref{subsec:background}. We provide details about the evaluation procedure, and the methodologies employed, including the VGS architecture and implementation details. We also present the baseline and upper bound results for the VGS models on the keyword detection and keyword spotting tasks. 
\paragraph{Chapter~\ref{chap:methods-for-keyword-localisation}: Methods for keyword localisation with VGS models.}
In this chapter we address in full our first research question. We extend the models from the previous chapter (Chapter~\ref{chap:problem_formulation_setup}) for the task of keyword localisation. We propose four different localisation methods to adapt the models for keyword localisation (see Section~\ref{sec:methodology}). We describe the evaluation protocol for visually grounded keyword localisation. We present the performance of the proposed keyword localisation methods, and perform a systematic and fair comparison of the four different keyword localisation methods. Furthermore, we provide a detailed qualitative analyses showcasing the success and failure modes of the models on the keyword localisation task.
\paragraph{Chapter~\ref{chap:cross_lingual_keyword_localisation}: Cross-lingual keyword localisation.}
 In this chapter we address our second research question. We move towards a more realistic setting of the tasks considered in this dissertation by introducing a multilingual and multimodal dataset that enables training and evaluating of VGS models in a real low-resource setting. We collect and release a new single-speaker dataset of audio captions for 6k of the Flickr8k images in \yoruba---a real low-resource language spoken in Nigeria. We train an attention-based VGS model where images are automatically tagged with English visual labels and paired with \yoruba utterances. We describe the evaluation criterion used to assess the performance of the models on cross-lingual keyword localisation. We present details about the architecture and implementation of the VGS systems. Finally, we present the quantitative and qualitative analyses of our visually grounded cross-lingual system.

\paragraph{Chapter~\ref{chap:conclusion}: Summary and conclusions.}
This chapter summarises and highlights the main findings from the previous chapters, presents direct answers to the research questions posed in Section~\ref{sec:research_questions}, and finally presents avenues for possible future work.

%% file: related_work.tex
\chapter{Related work}
\label{chap:related_work}
In this chapter, we review previous studies related to the topic of this dissertation. We first review the various datasets that have been developed and curated to support multimodal learning (Section~\ref{sec:audio_visual_datasets})---focussing primarily on three modalities: texts, visual signals (either image or video), and audio. However, all experiments conducted in this dissertation are performed specifically on audio and still images (at training time) and audio-only (at test time). We review different methods for exploiting complementary information from multiple modalities in Section~\ref{sec:methodologies_multimodal_learning}. We then review studies that focussed on analysing and explaining multimodal systems in Section~\ref{sec:analysis_vgs_systems}, focussing particularly on visually grounded speech (VGS) models. As a reminder, VGS models learn from images and their spoken captions, and these are the type of models we consider throughout this dissertation.

We take an even closer look at VGS systems in Section~\ref{sec:visually_grounded_speech_models}, reviewing studies most related to this dissertation---specifically studies on VGS models that considered the cross-modal distillation approach described in Section~\ref{subsec:cross-modal-distillation}. These VGS systems are specifically trained to perform coarse (non-verbatim) speech recognition. By coarse speech recognition we mean recognising isolated keywords (either semantic or exact) that are uttered in speech audio. For example, given an utterance with the transcription \textit{three dogs in the field looking at something in the grass}; a coarse speech recognition involves the use of VGS model to predict whether the word \textit{grass} is present in the utterance. This is different from the traditional automatic speech recognition (ASR) systems which are assessed on their ability to produce the exact transcription of the entire utterance. Furthermore, we review other types of multimodal learning (both within and outside the speech domain) that predate VGS modelling. This includes audio-visual speech recognition (AVSR) which involves training an ASR system by leveraging visual information available during speech production (Section~\ref{sec:avsr}), and multimodal learning from images and text (Section~\ref{sec:learning-from-images-and-text}). Finally we review studies on keyword spotting and localisation with and without visual supervision (Section~\ref{sec:speech_spotting_localisation}). These are the core tasks that we try to tackle in this dissertation, specifically by using VGS models.
\section{Multimodal datasets: text, vision, and speech}
\label{sec:audio_visual_datasets}
There are many different types of modalities that can  be combined to create multimodal datasets. Therefore, in order to make this section more focussed on the core subject of this dissertation, we only cover datasets involving three modalities: texts which can be both written or spoken, vision which can be either still image or video, and audio. A multimodal dataset is created by a combination of two or more of these modalities.
\subsection{Still images with speech}
\label{subsec:vgs_datasets_chp2}
One of the key factors responsible for the progress in the development of VGS models is the availability of VGS datasets~\cite{chrupala2022}. Efforts have gone into collecting spoken descriptions of images depicting everyday objects and situations in order to facilitate the development of modelling approaches as well as enable speech technology for downstream tasks in the absence of conventional supervision. See~\cite{chrupala2022} for a recent survey of some of the existing VGS datasets. 

Table~\ref{tbl:vgs_corpora_overview_modalities} presents the modalities of some of these datasets, and Table~\ref{tbl:vgs_corpora_overview_quantitative} presents their quantitative characteristics. Among all these publicly available VGS datasets, the most relevant to this dissertation is the Flickr Audio Captions Corpus (FACC) (see Section~\ref{subsec:FACC} for more details) introduced by Harwath and Glass~\cite{harwath2015}. This dataset is based directly on Flickr8k~\cite{rashtchian2010,hodosh2015}. The original Flickr8k is a bimodal (image and text) dataset consisting of 8k images and 40k English text captions. FACC extended Flickr8k by using Amazon's Mechanical Turk to crowdsource English audio captions for each of the text captions. The release of the FACC dataset has subsequently enabled several visually-grounded tasks, such as image--speech alignment \cite{harwath2015}, semantic speech retrieval \cite{kamper2019b}, image-to-speech synthesis \cite{hasegawa2017} and speech-to-image generation \cite{li2020}. In this dissertation we further augment the Flickr8k dataset with a truly low-resource language (\yoruba), by collecting spoken captions and manual alignments with start and end timestamps for selected keywords to enable cross-lingual keyword spotting and localisation for an under-resourced language (details in Chapter~\ref{chap:cross_lingual_keyword_localisation}). 

Other bigger speech--image datasets that have been developed are based on the MIT Places image dataset~\cite{zhou2014b} which contains about 7M labelled pictures of scenes. This image dataset has been augmented with 400k spoken captions in the English language~\cite{harwath2016}, 100k spoken captions in the Hindi language~\cite{harwath2018b}, and 100k spoken captions in the Japanese language~\cite{ohishi2020a}. All of these languages can be regarded as well-resourced. 

\begin{table*}[t]
	\centering
	\caption{%
		Overview of audio-visual datasets with their corresponding modalities
	}
	\begin{tabular}{@{}l@{}lccccccc@{}}
		\toprule
		&  & \multicolumn{6}{c}{Modalities}                                         &  \\
		\cmidrule(lr){4-8}
		Dataset                                   & & Lang. & Audio & Text & Images & Alignments    \\
		\midrule
		Flickr audio caption corpus (2015)~\cite{harwath2015} & & en & \cmark & \cmark    &     \cmark  &  \cmark   \\
		English Places Audio Captions (2016)~\cite{harwath2016}            & & en & \cmark & \cmark    &   \cmark &        &              \\
		Synthetically spoken COCO (2017)~\cite{chrupala2017}          & & en & \cmark & \cmark    &    \cmark       &        &              \\
		Synthetically spoken STAIR (2017)~\cite{yoshikawa2017}           &  & ja & \cmark & \cmark    &  \cmark \\
		Speech-COCO (2017)~\cite{havard2017}            & & en & \cmark & \cmark    &    \cmark & \cmark  \\
		Hindi Places Audio Captions (2018)~\cite{harwath2018b}            & & hi & \cmark & \cmark    &  \cmark &   \\
		Conceptual Spoken Captions (2019)~\cite{ilharco2019}            & & en & \cmark & \cmark    &  \cmark &   \\
		SpokenCOCO (2020)~\cite{hsu2020}                          &    & en & \cmark & \cmark  & \cmark &  \\
		Japanese Places Audio Captions (2020)~\cite{ohishi2020a}  &    & ja & \cmark & \cmark  & \cmark &  \\
		\bottomrule
	\end{tabular}
	\label{tbl:vgs_corpora_overview_modalities}
\end{table*}

\begin{table*}[t]
	\centering
	\caption{%
		Overview of audio-visual datasets with their quantitative characteristics. 
	}
	\begin{tabular}{@{}l@{}lccccc@{}}
		\toprule
		&  & \multicolumn{1}{c}{Speakers}         & \multicolumn{1}{c}{Duration}        & \multicolumn{1}{c}{Utterances}       & \multicolumn{1}{c}{Sampling} \\
		Dataset                                   &  & \multicolumn{1}{c}{\mylabel{number}} & \multicolumn{1}{c}{\mylabel{hours}} & \multicolumn{1}{c}{\mylabel{number}} & \multicolumn{1}{c}{\mylabel{kHz}} \\
		\midrule
		Flickr audio caption corpus (2015)~\cite{harwath2015} &  &  183 & 36 & 40000 & 16 \\
		English Places Audio Captions (2016)~\cite{harwath2016}           &  & 2683  & 936 & 400000 & 16 \\
		Synthetically spoken COCO (2017)~\cite{chrupala2017}             &  & 1 & 601 & 616767 & 24 \\
		Synthetically spoken STAIR (2017)~\cite{yoshikawa2017}             &  & 1 & 793 & 616767 & 24\\
		Speech-COCO (2017)~\cite{havard2017}                              &  & 8 & 601 & 616767 & 16 \\
		Hindi Places Audio Captions (2018)~\cite{harwath2018b}            &  & 139 & 316 & 100000 & 16 \\
		Conceptual Spoken Captions (2019)~\cite{ilharco2019}            &  & 6 &  & 3.3M & 16 \\
		SpokenCOCO (2020)~\cite{hsu2020}                              &  & 2352 & 742 & 605000 & 48 \\
		Japanese Places Audio Captions (2020)~\cite{ohishi2020a}	&  & 303 & 540 & 98555 & 16 \\
		\bottomrule
	\end{tabular}
	\label{tbl:vgs_corpora_overview_quantitative}
\end{table*}
\subsection{Video datasets}
Beyond image--speech datasets described above, there are also video datasets which contain either written or spoken (or both) descriptions related to the activity depicted in the video. These datasets are sourced from a range of contexts including cooking~\cite{regneri2013, das2013, rohrbach2014, rohrbach2016, damen2018, zhou2018, damen2020}, sports~\cite{soomro2012, idrees2017, karpathy2014b}, movies~\cite{rohrbach2015, rohrbach2017} or general activities~\cite{miech2019}. These datasets target tasks pertaining to video understanding, such as video description~\cite{chen2011, rohrbach2014, rohrbach2015, xu2016, rohrbach2017, gella2018} and event detection~\cite{rohrbach2016, krishna2017b, damen2018, damen2020}.

Regneri~\etal~\cite{regneri2013} presented a multimodal dataset that aligns videos with multiple natural language descriptions of the actions depicted in the videos, together with an annotation of how similar the action descriptions are to their paired video segments. The purpose of the dataset is to demonstrate the impact of video scenes on a text-based model of action descriptions.
Das~\etal~\cite{das2013} collected a video dataset from YouTube to train a model to perform a video description task. The dataset consists of 88 videos and includes textual annotations of objects depicted. Zhou~\etal~\cite{zhou2018} introduced a multimodal dataset consisting of 2k instructional videos downloaded from YouTube. The dataset includes temporal location information and textual descriptions of procedure segments from each of the videos. The dataset is used to facilitate the task of segmenting a (visual only) video into category-independent procedure segments.
The Spoken Moments in Time (S-MiT) dataset~\cite{monfort2021a} includes spoken audio captions for 500k unique three-second clips each with different source videos from the Moments in Time dataset. S-MiT contains spoken descriptions of elements depicted in the videos rather than textual description, allowing speech--video models to be trained jointly.
The MP II movie description corpus~ \cite{rohrbach2015} contains transcribed audio descriptions from 94 Hollywood movies split into 68k clips where each clip is paired with a sentence from the movie script and an audio description of the visual content in that clip. Similarly, the Large Scale Movie Description Challenge dataset~\cite{rohrbach2017} contains 200 movies with 120k sentence descriptions. VideoStory~\cite{gella2018} contains 20k social media videos where each video contains a paragraph length description. 
The ActivityNet Captions dataset~\cite{krishna2017a} has 20k videos with 100k text descriptions. The Microsoft Video
Description dataset~\cite{chen2011} contains 2k YouTube clips with a 10-25 second duration and an average of 41 single sentence descriptions per clip. MSR-Video to Text~\cite{xu2016} contains 7k videos split into 10k clips with 20 captions per video.
HowTo100M~\cite{miech2019} contains 136M clips sourced from 1.22M instructional videos with spoken narrations generated from subtitles associated with each video. 
VaTeX~\cite{wang2019} contains 41k videos annotated with 10 English captions and 10 Chinese captions for multilingual captioning.

All of these video datasets make it possible to learn models for downstream tasks multimodally. The models considered in this dissertation are all trained on spoken captions with still images. Extending the approaches developed in this dissertation to deal with video could be an interesting direction for future work.
\section{Methodologies for multimodal learning}
\label{sec:methodologies_multimodal_learning}
The success of multimodal learning is largely dependent on how well the representation of each modality interacts to leverage the complementary information among them~\cite{baltruvsaitis2018}.
In this section we review studies that introduced different approaches to multimodal learning. Some of the approaches involve learning representations from each modalities and then fusing the representations together in either a joint space (Section~\ref{subsec:joint_representation_learning}) or a coordinated space (Section~\ref{subsec:coordinated_representation_learning}). Another approach for multimodal learning involves cross-modal distillation (Section~\ref{subsec:cross-modal-distillation}), where knowledge is transferred from one modality (the teacher model) to another modality (student model). 
\subsection{Joint representation}
\label{subsec:joint_representation_learning}
\begin{figure}[t]
	\centering
	\includegraphics[width=0.99\columnwidth]{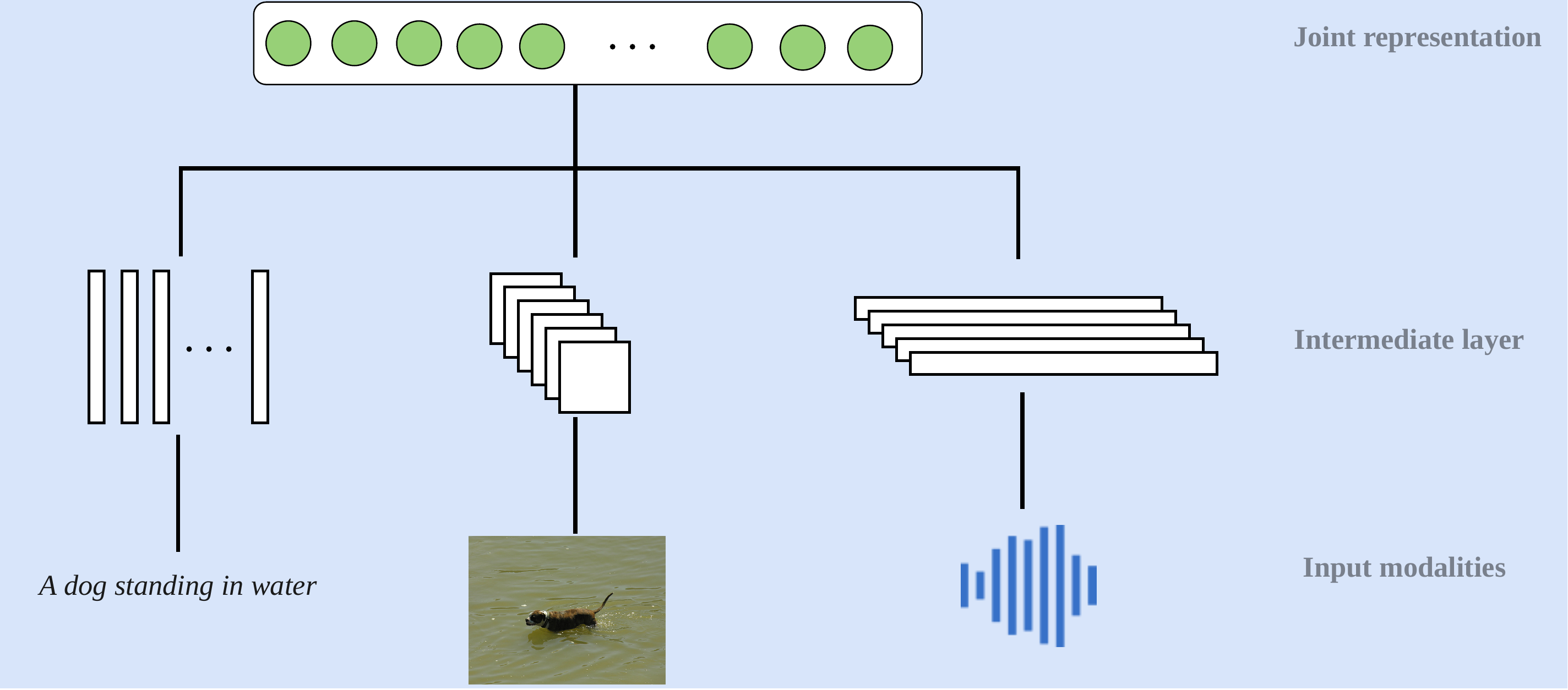}
	\caption{Joint representations are projected into the same embedding space. In this figure, three modalities are given: text, still image and speech.}
	\label{fig:joint_gen_rep}
\end{figure}

\begin{figure}[t]
	\centering
	\includegraphics[width=0.99\columnwidth]{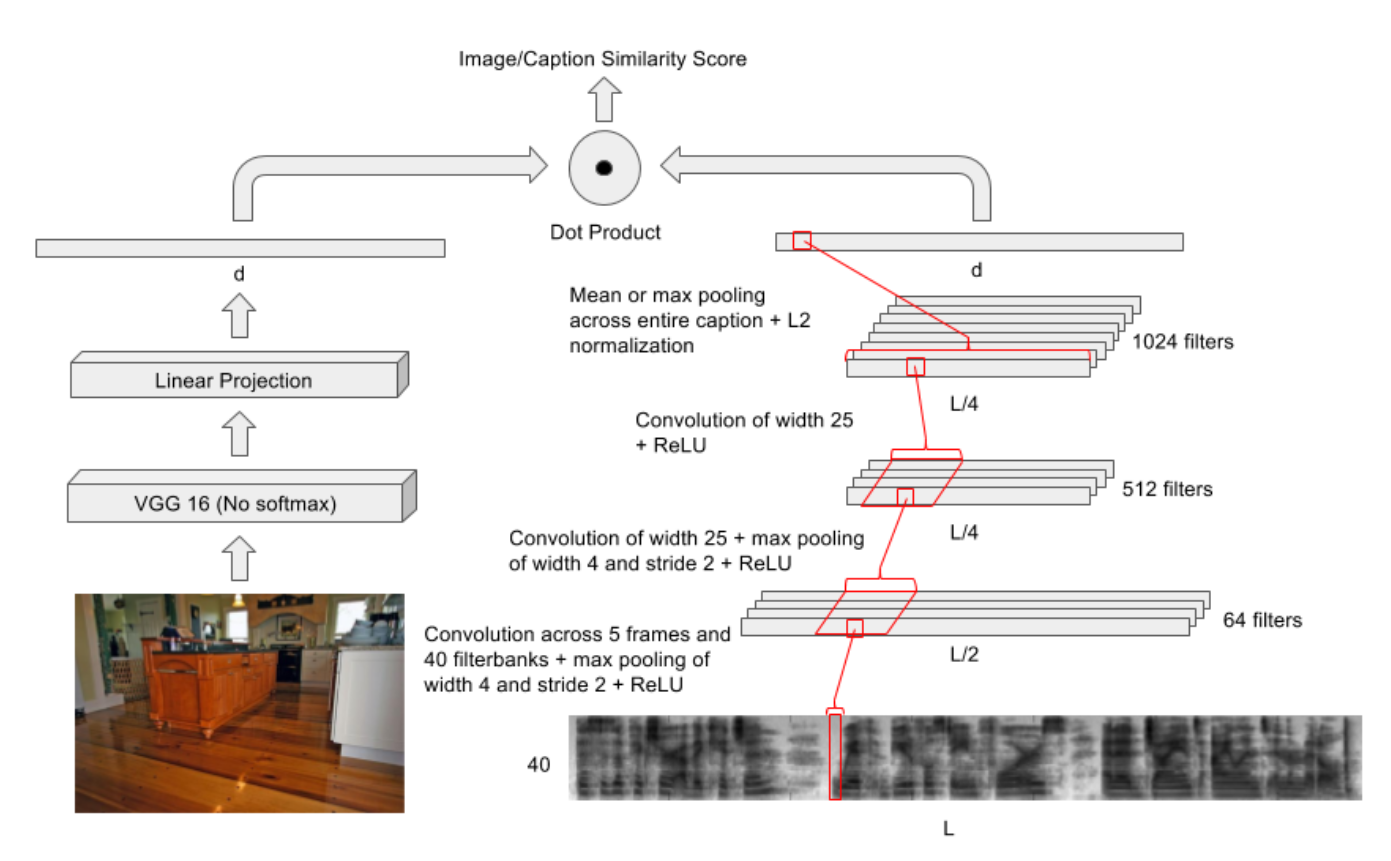}
	\caption{The joint speech-image embedding space with embedding dimension $d$ and the caption length $L$. The left branch processes the image and the right branch processes the speech input. Both branches are subsequently tied together using a dot product~\cite{harwath2016}.
	}
	\label{fig:joint_rep}
\end{figure}
In the joint representation methodology, the representation of each modality is projected into the same multimodal space~\cite{baltruvsaitis2018}. Figure~\ref{fig:joint_gen_rep} presents an illustration of the joint representation method where a still image and its texual and spoken captions are projected into a common embedding space. Several studies have been done on constructing joint representation from images and texts. The resulting representations have enabled tasks such as visual question answering~\cite{antol2015}, and image captioning~\cite{karpathy2015}. The joint representation approach has also been extended beyond the image and text modalities to also include audio~\cite{mroueh2015}. This has been used for audio-visual speech recognition (AVSR, reviewed in Section~\ref{sec:avsr}), and for video classification~\cite{wu2014b}. 
We will focus this subsection on reviewing studies on the construction of joint representation between still images and their spoken captions. 

One of the earliest studies proposing neural network for constructing  joint representation for image and speech is~\cite{synnaeve2014b}.
The authors used multilayer perceptrons (with ReLU activations) to map speech and image features into the same embedding space. The model learned to associate content words and image fragment that co-occur together using the cosine-squared-cosine loss:
\begin{equation}
	\text{L}_{\cos\cos^2} = 
	\begin{cases}(1-\cos(\mathbf{v}, \mathbf{s})) & \text{ if } \mathbf{v} \text{ and } \mathbf{s} \text{ match}\\ 
		\cos^2(\mathbf{v}, \mathbf{s}) & \text{ if different} 
	\end{cases}
\end{equation}
with $$\cos(\mathbf{e_1},\mathbf{e_2}) = \frac{\mathbf{e_1^{\top}}\mathbf{e_2}}{\left \| \mathbf{e_1}  \right \| \left \| \mathbf{e_2} \right \|}$$
where $\mathbf{v}$ and $\mathbf{s}$ are the output representations for an input image fragment and an input spoken word token, respectively.
The resulting trained VGS model was used for spoken word retrieval given an image, and vice versa.

In seminal studies kicking off much of the work on VGS modelling, Harwath~\etal~\cite{harwath2015, harwath2016} scaled up this approach using a much larger dataset (FACC, reviewed in Section~\ref{subsec:vgs_datasets_chp2}). In~\cite{harwath2016}, the authors used a convolutional neural network to construct representations for each modality. Figure~\ref{fig:joint_rep} presents an illustration of the architecture used by the authors. The model consists of an image branch and a speech spectrogram branch (of length $L$). Each branch produces a $d$-dimensional representation. A dot product is applied to the representations and the output of the dot product operation is taken to represent the similarity between the image and the speech.
The authors also introduced a contrastive margin loss that pushes matching image-speech pairs closer to each other compared to mismatched image--speech pairs; variants of this loss has since become the standard in most VGS models. In~\cite{harwath2015}, the authors also employed a pair of convolutional neural networks to construct image object and
speech representation at the word level; the networks are tied together with an embedding and alignment model which learns a joint semantic space over both modalities. The resulting model was evaluated
on an image retrieval task---asking the model to find which image belongs with a given caption---and an image annotation task---asking the model to search over all the captions to find which one belongs with a given image.

Other network architectures for constructing joint representations have also been explored: \chrupala~\etal~\cite{chrupala2017} used a gated recurrent neural network with attention for the speech branch, while recently Peng and Harwath~\cite{peng2021} proposed a Transformer-based model. {Other studies have proposed multilingual variants where VGS systems are trained with two or more spoken languages with the same or different visual component. Harwath \etal~\cite{harwath2018b} demonstrated that training a bilingual model on spoken English and Hindi captions simulataneously with a common visual component produces VGS system capable of performing semantic cross-lingual speech-to-speech retrieval. Azuh~\etal~\cite{azuh2019} presented a method for the discovery of word-like units and their approximate translations from VGS models across two languages. Ohishi~\etal~\cite{ohishi2020a} explore trilingual VGS models (trained on spoken English, Hindi and Japanese captions) for cross-modal and cross-lingual retrieval, and Ohishi~\etal~\cite{ohishi2020b} proposed a data expansion method for learning a multilingual semantic embedding model using disjoint datasets containing images and their multilingual audio captions with no shared images among the multiple languages.
	
In this dissertation we do not use the joint representation approach to capture the complementary information between the modalities. We instead use as a starting point a different approach which will be reviewed in Section~\ref{subsec:cross-modal-distillation}.
	\subsection{Coordinated representation}
	\label{subsec:coordinated_representation_learning}
	
	\begin{figure}[t]
		\centering
		\includegraphics[width=0.99\columnwidth]{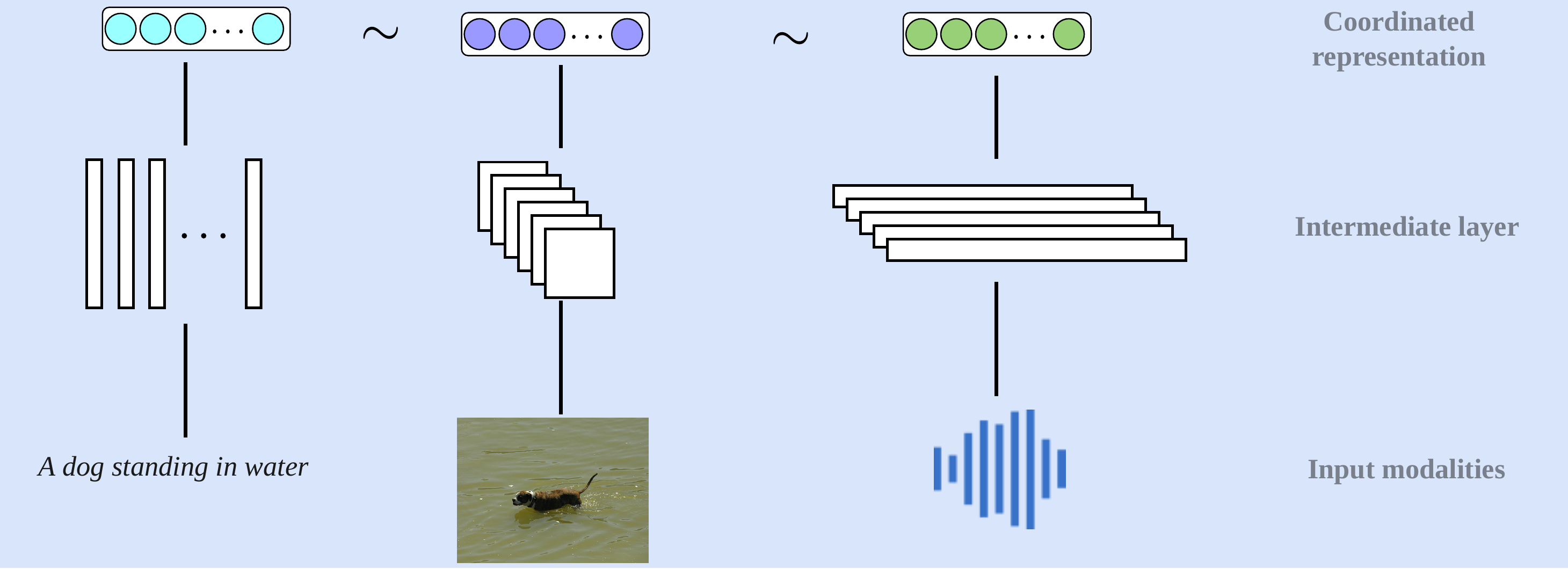}
		\caption{Coordinated representations exist in their own space, but are coordinated
			through a constraint to enforce similarity.}.
		\label{fig:coordinated_gen_rep}
	\end{figure}

	\begin{figure}[!b]
	\centering
	\includegraphics[width=0.99\columnwidth]{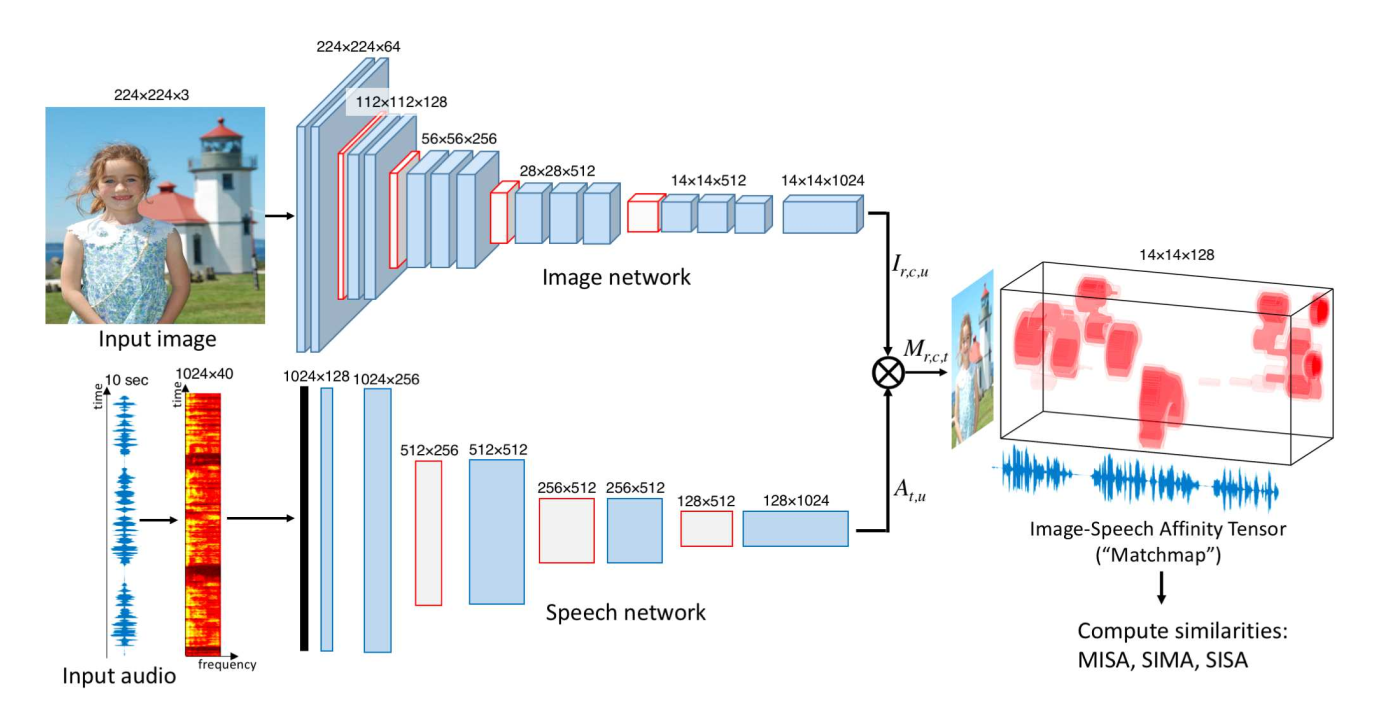}
	\caption{The audio-visual model architecture of~\cite{harwath2018a}. CNN layers shown in blue, pooling
		layers shown in red, and BatchNorm layer shown in black. Each CNN layer is followed by
		a ReLU. An example spectrogram input of approx. 10 seconds (1024 frames) is shown to
		illustrate the pooling ratios. An example output (right), displaying spatial and temporal
		similarity between pairs of image and speech features.
	}
	\label{fig:coordinated_rep}
\end{figure}
	Instead of projecting the modalities into a joint space, like the approach reviewed in Section~\ref{subsec:joint_representation_learning}, in the coordinated representation, separate representations are learned for each modality but are coordinated through a constraint to enforce similarity between the representations~\cite{baltruvsaitis2018}. Figure~\ref{fig:coordinated_gen_rep} presents an illustration of the coordinated representation approach where a separate representation exists for a still image, its written caption, as well as its spoken caption, but the representations are coordinated through a similarity function. The earliest studies that used the coordinated representation are based on the image and written text modalities. Weston~\etal~\cite{weston2010, weston2011} conducted one of the earliest studies based on such representation by constructing a feature space using a simple linear map and coordinating the space---using the weighted approximate-rank pairwise loss---such that matching image and textual caption representation have a higher inner product between them than non-matching ones. 
	
Instead of a linear map, more recent work has employed neural networks. Frome~\etal~\cite{frome2013} use a deep neural network to learn a visual-semantic embedding space and coordinate the space with the hinge rank loss; in this coordinated space, the written word \textit{car} and an image of \textit{cars} are closer to each other than the written word \textit{car} and the image of an \textit{insect}. Kiros~\etal~\cite{kiros2014} extended this to sentence and image coordinated representations by using an LSTM model to construct the feature space and a pairwise ranking loss to coordinate the space. Socher~\etal~\cite{socher2013} tackle the same task, but injected compositional semantics into the sentence embeddings using a decision tree RNN model in order to retrieve images that are described by written sentences.
	
	A similar approach has now been extended beyond still images and written captions. Pan~\etal~\cite{pan2016} proposed a model similar to Socher~\etal's using videos instead of images. Xu~\etal~\cite{xu2015b} also constructed a coordinated space between videos and sentences; the representation was then used for the task of cross-modal retrieval and video description. More recently, Harwath~\etal~\cite{harwath2018a} extended this approach using VGS dataset with still images and their spoken captions (see Figure~\ref{fig:coordinated_rep}).  
	The authors constructed a coordinated space in which the relevant regions of the image can be highlighted as they are being described in the speech. 
	Their model is primarily trained to perform semantic retrieval at the whole-image and whole-caption level. But it also exhibits the ability to detect and localise both visual objects and spoken words without any additional training. What differentiates this model of Harwath~\etal from the approach reviewed in Section~\ref{subsec:joint_representation_learning} is that instead of mapping entire images and spoken utterances to fixed points in an embedding space, they learn representations that are distributed spatially and temporally, enabling their models to directly co-localise within both modalities. 
	In this dissertation we do not use the coordinated representation to capture the complementary information between the modalities---we employ a different approach which we review next. Furthermore, while Harwath~\etal's approach could localise some words when presented with a corresponding image, we consider keyword localisation when a model is presented with a written keyword---a different task.
	
	\subsection{Cross-modal distillation}
	\label{subsec:cross-modal-distillation}
		 \begin{figure}[!t]
		\centering
		\includegraphics[width=0.55\columnwidth]{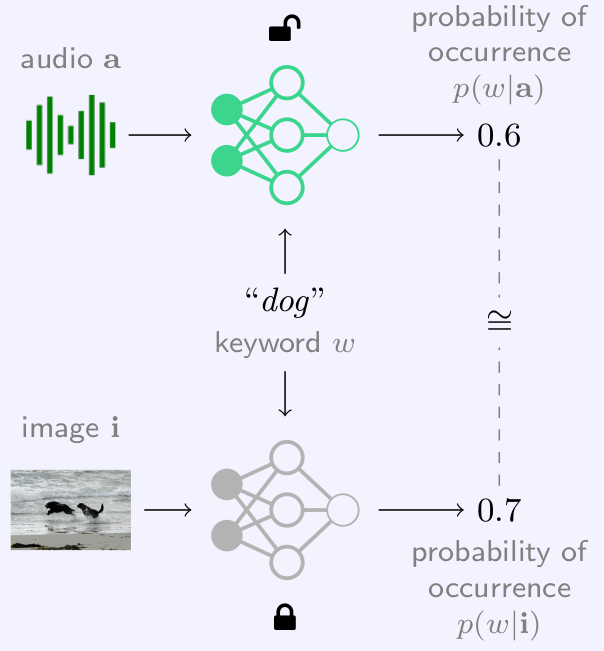}
		\caption{Structure of offline cross-modal distillation. The training of the teacher model (image model; bottom) is detached from the training of the student model (audio model; top).
		}
		\label{fig:cross_modal_distill}
	\end{figure}
In Sections~\ref{subsec:joint_representation_learning} and~\ref{subsec:coordinated_representation_learning} we reviewed two different approaches to multimodal learning. In this section, we discuss and review the approach most relevant to this dissertation. 
While the approaches above define either a joint multimodal space or a coordinated space in which the multimodal system are trained in an end-to-end fashion, the approach to multimodal learning adopted in this dissertation involves extracting complementary knowledge from one modality (ideally a well-resourced modality) with a system (ideally a well-established  ``teacher" model), and using this knowledge to guide the learning of a (``student") model trained on another modality (ideally an under-resourced modality). Figure~\ref{fig:cross_modal_distill} illustrates the specific approach used in this dissertation, originally introduced by Kamper~\etal~\cite{kamper2017a} within the image-speech context (we will give more details about~\cite{kamper2017a} in Section~\ref{sec:visually_grounded_speech_models}). In this approach, the image model (bottom-middle, with a \faLock~symbol, indicating offline training) serves as the teacher and the audio model (top-middle, with the \faUnlock~symbol) serves as the student.

	By having a teacher network to supervise a student network, this methodology can be regarded as a knowledge distillation technique: a method for transferring knowledge from a larger deep neural network onto a small network~\cite{hinton2015,ba2013}. Knowledge distillation~\cite{gou2021} was originally motivated by the need to tackle the challenge of deploying deep learning models on devices with limited resources, such as mobile phones and embedded systems. Figure~\ref{fig:knowledge_distillation} presents an illustration of a teacher-student network for knowledge distillation---with the \textit{Teacher Model} on the left of the figure depicting a deeply-layered neural network which is the source of the \textit{Knowledge} distilled into a shallow \textit{Student Model} on the right. The original methods used large networks to train smaller ones in unimodal settings, i.e.\ these approaches were used for model compression. A comprehensive review of these methods is outside the scope of this dissertation but a recent survey is provided in~\cite{gou2021}.
	
	Compared to the earlier knowledge distillation methods, we share similarities in how knowledge is transferred (via the logits produced by the teacher) and by the fact that the teacher is fixed (offline), but we differ in that our method operates across modalities.
	This idea falls under the category of cross-modal distillation: designed to aid the modelling of a (resource-poor) modality by exploiting knowledge from another (resource-rich) modality~\cite{baltruvsaitis2018}. 
	
	Cross-modal distillation has been previously applied to perform diverse tasks. 
	Gupta~\etal~\cite{gupta2016} proposed a technique that obtains supervisory signals with a teacher model trained on a well labelled (RGB) image modality and used the supervisory signals as labels for unlabelled depth and optical flow images. In particular, the authors trained a convolutional neural network model on these unlabelled depth and optical flow images to reproduce the mid-level semantic representations learned from the well labelled (RGB) image modality. Both modalites are parallel.
	Zhao~\etal~\cite{zhao2018} employed a vision model to provide cross-modal supervision for a deep neural network that parses radio signals to estimate 2D human poses. In particular, during training the system uses synchronised wireless and visual inputs, extracts pose information from the visual modality, and uses it to guide the training process. Once trained, the network uses only the wireless signal for pose estimation.
	By leveraging the natural synchronisation between vision and sound in unlabelled videos, Aytar~\etal~\cite{aytar2016} proposed a student-teacher training procedure which transfers discriminative visual
	knowledge from well established visual recognition models into the sound modality for the purpose of learning natural sound representations.
	Similar to~\cite{aytar2016}, Albanie~\etal~\cite{albanie2018} trained a convolutional neural network for speech emotion recognition by distilling the knowledge of a pretrained facial emotion recognition network using unlabelled video as a bridge.
	Thoker and Gall~\cite{thoker2019} investigated how an action recognition model trained on RGB videos can be adapted to a skeleton-based human action recognition model. Specifically, the authors extracted the knowledge of a teacher network---trained on RGB videos---and transferred the knowledge to a small ensemble of student networks which takes as input skeleton sequence. Each RGB video is paired with each skeleton sequence.
	To improve the performance of the human action recognition model, Garcia~\etal~\cite{garcia2018} introduced an additional modality---depth images---to perform cross-modality distillation. At test time, the trained model receives RGB video stream and generates hallucination stream.
	 As an attempt to help the student model to capture important information from the teacher network, Tian~\etal~\cite{tian2019} introduced a contrastive loss to transfer structural relationship across different modalities. The authors demonstrated that a new contrastive objective loss achieves competitive performance on knowledge
	 transfer tasks in several knowledge distillation settings, including single model compression, ensemble distillation, and cross-modal transfer.
	 \begin{figure}[t]
	 	\centering
	 	\includegraphics[width=0.99\columnwidth]{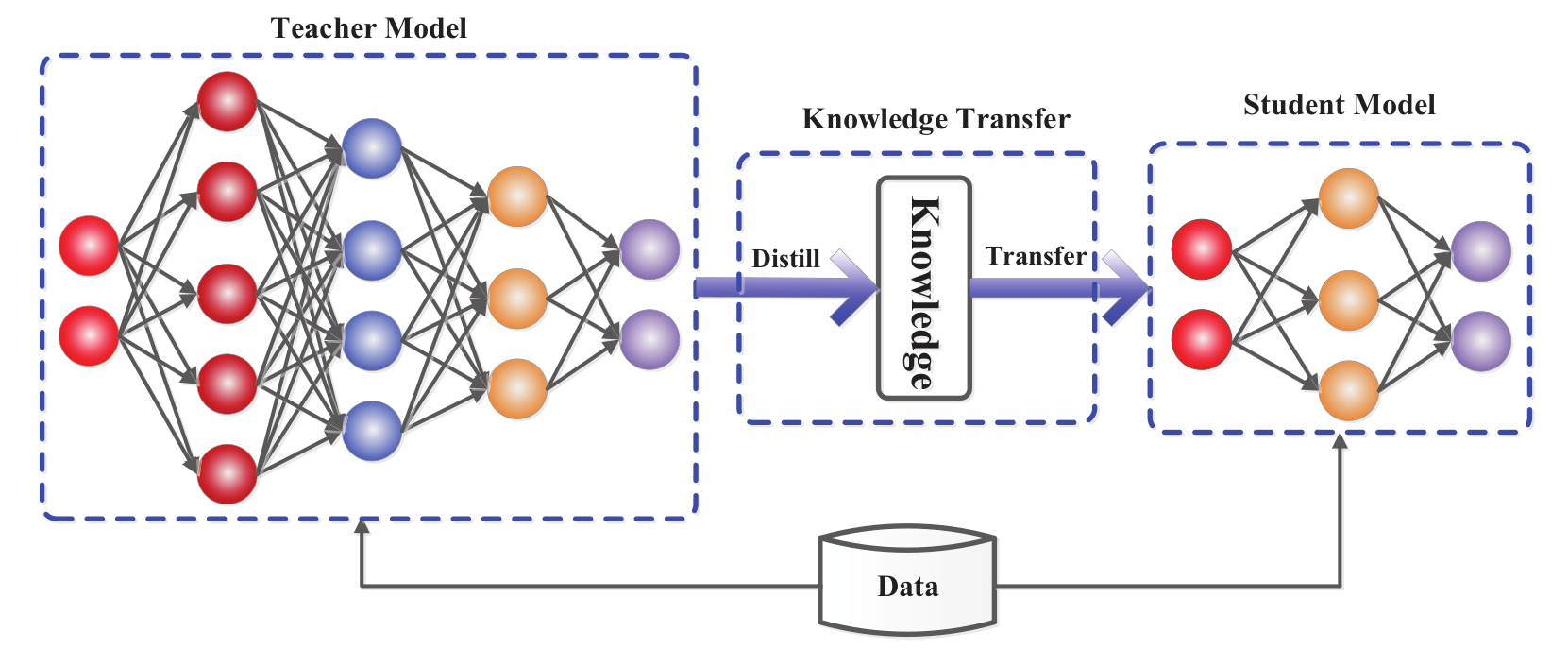}
	 	\caption{A teacher--student network for knowledge distillation~\cite{gou2021}.
	 	}
	 	\label{fig:knowledge_distillation}
	 \end{figure}

	Similar to the cross-modal distillation approach employed in this dissertation, all these studies leverage a visual network for supervision, and do not require any annotated data. Instead they all used pairs of sequences of both modalities as supervision. Furthermore, the helper (teacher) modality are used only during model training and is not used during test time.
	Other studies have also shown that the teacher--student roles can be reversed. Owens \etal \cite{owens2016b,owens2018} used sound information to supervise the visual network, and more generally, Alwassel~\etal~\cite{alwassel2020}, which learn two networks (audio and visual) in tandem on the pseudo-labels obtained by clustering the features from the other modality. The representations learnt by these methods were successfully transferred to tasks such as sound classification~\cite{aytar2016,alwassel2020}, video classification~\cite{alwassel2020}, and object or scene recognition~\cite{owens2018}.
	
	In this dissertation, our downstream tasks are related to speech, specifically keyword detection and keyword localisation. As a starting point, we use the models of Kamper~\etal~\cite{kamper2017a, kamper2019b} whose cross-modal distillation approach is the most relevant to the studies conducted throughout this dissertation. While we described some aspects of the work of Kamper~\etal~\cite{kamper2017a} here to show how it can be regarded as a knowledge distillation approach, we only give a complete overview of this methodology with the full details in Section~\ref{sec:visually_grounded_speech_models}.

\section{Interpretability and analysis of VGS systems} 
\label{sec:analysis_vgs_systems}
Apart from studies looking into finding optimal ways to combine representations from multiple modalities, several studies have also been done on analysing and interpreting the representations. In this section, we  will review such studies, particularly the ones relating to VGS models.

Harwath and Glass~\cite{harwath2019a} investigate how interpretable sub-word speech units emerge within a convolutional-based VGS system trained to map raw speech waveforms to semantically related images. Gelderloos and \chrupala~\cite{chrupala2016} show that their VGS model captures linguistic information in a sequence of phonemes hierarchically, with the lower layer exhibiting more sensitivity to form and the higher layer exhibiting more sensitivity to meaning. Drexler and Glass~\cite{drexler2017} show that features generated with a VGS model contain more discriminative linguistic information and are less speaker-dependent than traditional speech features. Havard~\etal~\cite{havard2019b} show that VGS model does not rely on a global acoustic pattern to activate accurate word representation, but only require access to the first phone of the target word. They also found that only a subset of all speech frames play a crucial role in the representation of a given word. Scholten~\etal~\cite{scholten2021} investigate whether VGS models can be used to recognise words by embedding isolated words and using them to retrieve images of their visual referents. 
Havard~\etal~\cite{havard2019a} show that attention-based visually grounded speech models trained on English and Japanese captions learn to locate nouns in an utterance using images as a guide. 
Havard~\etal~\cite{havard2019b} and Scholten~\etal~\cite{scholten2021} investigate how well a visually grounded speech model learned a mapping from words to their visual referents.

Kamper~\etal~\cite{kamper2019b} investigate how a model that learns from images and their (untranscribed) spoken captions captures aspects of semantics in speech. In Chapter~\ref{chap:methods-for-keyword-localisation} we qualitatively show that some of the incorrect keyword localisations that the VGS models make are due to semantic confusions, for instance locating the word \textit{backstroke} for the query keyword \textit{swimming}. We further show in Chapter~\ref{chap:cross_lingual_keyword_localisation} that such semantic confusions hold even in a cross-lingual setting where VGS models are trained to localise English keyword in \yoruba speech. 

In Chapter~\ref{chap:methods-for-keyword-localisation}, we provide additional task-based analyses of VGS models: we perform detailed experiments revealing the limits of the VGS detection step for localisation; the impact of the VGS architecture on the tasks; and the performance of the model on individual keywords. In Chapter~\ref{chap:cross_lingual_keyword_localisation} we investigate what impact using more data will have on the VGS model, and how initialising the training of a cross-lingual VGS model from a pretrained model impacts the performance of the model.
		
\section{Visually grounded (coarse) speech recognition}
\label{sec:visually_grounded_speech_models}
\begin{figure}[t]
	\centering
	\includegraphics[width=0.99\columnwidth]{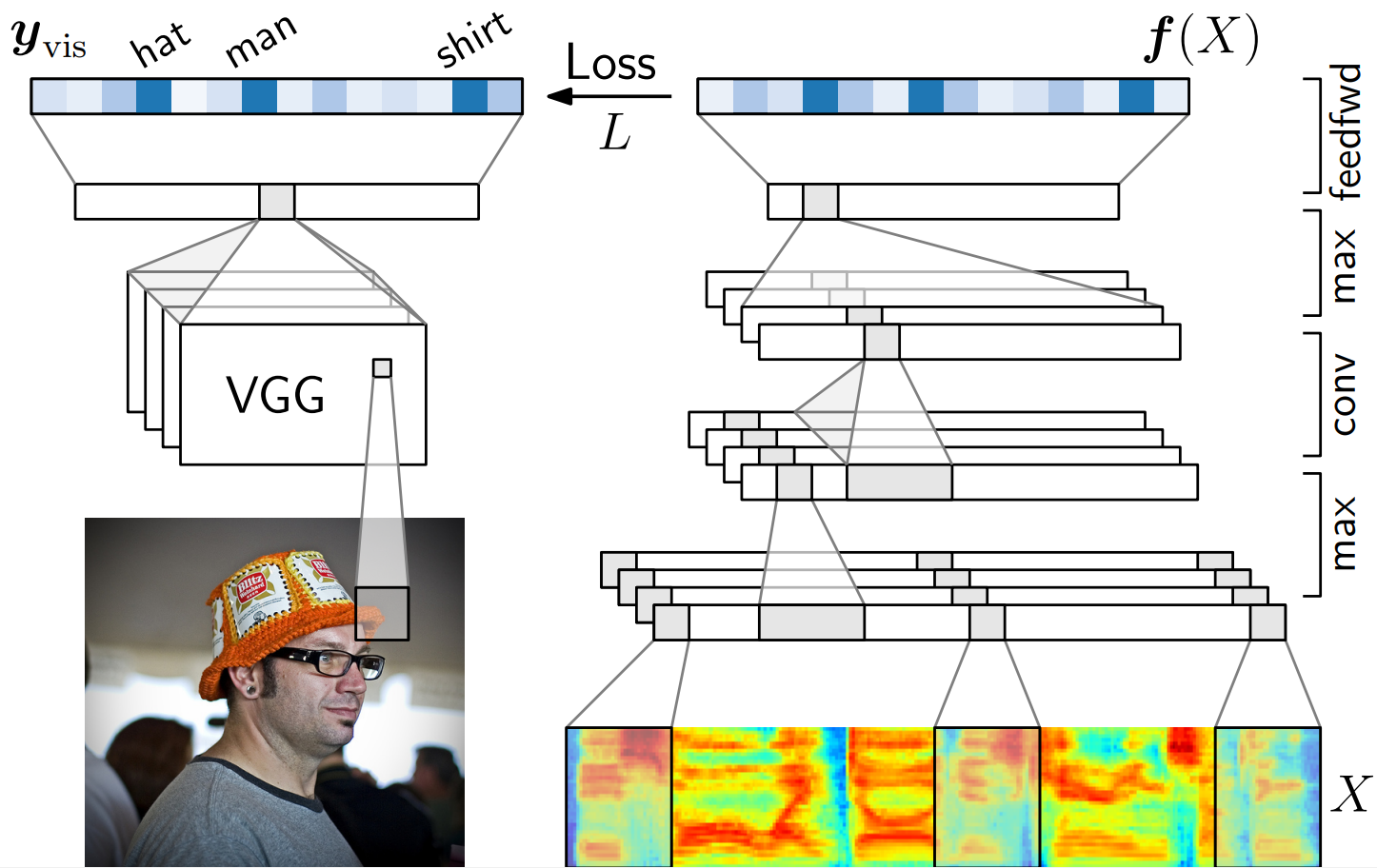}
	\caption{A multi-label image classifier (left) used to produce probability targets for training a speech system (right)~\cite{kamper2017a}.
	}
	\label{fig:vgs_kamper}
\end{figure}
In Section~\ref{sec:methodologies_multimodal_learning} we reviewed, in a more general fashion, studies relating to multimodal learning. Here we specifically review the VGS methodology that is used throughout the dissertation, proposed by Kamper~\etal~\cite{kamper2017a}. This approach has already been briefly introduced in Section~\ref{subsec:cross-modal-distillation}, because this model is a knowledge distillation approach. By using this approach, a VGS model is trained on still images and their spoken captions to do a coarse form of speech recogntion where given keyword can be detected in a spoken utterance.

The studies reviewed in Section~\ref{subsec:joint_representation_learning}, illustrated in Figure~\ref{fig:joint_gen_rep}, and Section~\ref{subsec:coordinated_representation_learning}  illustrated in Figure~\ref{fig:coordinated_gen_rep}, define a joint image-speech space and models that can move between spaces, respectively. While these studies are useful in themselves, they do not provide a mapping of speech to textual labels.
This was addressed by Kamper~\etal~\cite{kamper2017a, kamper2019b}.
In their pioneering work, which is the first application of the offline cross-modal distillation scheme (see Section~\ref{subsec:cross-modal-distillation}) within the VGS context, Kamper~\etal~\cite{kamper2017a} followed a two-stage training process: (i) the large teacher model (the visual model) is first trained on a set of well-labelled still images; and (ii) the teacher model is then used to extract the knowledge in the form of logits, which are then used to guide the training of the student model (the audio model) during distillation. Their architecture, illustrated in Figure~\ref{fig:vgs_kamper} involves an image branch where the well-established VGG-16 network plays the role of a teacher, and the speech branch where a convolutional neural network with a different structure played the role of a student. (The key distinction between offline and online distillation is that in online distillation, both the teacher model and the student model are trained end-to-end, and are updated simultaneously. But in offline distillation, the training of the teacher model is detached from the training of the student model, as illustrated in Figure~\ref{fig:cross_modal_distill}.)

Throughout this dissertation, we use this specific model~\cite{kamper2017a} as a starting point.
However, the speech input in our case are Mel-frequency cepstrum (MFCCs) and not spectograms as depicted in Figure~\ref{fig:vgs_kamper}.
Concretely, to detect whether a written keyword occurs in an utterance, the authors  used a pretrained visual tagger (teacher) that generates soft labels for the training images, which then serve as labels for a convolutional neural network (student) that maps speech to soft unordered word targets. At test time, the speech model is then able to map an input utterance to a set of textual words that are likely to appear in the utterance. The key ingredient is the availability of a pretrained image tagger, which is used to supervise the speech model.
More precisely, given an image, the visual tagger generates soft labels for a fixed vocabulary of visual categories,
which the speech model tries to mimic on the corresponding spoken utterance. The numerous large-scale labelled image datasets~\cite{imagenet2009,xiao2010,lin2014,krishna2017a,kuznetsova2020} and performant vision models~\cite{he2016,tan2019,dosovitskiy2021,tolstikhin2021} ensure the availability of the pretrained visual component. Pasad {\it et al.}~\cite{pasad2019} showed that this approach is even complementary in settings where (limited) text labels are available.

Although these studies have demonstrated that keywords can be detected, the question remains of whether these VGS models can also locate query keywords. In this dissertation, we extend this methodology to produce VGS models which are also capable of localising query keywords in speech. We apply these models both in an artificial low-resource linguistic environment and in a real low-resource environment.

\section{Multimodal paradigms closely related to VGS modelling}
\label{sec:multimodal_paradigms}
In this section we briefly review three multimodal learning paradigms that specifically inspired the recent line of VGS studies. Some of these paradigms preceded many of the studies on VGS modelling reviewed above.
\subsection{Audio-visual (fine-grained) speech recognition}
\label{sec:avsr}
While the studies reviewed in Section~\ref{sec:visually_grounded_speech_models} focussed on coarse speech recognition, which we defined as the task of sparsely recognising keywords in speech, the studies reviewed in this section involves fine-grained recognition of speech. 
\begin{figure}[t]
	\centering
	\includegraphics[width=0.99\columnwidth]{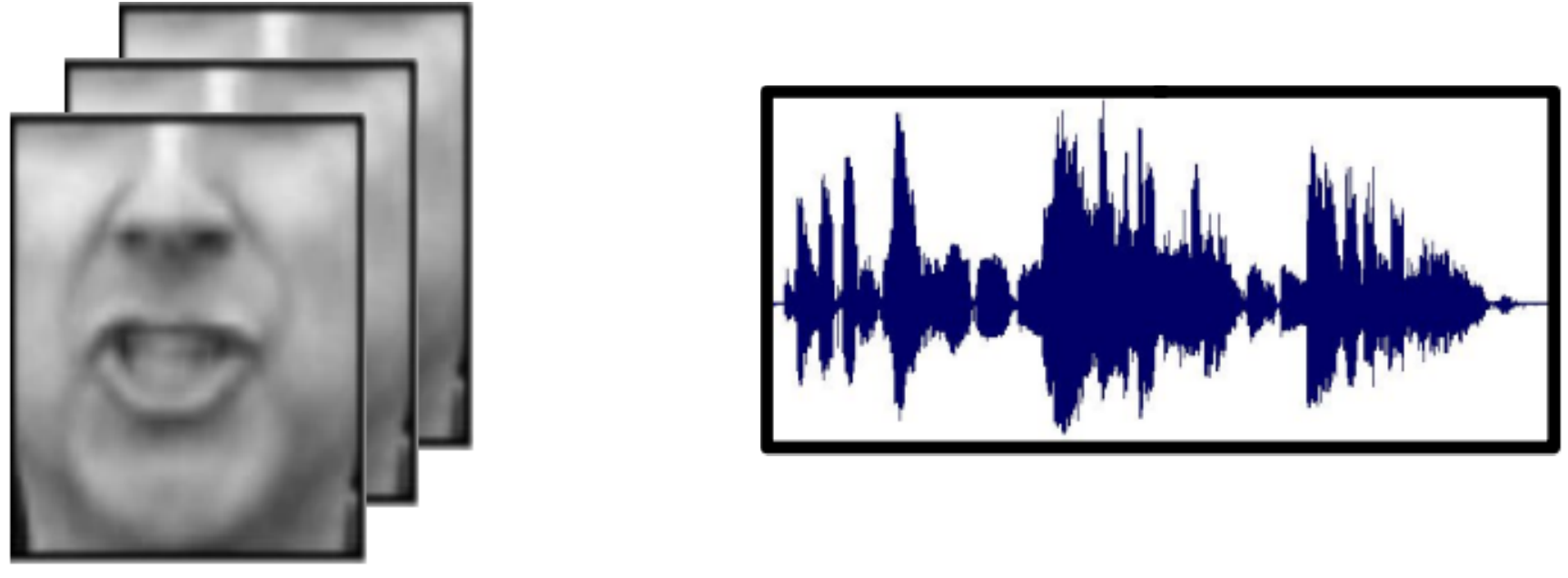}
	\caption{A sample training item containing a sequence of mouth regions of a speaker (left), and the corresponding utterance (right) produced by the speaker~\cite{afouras2018}.
	}
	\label{fig:lip_speech}
\end{figure}
The goal of AVSR is similar to that of automatic speech recognition (ASR)---which is to train a speech recogniser to automatically map utterances to transcriptions---but differs from ASR with the use of an additional modality (visual signal) during model training, i.e.\ speech occur in the context of a visual image, typically of the face of someone speaking, as illustrated in Figure~\ref{fig:lip_speech}. AVSR systems specifically learn from speech audio, textual transcriptions, and visual context associated with the speech. This approach to training a speech recogniser stems from the observation that most verbal communication occurs in contexts where the listener can see and hear the speaker~\cite{mcgurk1976}.
Lip-reading visible facial speech articulations to enhance speech recognition is frequent among humans, especially when the acoustic signal is degraded by noise or hearing impairment~\cite{yuhas1989}. The benefit of using visual information for processing speech in noisy conditions has already been noted as early as 1954~\cite{sumby1954}. Subsequently, a study of the interaction between hearing and vision during speech perception was performed in~\cite{mcgurk1976}, demonstrating the integration of audio and visual stimuli in recognising speech. A line of follow-up studies~\cite{petajan1984, petajan1988, yuhas1989} has been done on combining visual information with information from speech to improve ASR. The central questions most of these studies seek to answer are: (i) can the speech information conveyed by visual speech signals be extracted automatically? (ii) how can this information be combined with information from the acoustic signal to improve ASR?~\cite{yuhas1989}.  More related to this dissertation is the latter question of finding ways to combine visual information with speech information to facilitate downstream tasks in a linguistic environment where adequately transcribed speech data is unavailable. 

Many recent studies use neural network techniques to learn and combine representation from the text, audio and visual modalities. Ngiam~\etal~\cite{ngiam2011} performed audio-visual speech classification on speech audio coupled with videos of the lips, demonstrating that better features for one modality can be learned if multiple modalities are present at learning time. Huang and Kingsbury~\cite{huang2013} use deep belief networks to learn noise robust speech recognition from audio and visual features. On a continuously spoken digit recognition task, they show that their models performed better in terms of the word error rate. Mroueh~\etal~\cite{mroueh2015} presents a multimodal learning method based on a deep neural network architecture for fusing speech and visual information for AVSR. The authors showed that the AVSR system outperformed an audio network---trained without the visual component---on a phone classification task. Afouras~\etal~\cite{afouras2018} investigate how lip reading complements audio speech recognition when the audio signal is noisy. Petridis~\etal~\cite{petridis2018} present an end-to-end audio-visual model based on residual networks and bi-directional gated recurrent units for within-context word recognition on a large publicly available dataset. Their model consists of two streams, one for each modality, which extract features directly from mouth regions and raw audio waveforms. Ma~\etal~\cite{ma2021} present a hybrid connectionist temporal classification and attention model based on a ResNet-18 and convolution-augmented transformer (Conformer) for learning to extract features directly from raw pixels and an audio waveform. In all of these AVSR systems, the labels of the visual and audio modalities are available to guide the training of the system. 

An important distinction between the datasets that we used throughout this dissertation and the type of datasets used for AVSR is that the former consists of still images and their spoken captions while the latter is typically images or videos of faces of some speaking and the speech they produce. Furthermore, our goal is not ASR but detecting, spotting and localising pre-specified keyword types in an utterance. Prajwal~\etal~\cite{prajwal2021} considered the task of spoken keyword spotting in silent video, but we consider written keyword spotting in audio utterances.
\subsection{Learning from images and text}
\label{sec:learning-from-images-and-text}
Another multimodal learning paradigm that predates VGS involves learning from images and text.
Supported by the claim that much of human infants’ language acquisition is rooted in the visual world around them~\cite{yu2007, thiessen2010, Cunillera2010}, a lot of earlier work considered training multimodal systems on paired images and human-generated written captions.
Efforts in this direction showed that such systems can perform image captioning by describing a visual scene with a string of keywords~\cite{young2014, sharma2018}, and even link specific keywords in a caption to specific image regions~\cite{liu2017, lu2018, Rohrbach2018, selvaraju2019}.

In linking keywords in a caption to image regions, two possible paradigms are considered.
One involves sparsely connecting keywords to image segments where the multimodal system prioritises linking keywords belonging to a specific part of speech, for example, a noun, to specific image regions~\cite{plummer2015}.
The other involves densely connecting keywords in the caption to specific image regions regardless of the part of speech of each keyword, for example, ``holding" in ``a woman holding a balloon" should be linked to the image region where this action occurs~\cite{ponttuset2020}.
However, this approach requires multimodal image annotations in which alignment information between each keyword in the caption and the specific image region where the keyword occurs is available during training.

Although our work is related to these studies, we focus on a setting where written captions are unavailable during training: only images and their spoken captions are available.
Moreover, we do not have alignment information during training as in~\cite{ponttuset2020}.
\subsection{Keyword search in sign language}
\label{subsec:keyword_search_sign_language}
\begin{figure}[t]
	\centering
	\includegraphics[width=0.85\columnwidth]{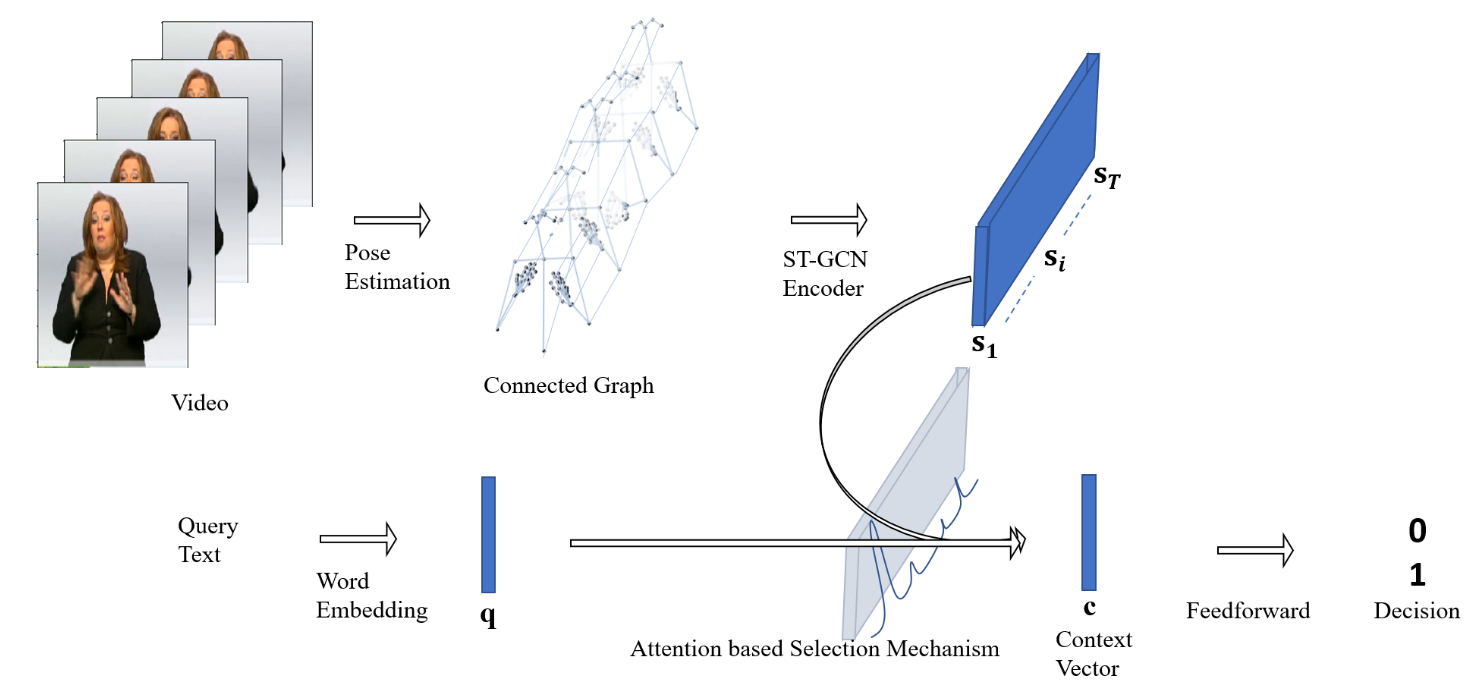}
	\caption{The attention-based spatial temporal graph convolution networks used in~\cite{tamer2020} for keyword search in sign language. Our attention-based approach (Section~\ref{ssec:attention}) is heavily inspired by this work.
	}
	\label{fig:_sign_language}
\end{figure}
Another multimodal learning paradigm related to VGS modelling involves building a system that can search for a written query in sign language~\cite{tamer2020, varol2022}. To extend the technology to sign language, Tamer and Saraçlar~\cite{tamer2020} proposed a keyword search system for sign language. As illustrated in Figure~\ref{fig:_sign_language}, the authors first extracted body and hand joints from the frames of a sign language sentence and represent it with a unified spatio-temporal graph of skeleton joints. They then train a graph-convolutional-network encoder, query embedding, and selection mechanism together in a weakly supervised end-to-end fashion.
This study is related to this dissertation in two ways. Firstly, sign language can be regarded as a low-resource language (the data available with sign language labels, i.e.\ glosses, are limited~\cite{tamer2020}). In this dissertation we also consider a low-resource scenario. Secondly, we use a similar type of attention mechanism to perform keyword localisation as the one proposed by Tamer and Saraçlar.
However, in our case, we incorporate attention in a convolutional neural network to localise keywords in speech, and train the model with visual supervision instead of explicit labels for whether a keyword is present in an utterance or not. Full details about this approach are given in Section~\ref{ssec:attention}.
\section{Keyword spotting and localisation} 
\label{sec:speech_spotting_localisation}
In this section we review studies relating to the main tasks that we attempt to do with our VGS approach. We start by reviewing keyword spotting studies involving models trained without ASR or visual supervision (Section~\ref{subsec:keyword_spotting}). We then review studies relating to keyword localisation in speech (Section~\ref{subsec:keyword_localisation_speech_review}). Finally we review studies relating to cross-lingual learning (Section~\ref{subsec:cross_lingual_learning_review}), which is the main focus of Chapter~\ref{chap:cross_lingual_keyword_localisation}.
\subsection{Keyword spotting without ASR or visual supervision}
\label{subsec:keyword_spotting}
Conventionally, keyword spotting is performed using an ASR system, which requires a massive amount of transcribed audio speech. These resources are expensive to collect for under-resourced language, and several studies seeking alternative means of enabling speech technology for resourced-constrained settings---without relying on conventional ASR---have been conducted. 
Here we will focus on reviewing such studies, specifically the ones that involve learning speech representation and keyword spotting systems without visual supervision. This is distinct from the focus of this dissertation which includes building keyword spotting systems trained in a multimodal setting involving images and their spoken captions using the cross-modal distillation approach reviewed in Section~\ref{sec:visually_grounded_speech_models}.

Keyword spotting closely relates to keyword detection and spoken term detection. As mentioned in Chapter~\ref{chap:introduction}, the aim of keyword spotting is to find utterances containing either exact matches~\cite{wilpon1990, szoke2005, garcia2006} or  semantic matches~\cite{chelba2008, lee2012, li2013, lee2015} of keywords. 
Bird~\cite{bird2020} demonstrates that spoken term detection can enable sparse transcription of endangered spoken languages under the assumption that a small number of isolated spoken words are available to start the process. Ferrand~\etal~\cite{ferrand2020} further this effort by presenting a spoken term detection system with human expertise to
support speech transcription in almost-zero resource settings. Van der Westhuizen~\etal~\cite{vanderwesthuizen2022} consider feature learning for a computationally efficient method of keyword spotting that can be applied in severely under-resourced settings. The objective is to enable speech technology in places where almost no language resources are available. 

All of these studies use limited amounts of transcribed speech resources: they all rely on a small and easily-compiled set of isolated keywords. In this dissertation, we present VGS models, trained without any textual transcription or any set of isolated keywords, for keyword spotting. We go even further by augmenting these models with the ability to also localise keywords without requiring keyword alignment information during training. Speech recognition has also been attempted in a zero-resource setting~\cite{kamper2015a, kamper2015b, renshaw2015} but this dissertation does not cover such a scenario---in our case we rely on some form of supervisory signals obtained from processing a well-resourced image modality with a well-established off-the-shelf visual tagger.  
\subsection{Keyword localisation in speech}
\label{subsec:keyword_localisation_speech_review}
Keyword localisation involves finding where a given keyword occurs in an utterance. It differs from keyword detection which involves checking whether a given keyword occurs in an utterance without considering the location of the keyword. The task of keyword localisation in speech using VGS model is an incredibly challenging problem since neither textual transcriptions nor alignments are provided at training time~\cite{olaleye2022}. 

Visually grounded keyword localisation has application in the documentation of endangered languages~\cite{ferrand2020}. A linguist
can employ such tools to rapidly locate the speech segments containing a query keyword in a collection of utterances; the
only prerequisite is a collection of image-speech pairs from the target language to train the model.
Recent studies are starting to build speech models that not only care about generating transcriptions for speech but also aligning keywords in speech. Common methods currently in use for aligning keywords in speech are adopted from the vision domain. We discuss some of them below.

\subsubsection{Weakly-supervised localisation}
\label{subsec:weak_supervised_localisation}
Generally, a localisation task can be formulated as a multiple-instance learning problem \cite{amores2013,carbonneau2018}:
samples are bags of instances, which are labelled as positive if at least one instance is positive and negative, otherwise. The goal is to infer the labels of the instances based only on bag-level annotations.
In the computer vision community, the related task of weakly supervised object localisation has gained much attention recently, see \cite{choe2020,zhang2021} for surveys.

We draw inspiration from these methods, but focus on localisation methods that are able to exploit existing cues in already trained models.
Amongst others, we follow \cite{oquab2015,zhou2016,bency2016}, which use the network's activations as a means to localise
and \cite{zeiler2014,bazzani2016}, which occlude parts of the input to identify the relevant regions for classification.
Our work adapts these methods to work on the auditory signal, in particular for the task of keyword localisation. 
\subsection{Cross-lingual learning}
\label{subsec:cross_lingual_learning_review}
In Chapter~\ref{chap:cross_lingual_keyword_localisation} we consider a linguistic environment that is more extreme than the ones reviewed in Section~\ref{sec:visually_grounded_speech_models}. In this extreme setting, there is only a limited amount of untranscribed utterances in an endangered language, and written keywords are only available in a different well-resourced language and not in the endangered language. A number of studies has approached processing speech data in such linguistic environment through cross-lingual learning. 

The main idea of cross-lingual learning involves using data from a well-resourced language to support the processing of a low-resourced language.
Early work in cross-lingual learning adopted
a direct approach of using text transcriptions for training the system instead of visual grounding.
For instance, Sheridan~\etal~\cite{sheridan1997} proposed to cascade automatic speech recognition 
with text-based cross-lingual information retrieval~\cite{oard1998}. 
However, this approach is only possible when transcribed speech is available in the target language to build a recogniser. 
Recent work has proposed models that can translate speech in one language directly to text in another~\cite{duong2016, bansal2017, weiss2017, berard2018}, but these methods also require transcribed data: parallel speech with translated text.
Kamper and Roth~\cite{kamper2018} investigated cross-lingual keyword spotting without the need for transcriptions by using the visual information as the link between the languages (more details in Section~\ref{sec:previous_work_chapter5}).
Our approach shares this core idea, but differs in terms of the languages used and the task tackled.

\subsubsection{Challenges in cross-lingual learning}
As mentioned above, cross-lingual learning involves leveraging a source (often well-resourced) language to process a target (ideally low-resourced) language. Linguistic and cultural disparity across different languages may inject anomalies into cross-lingual models as discussed extensively in~\cite{hershovich2022}. Gao~\cite{gao2005} and Larina~\cite{larina2015} highlight the culture-specific differences between the communicative style in one language and the style in another language; pointing out that an expression that is considered polite in one language can be considered offensive in another language. As an example, German native speakers tend to use a high level of directness in communication in the German language which would be considered offensive in English language as discussed by House and Kasper~\cite{house2011}. The expression of emotion also varies across cultures as discussed in~\cite{ryder2008, kirmayer2001, hareli2015, loveys2018}.
We will revisit this discussion in Chapter~\ref{chap:cross_lingual_keyword_localisation}, where we describe how the cross-linguistic and cultural variations impact our cross-lingual dataset and cross-lingual speech-keyword retrieval tasks.
\section{Summary of related work}
In this chapter, we reviewed topics related to existing datasets developed and curated to support multimodal learning. We then reviewed different methods for exploiting complementary information from multiple modalities. Furthermore, we reviewed studies that focussed on analysing multimodal systems, focussing on visually grounded speech models (VGS)---the type of multimodal system that we consider throughout this dissertation.
We zoomed in on VGS to review studies most relevant to this dissertation. 

In this chapter, we also reviewed other types of multimodal systems including audio visual speech recognition systems and systems that learn from images and text. We then reviewed studies relating to the main tasks conducted in this dissertation: we started by reviewing keyword spotting studies involving models trained without ASR and visual supervision, we then reviewed studies relating to keyword localisation in speech, and finally we reviewed studies relating to cross-lingual learning.

%% file: vgs.tex
\chapter{Visually grounded keyword detection and spotting}
\label{chap:problem_formulation_setup}

In this chapter we present visually grounded speech (VGS) models for keyword detection and spotting.
These models have been proposed before~\cite{kamper2017a, kamper2019b} but in this dissertation we update them for the use of keyword localisation (which is only considered in the subsequent chapters). In this chapter we also propose new keyword detection and spotting methods using attention which has not been considered before.  As a reminder, our keyword detection system is trained on soft labels extracted from images paired with spoken captions, and at test time, the trained VGS system receives an utterance and a written query to predict the presence of the query keyword in the utterance. In keyword spotting, the VGS system receives a collection of utterances and a written query and outputs utterances containing instances of the query.

The goal of this chapter is to explain the keyword detection models that we will extend in the rest of this dissertation for the task of keyword localisation which is our primary goal. We start by describing the one datasets on which we do a number of our experiments (Section~\ref{sec:dataset}). We provide details about the evaluation procedure in Section~\ref{sec:evaluation-tasks}. In Section~\ref{sec:keyword_detection_spotting_VGS}, we present the methodologies employed, including the VGS architecture and implementation details. In Section~\ref{sec:detection_spotting_results}, we present the VGS model's baseline results on keyword detection and keyword spotting tasks. Furthermore, we present upper bound results which we estimate using a model trained on unordered transcriptions instead of automatically obtained visual tags. 

This chapter lays the groundwork for answering research question 1 (Section~\ref{sec:research_questions}): Is keyword localisation possible with a VGS model? We do not answer the question in this chapter, but we do provide the detection models which we extend in the subsequent chapters.
\begin{tcolorbox}[width=\linewidth, colback=white!95!black, boxrule=0.5pt]
	\small
	\textit{Parts of this chapter were presented in:} \\
	K. Olaleye, D. Oneață, and H. Kamper, ``Keyword localisation in untranscribed speech using visually grounded speech models,'' \textit{Journal of Selected Topics in Signal Processing}, 2022.
\end{tcolorbox}

\begin{table}[b]
	\centering
	\caption{{
			The quantitative characteristics of the Flickr Audio Caption Corpus (FACC). The duration of the entire corpus is approximately 36 hours.
	}}
	\label{tbl:FACC_corpus_characteristics}
	\begin{tabular}{lcccc}
		\toprule
		Characteristics                     & Total & Train & Val & Test \\
		\midrule
		Speakers        & 183 & 183& 176 & 176\\ [3pt]   
		Duration (hours)        & 36.0  &27.0 & 4.5 & 4.5\\   [3pt]  
		Images & 8k & 6k& 1k & 1k \\ [3pt]
		Utterances & 40k & 30k & 5k & 5k\\ [3pt]

		\bottomrule
	\end{tabular}
\end{table}
\section{Dataset}
\label{sec:dataset}
In this section, we describe the main dataset used in this chapter and in Chapter~\ref{chap:methods-for-keyword-localisation}, and for some of the comparative experiments in Chapter~\ref{chap:cross_lingual_keyword_localisation}. Since our task involves three modalities---audio and text (keywords) at evaluation time, audio and vision at training time---we carry out our experiments on arguably the most common multimodal dataset encountered in the literature: Flickr8k and its audio extension, the Flickr Audio Caption Corpus (FACC).
\subsection{Flickr8k}
The Flickr8k dataset~\cite{rashtchian2010,hodosh2015} is a bimodal (image and text) dataset
consisting of 8k images and 40k English captions. Each image is annotated with a textual description by five different people.
The images (and captions) generally depict relations and actions involving people or animals.
Flickr8k was initially used for image captioning and cross-modal retrieval, but was subsequently
extended with translations \cite{elliott2016,li2016}, spoken captions \cite{harwath2015}, and bounding box annotations for the visual concepts \cite{plummer2015}.
We are particularly interested in the spoken version of this dataset, which is described next.

\subsection{Flickr Audio Caption Corpus (FACC)}
\label{subsec:FACC}
\begin{figure}[bt]
	\centering
	\includegraphics[width=0.99\linewidth]{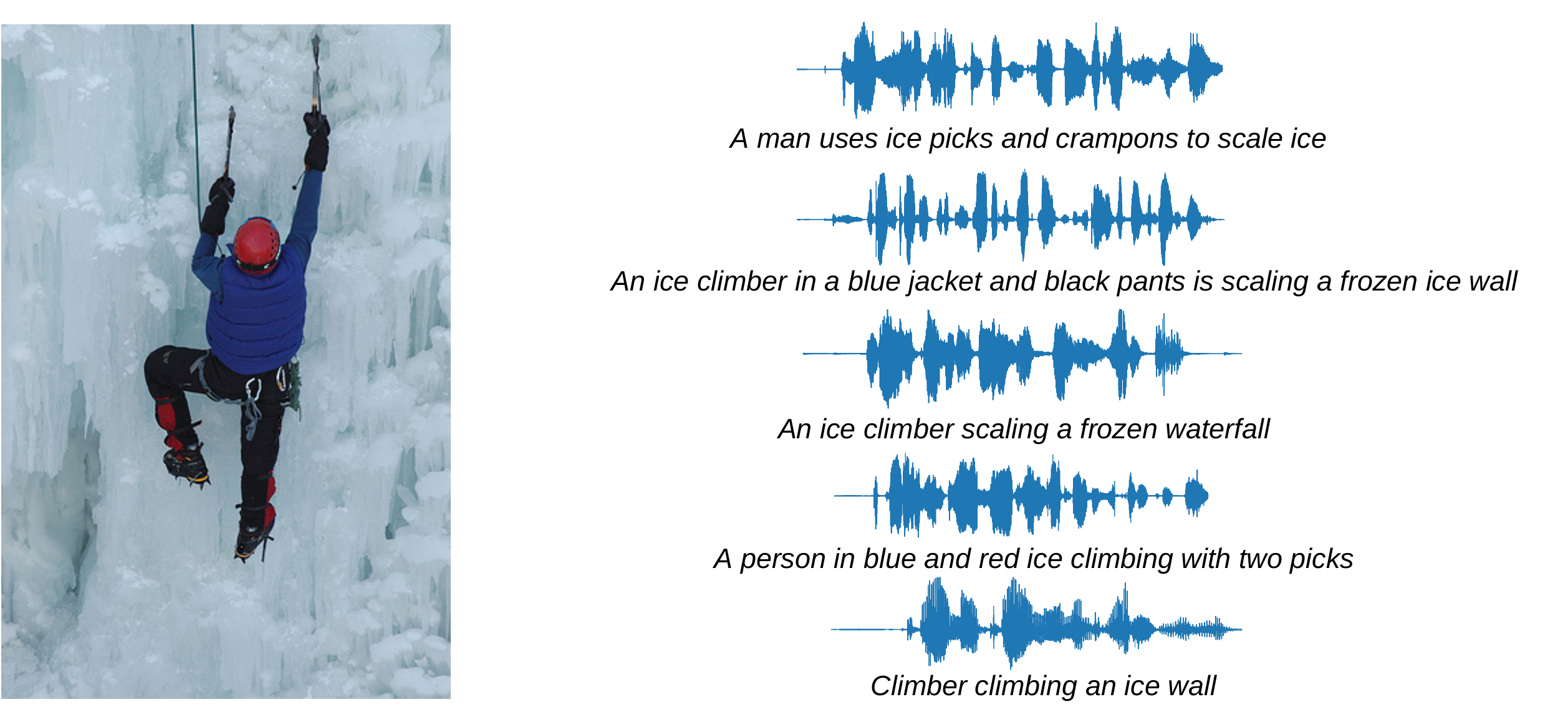}
	\caption{%
		An example from the Flickr Audio Caption Corpus (FACC) of an image with its corresponding spoken captions. The textual captions are not used in any of our models during training, and are given here only for illustrative purposes. We do use the textual transcriptions in the evaluation of our approaches.
	}
	\label{fig:image_speech_text_examples}
\end{figure}
The Flickr Audio Caption Corpus (FACC) was developed by Harwath and Glass~\cite{harwath2015}. They used Amazon's Mechanical Turk to crowdsource approximately 36 hours of non-silent English speech, at 16 kHz, for each of the text captions in the original Flickr8k dataset (above). 
FACC subsequently enabled visually-grounded tasks, such as image--speech alignment \cite{harwath2015}, semantic speech retrieval \cite{kamper2019b}, image-to-speech synthesis \cite{hasegawa2017} and speech-to-image generation \cite{li2020}; these studies are reviewed in Chapter~\ref{chap:related_work}. 

In the rest of this dissertation, we perform experiments on FACC, consisting of five English audio captions for each of the roughly $8$k images.
A split of $30$k, $5$k, and $5$k  utterances are used for training, development, and testing, respectively.
Our data partitioning corresponds with the standard training, validation, and testing splits used in \cite{rashtchian2010,harwath2015,kamper2017a}.
While none of the images or individual utterances overlap between the splits, there is speaker overlap between the sets.

Table~\ref{tbl:FACC_corpus_characteristics} presents the number of images and utterances per split (train, val, and test) in FACC. The table also shows the number of speakers that recorded the speech as well as the duration of the entire dataset. All experiments in this chapter and in Chapter~\ref{chap:methods-for-keyword-localisation} are based on FACC, and we also use the dataset for some of the comparative models in Chapter~\ref{chap:cross_lingual_keyword_localisation}. Figure~\ref{fig:image_speech_text_examples} shows an example image with its corresponding spoken caption and textual caption. The textual captions are never used during training of our VGS models; but we do actually use them to train the idealised models used to estimate upper bounds on our VGS models' performance. These bag-of-word (BoW) labels are considered as a weak form of supervision because they indicate the presence or absence of words without giving their location, order, or number of occurrences (details in Chapter~\ref{subsec:training}). The textual transcriptions are also used in the evaluation of our models.

\begin{table}[!t]
	\caption{A list of the 67 unique keyword types considered in this dissertation. The ``Train count" and ``Val count" columns give the number of occurrence of each keyword type in the training and validation sets, respectively.}
	
	\label{tbl:vocabulary_list}
	\centering
	\begin{tabular}{@{}rlcccrlcc}
		\toprule
		& Keyword       & Train count   & Val count & & & Keyword & Train count & Val count \\
		\midrule
		\ii{1} & air       & 154 & 23 &  & \ii{35} & race        & 58 & 7  \\
		\ii{2} & baby      & 58  & 10 &  & \ii{36} & red         & 378 & 66  \\
		\ii{3} & ball      & 264 & 47 &  & \ii{37} & rides       & 73 & 11 \\
		\ii{4} & beach     & 147 & 26 &  & \ii{38} & riding      & 145 & 21   \\
		\ii{5} & bike      & 122 & 25 &  & \ii{39} & road        & 67 & 9  \\
		\ii{6} & black     & 566 & 86 &  & \ii{40} & rock        & 117 & 10  \\
		\ii{7} & boy       & 536 & 93 &  & \ii{41} & running     & 336 & 57 \\
		\ii{8} & brown     & 393 & 64 &  & \ii{42} & sand        & 58 & 15   \\
		\ii{9} & building  & 70  & 9 &  & \ii{43} & shirt       & 264 & 41  \\
		\ii{10} & camera   & 108 & 15 &  & \ii{44} & sits        & 84 & 15  \\
		\ii{11} & car      & 61  & 12 &  & \ii{45} & sitting     & 202 & 40 \\
		\ii{12} & carrying & 66  & 8 &  & \ii{46} & skateboard  & 73 & 10   \\
		\ii{13} & children & 168 & 33 &  & \ii{47} & small       & 187 & 24  \\
		\ii{14} & climbing & 95  & 11 &  & \ii{48} & smiling     & 65 & 11  \\
		\ii{15} & dirt     & 88  & 16 &  & \ii{49} & snow        & 236 & 32 \\
		\ii{16} & dogs     & 312 & 47 &  & \ii{50} & snowy       & 64 & 14   \\
		\ii{17} & face     & 67  & 8 &  & \ii{51} & soccer      & 84 & 12  \\
		\ii{18} & field    & 191 & 20 &  & \ii{52} & stands      & 126 & 25  \\
		\ii{19} & football & 78  & 11 &  & \ii{53} & stick       & 69 & 14 \\
		\ii{20} & grass    & 238 & 40 &  & \ii{54} & street      & 163 & 22   \\
		\ii{21} & hair     & 62  & 8 &  & \ii{55} & swimming    & 61 & 13  \\
		\ii{22} & hat      & 94  & 17 &  & \ii{56} & tennis      & 65 & 12  \\
		\ii{23} & holding  & 206 & 40 &  & \ii{57} & three       & 188 & 34 \\
		\ii{24} & jacket   & 112 & 14 &  & \ii{58} & top         & 74 & 9   \\
		\ii{25} & jumps    & 149 & 30 &  & \ii{59} & toy         & 90 & 12  \\
		\ii{26} & large    & 180 & 35 &  & \ii{60} & tree        & 58 & 7  \\
		\ii{27} & little   & 265 & 61 &  & \ii{61} & walks       & 85 & 11 \\
		\ii{28} & mountain & 77  & 11 &  & \ii{62} & water       & 390 & 75   \\
		\ii{29} & mouth    & 160 & 29 &  & \ii{63} & wearing     & 468 & 69  \\
		\ii{30} & ocean    & 73  & 16 &  & \ii{64} & white       & 583 & 85  \\
		\ii{31} & orange   & 105 & 13 &  & \ii{65} & women       & 101 & 17 \\
		\ii{32} & park     & 74 & 15 &  & \ii{66} & yellow       & 194 & 20   \\
		\ii{33} & pink     & 93 & 25 &  & \ii{67} & young        & 393 & 72  \\
		\ii{34} & pool     & 104 & 20 &  &  &  &  &   \\
		
		\bottomrule
	\end{tabular}
\end{table}

\subsection{Visual keywords in FACC}
\label{subsec:keyword_types}
Some studies have done keyword detection and spotting on FACC through visual grounding. These studies~\cite{kamper2019b, pasad2019} considered 67 unique keyword types. Table~\ref{tbl:vocabulary_list} shows the keyword types and their number of occurrences in the training and validation sets.
The procedure used to select these keywords are detailed in~\cite{kamper2019b}; it includes a human reviewer agreement step, which reduced the original set from 70 to 67 words. These keywords also have timestamps, which are obtained automatically using forced alignment. In subsequent chapters, we will also consider keyword localisation using this same set of keywords.

\section{Evaluating VGS models for keyword retrieval tasks}
\label{sec:evaluation-tasks}

Throughout the dissertation we consider keyword detection, keyword spotting and keyword localisation, specifically using VGS models. But in this chapter, we will focus strictly on the first two tasks since they lay the groundwork for the keyword localisation methods proposed in Chapters~\ref{chap:methods-for-keyword-localisation} and~\ref{chap:cross_lingual_keyword_localisation}. This section will introduce the keyword detection and keyword spotting tasks. These tasks are typically accomplished using supervised speech-only models (see Section~\ref{subsec:keyword_spotting}), but in the rest of the dissertation we will consider existing VGS models and propose new models for tackling these tasks. While evaluating VGS models for keyword retrieval tasks, it is important to note that throughout this dissertation images are not used anywhere. 

\subsection{Keyword detection}
\label{subsec:keyword_detection_evaluation}
In keyword detection we are given a keyword and an audio utterance (which the model has never seen before), and the task is to predict whether the keyword is present in the utterance.
Given an audio utterance, the VGS model gives a score $\hat{y}_w \in [0, 1]$ for each keyword $w$ that we are considering. To perform a hard keyword detection, we set a threshold $\theta$ and output labels for all $w$ where $\hat{y}_w \ge \theta$. We compare the predicted labels to the true set of words in the transcriptions, and calculate precision ($P$), recall ($R$) and $F_1$ as the average over all the keyword types considered. 

\subsection{Keyword spotting}
\label{subsec:keyword_spotting_evaluation}
In keyword spotting we are given a keyword $w$ and a collection of utterances and we want to rank the utterances according to the probability that the utterances contain the keyword. To do this, we rank the utterances from the largest $\hat{y}_w$ value to the smallest $\hat{y}_w$ value. Using this ranked list, keyword spotting is then evaluated in terms of several metrics:
\begin{itemize}
	\item Precision at 10 ($P@10$) is the average precision of the ten highest-scoring proposals.
	\item Precision at $N$ ($P@N$) is the average precision of the top $N$ proposals, with $N$ the number of true occurrences of the keyword. For example, for the keyword \textit{race} in the FACC validation data, $N$ would be $7$, i.e. we will report the fraction of the top 7 retrieved utterances that contain the keyword \textit{race}.
	\item Equal error rate (EER) is the average error rate at which the false acceptance---the ratio between the number of utterances wrongly categorised as containing the keyword and the total number of utterances actually not containing the keyword---and the false rejection rate---the ratio between the number of utterances wrongly categorised as not containing the keyword and the total number of utterances actually containing the keyword---are equal.
\end{itemize}

\section{Keyword detection and spotting using visual grounding}
\label{sec:keyword_detection_spotting_VGS}
Below we give details about the training and test-time inference of our VGS models for keyword detection and spotting. The methodology followed here has been outlined already in Section~\ref{sec:visually_grounded_speech_models}. Here we formalise this approach. In short, we are given images and spoken captions---both unlabelled---and we pass the images through a visual tagger to generate soft labels for the spoken captions. These labels are used to train a speech network to predict the presence of keywords in held-out utterances. The training and testing procedures are illustrated in Figure~\ref{fig:overview_artificial}.

The main component of our work is a system that, given an audio utterance $\ab$ and a keyword $w$,
estimates the probability of the keyword occurring in that utterance.
The system is implemented in terms of an audio deep neural network $\dnna$ that outputs a vector of predictions $\dnna(\ab) \in \mathbb{R}^\vocab$, each element corresponding to a score for one of the $\vocab$ words in the system vocabulary.
This vector is then passed through the sigmoid function $\sigma$ to ensure that the predictions are in $[0, 1]$ and can be interpreted as probabilities:   

\begin{equation}
	\hat \yb \triangleq p(w|\ab) = \sigma(\dnna(\ab)).
\end{equation}
To obtain the probability for a single keyword $w$, we select the entry from $\hat \yb \in [0, 1]^\vocab$ corresponding to the $w$-th keyword in the vector, i.e., $\hat{y}_w$. We will explain below how we get the ground truth target $\mathbf{y}$. (In short, we either use visual or bag-of-word supervision.)
\subsection{Network structure}
\label{subsec:network_structure}
We assume that the audio network $\dnna$ consists of three components that are sequenced---an audio encoder $\enc$, a pooling layer $\pool$, and a classifier $\clf$:
\begin{equation}
	\dnna = \clf \circ \pool \circ \enc.
\end{equation}
The audio encoder $\enc$ produces a sequence of $\emb$-dimensional embeddings $\left[\hb_1, \dots, \hb_\temp\right] = \Hb \in \mathbb{R}^{\emb \times \temp}$,
which are then aggregated across the temporal dimension (of $\temp$ steps) by the pooling layer.
The result is finally passed through the classifier $\clf$ to obtain a $\vocab$-dimensional vector corresponding to the scores of the keywords. The main reason we adopt this type of structure (encoder, pooling, classifier) stems from the localisation task (which will be discussed at length in Chapter~\ref{chap:methods-for-keyword-localisation}), where the pooling layer plays a central role---some of the methods rely on the pre-pooling activations for localisation. 

\subsection{Training}
\label{subsec:training}
\begin{figure}[t]
	\centering
	\includegraphics[width=0.99\columnwidth]{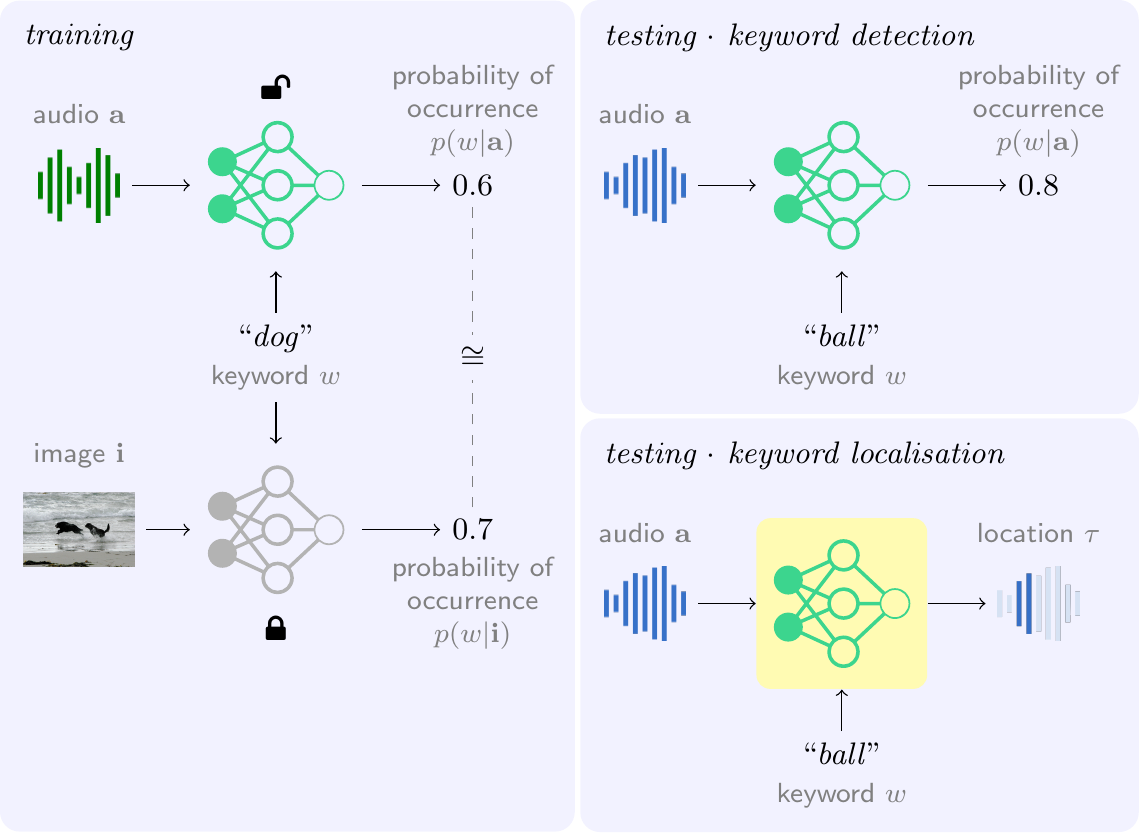}
	\caption{%
			Methodological overview.
			\textit{Left:} We train an audio neural network to predict whether a keyword $w$ occurs in an audio utterance $\ab$
			based on the supervision of a fixed pretrained image tagger ran on a corresponding image $\ib$.
			\textit{Top-right:}
			At test time, the network can be used
			to identify whether a certain keyword occurs in a given utterance (keyword detection).
			\textit{Bottom-right:} It can also be used
			to locate the keyword in the input (keyword localisation), which we consider in the next chapter.
	}
	\label{fig:overview_artificial}
\end{figure}
To train the model, we use \textit{visual supervision}.
Concretely, we assume that:
(1) for each utterance in the training data, there is a corresponding matching image, and
(2)~we have access to a pretrained image tagger to generate probability labels from images over the same vocabulary of $\vocab$ words.
The target outputs 
$\ybv \in [0, 1]^\vocab$ for the audio network will be the soft labels produced by an image tagger: an image deep neural network $\dnni$ (explained in more detail in the subsection directly below).
For the image $\ib$ associated with an utterance $\ab$, we therefore obtain:
\begin{equation}
	\ybv \triangleq p(w|\ib) = \sigma(\dnni(\ib)).
\end{equation}
We optimise the binary cross-entropy between the predictions of the audio network (taking $\ab$ as input) and the predictions from the image network (taking $\ib$ as input):

\begin{equation}
	L = \yv_w \log \hat y_w + (1 - \yv_w) \log (1 - \hat y_w).
	\label{eq:cross_entropy}
\end{equation}
$L$ is averaged over the keywords and audio-image pairs,
and it is optimised with respect to the weights of the audio network,
while the weights of the image network are frozen.
This training procedure is illustrated on the left in Figure~\ref{fig:overview_artificial}.

\subsubsection{Visual tagger}
Our VGS models are trained for keyword detection and spotting using soft text labels obtained by passing each training image through the multi-label visual tagger of~\cite{kamper2019b}. This tagger 
is based on VGG-16~\cite{simonyan2014} and is trained on images disjoint from the data used here, specifically images from MSCOCO~\cite{lin2014}.
This tagger has an output of 1000 image classes, but here we use a subset of the 1000 keywords---containing 67 unique keyword types---which is the same set of keywords used in several other studies~\cite{kamper2019b, pasad2019}. The procedure used to select these keywords are detailed in~\cite{kamper2019b}; it includes a human reviewer agreement step, which reduced the original set from 70 to 67 words. (See Section~\ref{subsec:keyword_types} for more details on this set of keywords).

\subsubsection{Bag-of-words (BoW) supervision}
The ground truth provided by the visual tagger is weak in two senses:
(1) there is no localisation information---the word can appear anywhere in the utterance---and
(2) the labels can possibly be erroneous since the audio description can be different from the visual information present in the image. 
To measure the impact of (2), i.e.\ the effect of using possibly incorrect labels from the visual tagger,
we also experiment with a case where the target is
a
bag-of-words (BoW).
A BoW labelling indicates
the presence or absence of words without giving their location, order, or number of occurrences.
In this scenario we generate the target labels $\yt$ from the text transcriptions $\tb$ of the audio sample, i.e.\ $\yt_w = \llbracket w \in \tb \rrbracket$, 
where $\llbracket - \rrbracket$ denotes the indicator function.
For this idealised setting, we use the same binary cross-entropy loss as in Equation~\ref{eq:cross_entropy}, but with the 
BoW
labels.
\subsection{Testing the VGS models}
\subsubsection*{Keyword detection}
At test time the audio neural network can be used in the same manner as it was trained:
given an audio utterance and a keyword, pass them through the network to obtain the probability of the occurrence of the specified keyword.
Note that no visual information is required at test time.
We can obtain a binary decision regarding the presence of the keyword $w$ by thresholding the probability score $\hat y_w$ with a value $\theta_w$,
that is
$\llbracket \hat y_w \ge \theta_w \rrbracket$.
This is illustrated in the top-right of Figure~\ref{fig:overview_artificial}. The accuracy of this prediction can be evaluated as described in Section~\ref{subsec:keyword_detection_evaluation} above.

\subsubsection{Keyword spotting}
The output $\hat y_w$ can also be used for keyword spotting. In this case the VGS model is given a keyword and a collection of utterances and must rank the utterances in descending order of their $\hat y_w$ value---the utterance with the best $\hat y_w$ value is given the topmost rank followed by the second-best, and in that order. This implies that utterances are ranked according to the probability that the utterances contain the keyword. Similar to keyword detection, no visual information is required. The performance of the VGS models on keyword spotting can be assessed as described in Section~\ref{subsec:keyword_spotting_evaluation}.

\subsection{VGS architectures and implementation details}
\label{subsec:exp_setup}
We define our specific model
architectures in terms of the three components mentioned in Section~\ref{subsec:network_structure}:
the encoder, pooling layers and the classifier.

\subsubsection*{Encoders (Enc)}

We use two types of encoders: CNN and CNN-Pool.
Both are based on convolutional neural networks and have been considered in previous work~\cite{kamper2017a, kamper2019b}. But here we change them slightly by switching from two-dimensional to one-dimensional convolutional neural network, and update them for the use of keyword localisation (which is only considered in the subsequent chapters). 
The main difference between CNN and CNN-Pool is that CNN-Pool uses intermediate max-pooling while CNN doesn't. 
The reason we consider the two alternatives is that in
earlier studies,
CNN-Pool yielded better scores on keyword detection while CNN gave better scores on keyword localisation in some settings that are described in the next chapters.

The CNN-Pool encoder consists of three one-dimensional convolutional layers with ReLU activations.
Intermediate max-pooling over 3 units is applied in the first two layers. 
The first convolutional layer has 64 filters with a width of 9 frames. 
The second layer has 256 filters with a width of 11 units, and the last layer has 1024 filters with a width
of 11. 
The CNN encoder consists of 6 one-dimensional convolutional layers with ReLU activations. 
The first has 96 filters with a kernel width of 9 frames. 
The next four have a width of 11 units with 96 filters. 
The last convolutional layer uses 1000 filters with a width of 11 units.
A padding size of 4 is used in the first convolutional layer and 5 in the remaining layers.

\subsubsection*{Pooling layers (Pool)}
All our models take in variable-duration speech data as input and then make a fixed number of output predictions.
We, therefore, need to summarise the output from the speech encoder into a fixed-dimensional vector.
We specifically compare 
three types of pooling layers:
max pooling over time, log-mean-exp pooling and attention pooling. 

The log-mean-exp pooling, defined as~\cite{palaz2016}:

 \begin{equation}
 	\pool(\Hb) = \sigma\left(\frac{1}{r} \log \mean_{\temp} \exp \left(r \Hb\right)\right)
 	\label{eq:log-mean-exp_chp3}
 \end{equation}
 is a soft pooling method regarded as a trade-off between max pooling and mean pooling depending on the temperature parameter $r$. A large part of the motivation for this pooling method is to improve localisation but we will only discuss this in Section~\ref{subsubsec:score-agg}. 

For the attention pooling, we use the following equation: 
\begin{equation}
	\pool(\Hb)_w = \sum_{t = 1}^{\temp} \alpha_{w, t} \mathbf{h}_t.
\end{equation}
Here $\alpha_{w, t}$ is a keyword-specific weight tied to a specific time point but we will give more details about how we use this weight to predict keyword location in Chapter~\ref{chap:methods-for-keyword-localisation}; for now we are using it to pool the intermediate features $\mathbf{h}_t$ at every time step $t$ to obtain the keyword detection score. This attention-based approach is also heavily inspired by~\cite{tamer2020}, but again we will explain this in detail in Section~\ref{ssec:attention}.

Note that all the different pooling methods we considered are built into existing VGS models~\cite{kamper2017a, kamper2019b}.

\subsubsection*{Classifiers (Clf)}
The classifier consists of a two-layer multi-layer perceptron (MLP) 
with ReLU activations. The first layer produces an output of dimension 4096.
There are differences in output dimensions of the second layer (precise details below): in some cases, the classifier outputs a probability for each of the $\vocab$ words in the output vocabulary (e.g., \ when trained for multi-label classification). Other times, it outputs a single value for whether a specific given keyword is present (e.g., when trained with attention pooling).

\subsubsection*{Architectures}
Based on the components described in Section~\ref{subsec:exp_setup}, we assemble four types of architectures (PSC, CNN-Pool, CNN-Attend and CNN-PoolAttend), summarised in Table~\ref{tbl:architectures_localisation}. We show in Chapter~\ref{chap:methods-for-keyword-localisation} that these architectures can be combined with four distinct localisation methods. In this chapter we evaluate each of these architectures on the keyword detection and keyword spotting tasks.
CNN, as used here, refers to the encoder type with no intermediate max pooling in its stack of convolutional layers (see above). CNN-Pool does include intermediate max pooling. PSC uses log-mean-exp pooling, as given in Equation~\ref{eq:log-mean-exp_chp3}; the reason for this architecture name is that this pooling approach was first proposed by Palaz, Synnaeve, and Collobert~\cite{palaz2016}.

\subsubsection*{Implementation details}
All architectures take as input mel-frequency cepstral coefficients. 
We augment the data using SpecAugment~\cite{park2019}, 
an augmentation policy that consists of warping the speech features, masking blocks of frequency channels, and masking blocks of time steps.
This policy achieves state-of-the-art performance in end-to-end speech recognition~\cite{park2019}. 
All models are implemented in PyTorch and use Adam optimisation~\cite{kingma2015} with a learning rate of $1\cdot{10}^{-4}$.
We train each model for 100 epochs and choose the best one based on its performance on the development set.
All models are trained on a single
GeForce RTX 2070 GPU with 8~Gb of RAM. 
\begin{table}[t]
\caption{Four architectures 
	in terms of their
	components (encoder, pooling layer, classifier).
}
\label{tbl:architectures_localisation}
\centering
\begin{tabular}{@{}llll}
	\toprule
	Architecture       & $\enc$   & $\pool$      & $\clf$    \\ 
	\midrule
	PSC            & CNN      & log-mean-exp & ---      \\
	CNN-Pool       & CNN-Pool & max          & MLP(1024, $\vocab$) \\ 
	CNN-Attend     & CNN      & attention    & MLP(1000, 1) \\
	CNN-PoolAttend & CNN-Pool & attention    & MLP(1000, 1) \\
	\bottomrule
\end{tabular}
\end{table}
\section{Results: Keyword detection and spotting}
\label{sec:detection_spotting_results}
The reason we consider keyword detection in this dissertation is two-fold. First, to perform our main task---keyword localisation---we often need to first detect whether the keyword to be localised is present in the utterance before localising the keyword. Second, the performance of the model on keyword detection can also serve as an upper bound for the performance of the model on the keyword localisation task. The keyword detection models that we consider here are developed in previous work~\cite{palaz2016, kamper2017a, kamper2019b} but we make various architectural changes, for example by adding attention to perform attention-based keyword detection and keyword spotting.

To establish the baseline performance of the detection models, we trained two types of weakly supervised models---BoW and VGS---as described in Section~\ref{subsec:training}. Then, we assess their performance on keyword detection and spotting. We mentioned earlier that we consider a BoW model here because it can be seen as a perfect VGS model and gives an idea of how well a VGS model can perform on the task if we have a perfect visual tagger. 

Table~\ref{tbl:detection_evaluation} shows the detection scores for the different architectures when supervised with BoW labels and visual context. 
	Of the two forms of supervision, BoW labels lead to consistently better detection
	than visual supervision. This is not surprising since
	different speakers could describe the same image in many different ways. Moreover, the visual tagger (which provides the training signal here) can assign high probabilities to concepts that no speaker would refer to (but which is nevertheless present in an image) or could tag semantically related words. CNN-Pool also appears to give better
	scores than the PSC structure.
	The one reason that we did consider the CNN-PoolAttend and CNN-Pool architectures that use intermediate max pooling in their encoder structure
	is that it gives better
	scores than the other structure when only considering keyword detection, irrespective of the localisation. 
	But here we see that the visual CNN-Attend achieves slightly better detection $F_1$ than CNN-Pool ($32.7\%$ vs $31.2\%$); we will show later in Chapter~\ref{chap:methods-for-keyword-localisation} that the CNN-Attend also has the benefit of dramatically better localisation performance. Note that an attention-based model has not been considered for this task before.
	
Table~\ref{tbl:spotting_evaluation} shows the keyword spotting performance for the same models used for keyword detection. We find, for the most part, that similar findings on the keyword detection task hold for keyword spotting: the BoW models outperform VGS models consistently; CNN-Pool achieves better $P@10$ (44.4\%), $P@N$ (28.2\%)  and EER (22.8\%) than PSC (19.7\%, 14.0\% and 35.5\%, respectively); and CNN-Attend achieves slightly better scores in most of the metrics (43.7\%, 30.1\% and 22.7\%, respectively) than CNN-Pool.
	
	\begin{table*}[!t]
		\captionof{table}{
			Keyword detection scores (\%) for the four different architectures, where the task is to detect whether a given keyword occurs in an utterance (regardless~of~location).
		} 
		\label{tbl:detection_evaluation}
		\centering
		\renewcommand{\arraystretch}{1.1}
			\begin{tabularx}{1.0\textwidth}{@{}lCCCCCC@{}} 
				\toprule
				& \multicolumn{3}{c}{Visual} & \multicolumn{3}{c}{Bag-of-words}\\
				\cmidrule(lr){2-4}\cmidrule(l){5-7}
				Architecture & $P$ & $R$ & $F_1$ & $P$ & $R$ & $F_1$\\
				\midrule
				\textbf{PSC} & 28.5 & 11.3 & 16.2 & 86.9 & 75.7 & 80.9\\[3pt]
				\textbf{CNN-Pool} & 37.4 & 26.8 & 31.2 & 87.7 & 75.7 & 81.2\\[3pt]
				\textbf{CNN-Attend} & \bf 38.9 & \bf 28.2 & \bf 32.7 & \bf 89.6 & \bf 79.6 & \bf 84.3\\[3pt]
				\textbf{CNN-PoolAttend} & 35.1 & 22.6 & 27.5 & 81.5 & 65.0 & 72.3\\[2pt]
				\bottomrule
			\end{tabularx}
	\end{table*}

\begin{table*}[!t]
	\captionof{table}{
		Keyword spotting scores (\%) for the four different architectures, where the task is to rank utterances according to the probability that the utterances contain each keyword in the vocabulary of the system (regardless~of~location).
	} 
	\label{tbl:spotting_evaluation}
	\vspace*{-5pt}
	\centering
	\renewcommand{\arraystretch}{1.1}
		\begin{tabularx}{1.0\textwidth}{@{}lCCCCCC@{}} 
			\toprule
			& \multicolumn{3}{c}{Visual} & \multicolumn{3}{c}{Bag-of-words}\\
			\cmidrule(lr){2-4}\cmidrule(l){5-7}
			Architecture & $P@10$ & $P@N$ & EER & $P@10$ & $P@N$ & EER\\
			\midrule
			\textbf{PSC} & 19.7 & 14.0 & 35.5 & 80.2 & 68.1 & 10.0\\[3pt]
			\textbf{CNN-Pool} & \bf 44.4 & 28.2 & 22.8 & 95.5 & 75.0 & 6.2\\[3pt]
			\textbf{CNN-Attend} & 43.7 & \bf 30.1 & \bf 22.7 & \bf 95.7 & \bf 80.2 & \bf 5.9\\[3pt]
			\textbf{CNN-PoolAttend} & 31.6 & 24.4 & 27.0 & 92.8 & 72.6 & 10.2\\[2pt]
			\bottomrule
		\end{tabularx}
\end{table*}

	\section{Summary}
	
	This chapter presented visually grounded models for keyword detection and spotting. We started our investigation from models that were proposed before~\cite{kamper2017a, kamper2019b}, and further modified and extended them using attention, which has not been considered before.
	We described the keyword detection models that we will extend in the rest of this dissertation for the task of keyword localisation which is our primary goal. We also described the one dataset that we will use later on. Finally, we presented the VGS model's baseline and upper bound results---with a bag-of-words (BoW) model---and the analysis of the VGS models on the keyword detection and keyword spotting tasks. In particular, we trained four different  architectures for keyword detection to establish the keyword detection and spotting baselines. As mentioned, BoW models are helpful in this study to estimate the upper bound performance of a VGS model and to serve as a sanity check. 
	
	We found that BoW labels consistently lead to better keyword detection and spotting than visual supervision. This result is unsurprising since different speakers could describe the same image in many different ways. We also found that the attention-based VGS model achieved better keyword detection and spotting performance (in most cases) than the non-attention models. In the next chapter, Chapter~\ref{chap:methods-for-keyword-localisation}, we extend these models for the use of keyword localisation---the first time that these models are considered for this task.
	

%% file: localisation_methods.tex
\chapter{Methods for keyword localisation with VGS models}
\label{chap:methods-for-keyword-localisation}

In Chapter~\ref{chap:problem_formulation_setup} we presented the methodological framework for training and evaluating VGS models for keyword detection and spotting. We considered existing VGS models~\cite{kamper2017a, kamper2019b} which are explicitly trained for keyword detection and spotting using soft text labels obtained by passing each training image through the multi-label visual tagger of~\cite{kamper2019b}. Furthermore, we introduced a new VGS-based detection method using attention. 
In this chapter we address in full our first research question (Section~\ref{sec:research_questions}): Is keyword localisation possible with a VGS model? We extend the VGS models from Chapter~\ref{chap:problem_formulation_setup} for the task of keyword localisation. As a reminder, visually grounded keyword localisation involves using models trained on images and their spoken captions to find where in an utterance a given written keyword occurs. 

In this chapter we propose four different localisation methods to adapt the models for keyword localisation. One method employs the popular explainability method  \textit{Grad-CAM}~\cite{selvaraju2017}, which was originally developed in the context of images. This method can be used within any convolutional neural network architecture. The second and third methods use the fact that for certain architectures, the internal activations can be interpreted as localisation scores. Concretely, the second approach (\textit{score aggregation}) assumes that the final pooling layer is a simple transformation (for example, average or max pooling) so that the output score can be regarded as an aggregation of local scores. These can be used to select the most likely temporal location for the query keyword. The third approach (\textit{attention}) assumes that the network involves an attention layer that pools features over the temporal axis; we use the attention weights as localisation scores. The fourth approach (\textit{input masking}) involves masking the input signal at different locations and measuring the response score predicted by the trained model on the partial inputs; large variations in the output suggest the presence of a keyword. Input masking is similar to the first approach because it can be applied to any classification model after training, irrespective of architecture.

All of the different models adapted for keyword localisation are trained in exactly the same way as the keyword detection models in the previous chapter: there is no alignment data during training (the model is never given the location of any keyword during training). The only difference appear at prediction time, where we extract additional information from the models.

\begin{tcolorbox}[width=\linewidth, colback=white!95!black, boxrule=0.5pt]
	\small
	\textit{The first and second localisation methods and their results were reported in:} \\
	K. Olaleye, B. van Niekerk, and H. Kamper, ``Towards localisation of keywords in speech using weak supervision,'' \textit{NeurIPS Self-Supervised Learning for Speech and Audio Processing Workshop}, 2020. \\
	
	\textit{The third localisation method was reported in:} \\
	K. Olaleye, and H. Kamper, ``Attention-based keyword localisation in speech using visual grounding,'' \textit{Interspeech}, 2021. \\
	
	\textit{The fourth localisation method was reported in:}\\
	K. Olaleye, D. Oneață, and H. Kamper, ``Keyword localisation in untranscribed speech using visually grounded speech models,'' \textit{Journal of Selected Topics in Signal Processing}, 2022.
	
\end{tcolorbox}

In the remainder of this chapter, we describe the evaluation of these VGS models for different keyword localisation tasks (Section~\ref{sec:keyword_localisation_evaluation}); and we present details about the proposed localisation methods and give a brief survey of existing studies on these methods (Section~\ref{sec:localisation_method}). Finally, we present the performance of the proposed localisation methods, and give a systematic and fair analysis of the methods (Section~\ref{sec:main_findings}).

\section{Evaluating VGS models for keyword localisation}
\label{sec:keyword_localisation_evaluation}
	The main task performed in this chapter (but also in the overall dissertation) is keyword localisation: given an audio utterance and a keyword, find where in the input the keyword occurs.
	As mentioned in the introduction of this chapter, the chief idea is to perform localisation with the same audio networks trained for detection in Chapter~\ref{chap:problem_formulation_setup}---i.e., \ the network is trained without keyword location labels.
	We give specific details about the different proposed methods that can adapt a keyword detection model for localisation; each of them is described in Section~\ref{sec:localisation_method}. Here, we only give details about how we evaluate the VGS models on the localisation tasks.
	
\begin{figure}[t]
		\centering
		\includegraphics[width=0.75\columnwidth]{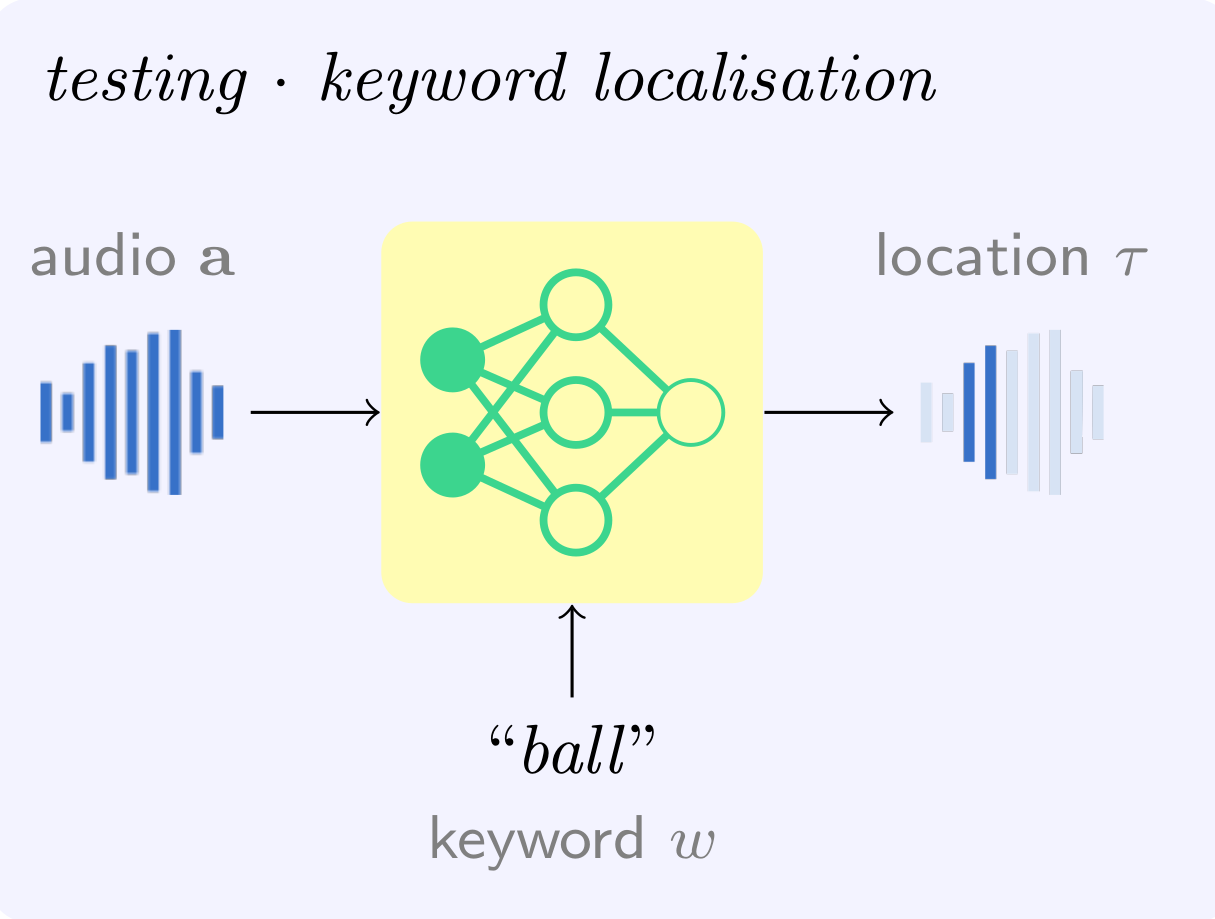}
		\caption{In keyword localisation, the task at test time is to locate a given keyword. In this example, the network is used to locate the keyword \textit{ball} in the input utterance $\ab$. 
		}
		\label{fig:testing_localisation}
	\end{figure}

	To compute
	location, each of the localisation methods assigns a score $\alpha_{w,t}$ for each keyword $w$ and location $t$ of the input audio sequence.
	The predicted location $\tau_w$ for the input keyword $w$ is the position of the highest score across time:
	\begin{equation}
		\tau_w = \argmax \left\{ \alpha_{w,1}, \dots, \alpha_{w,\temp} \right\}.
		\label{eq:location-max}
	\end{equation}
	This is illustrated in Figure~\ref{fig:testing_localisation} (which is the bottom-right part of Figure~\ref{fig:overview_artificial} presented earlier). We use $\temp$ to denote the length of the input sequence.
\begin{figure}[bt]
	\centering
	\includegraphics[width=0.99\linewidth]{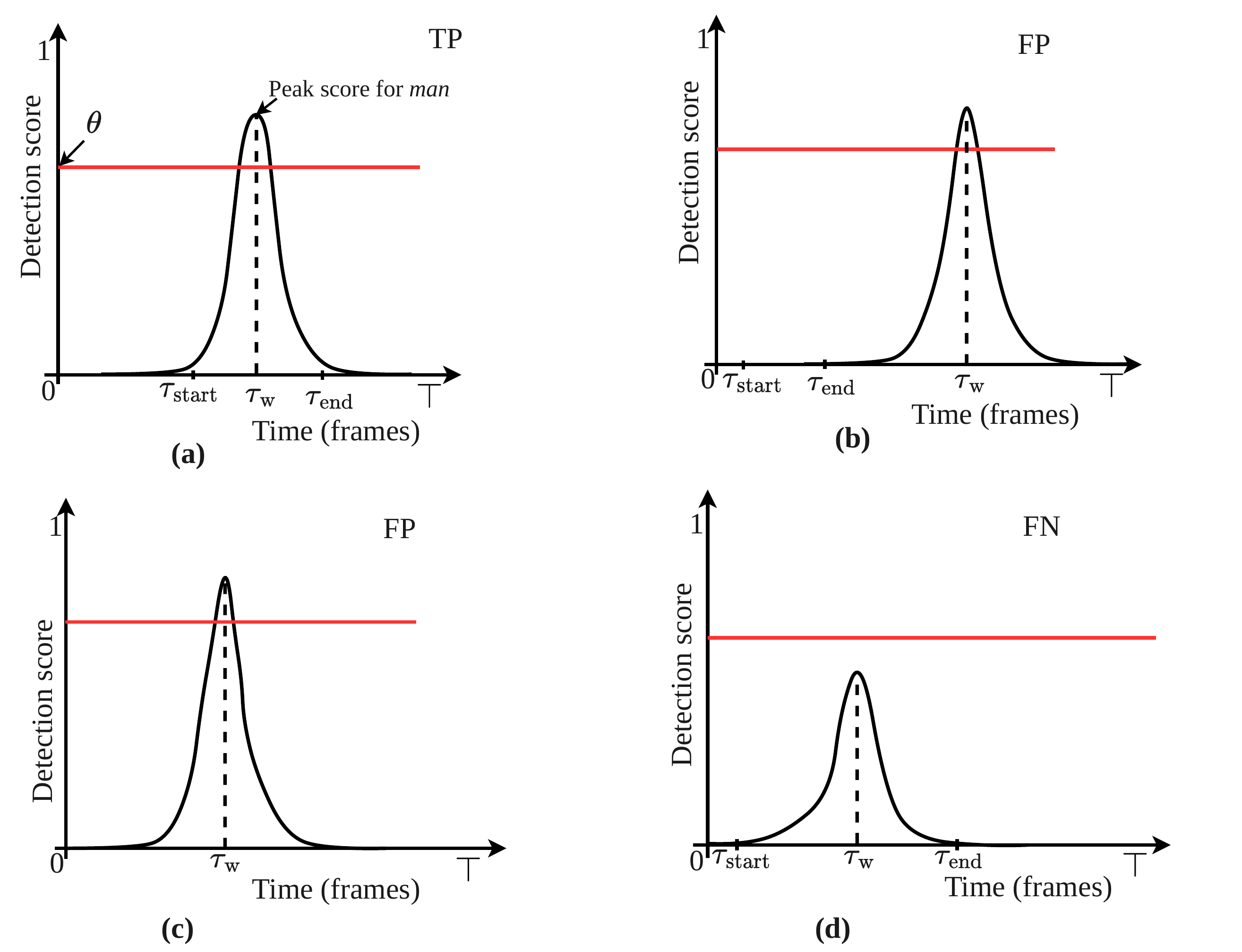}
	\caption{An illustration of how to assess the performance of our mechanism on locating
		the keyword ``man" in four utterances represented by curves (a), (b), (c), and (d). Curve
		(a) is completely labelled to specify what each part of the curves represents.
	}
	\label{fig:localisation_evaluation}
\end{figure}

	Keyword localisation can be evaluated in different ways depending on the final use case.
	In some cases, we might know that a keyword occurs within an utterance and are asked to predict where the keyword occurs (oracle keyword localisation, below).
	In other cases, we might first need to detect whether a keyword occurs before doing localisation (actual keyword localisation).
	Alternatively, we might be required to rank utterances in a speech collection and only do localisation on the utterances most likely to contain the given keyword (keyword spotting localisation).

		In all the different localisation scenarios, we consider a keyword localised if its predicted (point) location falls within the interval corresponding to the true location of the keyword, according to forced alignments.
		We give specific details below.
	
		\subsection{Oracle keyword localisation} 
		\label{ssec:oracle_keyword_localisation_measure}
		In this setup, we already know that a specified keyword occurs in an utterance, and we need to find the location of that keyword.
		As such, we do not need first to do detection since we already know that the keyword is present.
		This scenario is referred to as \textit{oracle localisation performance} in~\cite{palaz2016} and represents a setting where detection is perfect, and we only consider keyword localisation performance.
		Localisation is performed as in Equation~\ref{eq:location-max}, and we report the accuracy: the proportion of proposed word locations that fall within the ground truth word boundaries.
		
		For clarity, we give an example in Figure~\ref{fig:localisation_evaluation}. We assume there are four utterances in the test corpus and only one keyword \textit{man} to localise.
		To compute oracle accuracy, we consider only the utterances that include the keyword \textit{man}. In all cases, the keyword \textit{man} starts at ($\tau_{\text{start}}$ and ends at $\tau_{\text{end}}$), according to forced alignments of the data. In this case, oracle accuracy would be:
		\begin{equation*} 
			\begin{aligned}
				\text{Oracle accuracy} &= \frac{\text{\# peaks on \textit{man}}}{\text{\# \textit{man} in ground truth}} \\ &= \frac{\# [(a), (d)]}{3} = \frac{2}{3}. \\\\
			\end{aligned}
		\end{equation*}
	
		\subsection{Actual keyword localisation}
		\label{ssec:actual_keyword_localisation_measure}
		In the \textit{actual keyword localisation setting}, we don't know whether a specified keyword occurs in an utterance. Hence, in order to localise the keyword, we first check whether the detection score $\hat{y}_w$ (as specified in Section~\ref{sec:evaluation-tasks}) of keyword $w$ is greater than a specified threshold $\theta$,
		and then select the position of the highest attention weight $\alpha_{t, w}$ as the predicted location $\tau_w$, as in Equation~\ref{eq:location-max}.
		The location $\tau_w$ is accepted as correct if it falls within the interval corresponding to the true location of the word (which is determined based on the forced alignment between the text caption and audio utterance).
		The model is penalised whenever it localises a word whose detection score is less than or equal to $\theta$, i.e., \ if a word is not detected 
		then it is counted as a failed localisation even if $\alpha_{t, w}$ is at a maximum within that word.
		We set the value of $\theta$ to $0.5$, which gave the best performance on development data.
		We report 
		precision ($P$), recall ($R$), and $F_1$.

	Again, to understand this better, we consider the illustrative example in Figure~\ref{fig:localisation_evaluation}. To compute the actual localisation performance, in this case, we consider all four utterances---Figure~\ref{fig:localisation_evaluation}, curves (a), (b), (c) and (d). Precision, recall and $F_1$ are then calculated as follows:
	\begin{equation*} 
		\begin{aligned}
			P &= \frac{\text{TP}}{\text{TP} + \text{FP}} = \frac{1}{1 + 2} = \frac{1}{3}. \\\\
			R &= \frac{\text{TP}}{\text{TP} + \text{FN}} = \frac{1}{1 + 1} = \frac{1}{2}. \\\\
			F_1 &= 2 \times \frac{P \times R}{P + R} = \frac{2}{5}.
		\end{aligned}
	\end{equation*}
		
		\subsection{Keyword spotting localisation}
		As a reminder from Section~\ref{sec:keyword_detection_spotting_VGS}, in keyword spotting
		we are given a keyword and a collection of utterances and
		we want to rank the utterances according to the probability that the utterances contain the keyword.
		Keyword spotting is evaluated
		in terms of 
		$P@10$, the average precision of the ten highest-scoring proposals; and
		$P@N$, the average precision of the top $N$ proposals, with $N$ the number of true occurrences of the keyword (see Section~\ref{subsec:keyword_spotting_evaluation}). 
		Similar to the actual keyword localisation evaluation procedure (described in Section~\ref{ssec:actual_keyword_localisation_measure}) which we use for measuring our models' ability to localise detected keywords, we adapt the related task of keyword spotting to the localisation scenario.
		
		In the case of \textit{keyword spotting localisation}, we additionally consider the localisation of the given keyword: a sample is deemed correct if it is both highly ranked (spotted) and the predicted location falls within the ground truth word.
		In this chapter, we report keyword spotting localisation performance as $P@10$, which is the proportion of top-10 utterances for which the location is correctly predicted. If an utterance is erroneously ranked in the top 10 (i.e., \ the utterance does not contain the specified keyword), localisation performance is penalised (i.e., \ this utterance is considered a failed localisation).

\begin{table}[t]
	\caption{Four architectures 
		in terms of their
		components (encoder, pooling layer, classifier) and
		localisation method (Grad-CAM \textsf{GC}, input masking \textsf{M}, score aggregation \textsf{SA}, and attention \textsf{Att}).}
	\label{tbl:architectures_localisation_chp4}
	\centering
	\begin{tabular}{@{}llllccccc}
		\toprule
		Architecture       & $\enc$   & $\pool$      & $\clf$    &  & \textsf{GC} & \textsf{M} & \textsf{SA} & \textsf{Att} \\
		\midrule
		PSC            & CNN      & log-mean-exp & ---       &  &             &            & $\bullet$   & \\
		CNN-Pool       & CNN-Pool & max          & MLP(1024, $\vocab$) &  & $\bullet$   &            &             & \\
		CNN-Attend     & CNN      & attention    & MLP(1000, 1) &  &             & $\bullet$  &             & $\bullet$ \\
		CNN-PoolAttend & CNN-Pool & attention    & MLP(1000, 1) &  &             & $\bullet$  &             & $\bullet$ \\
		\bottomrule
	\end{tabular}
\end{table}
\subsection{On point-based evaluation}
An alternative to the point-based localisation that we consider throughout this dissertation would have been to predict and evaluate an interval for when a word starts and ends. However, we opt for the point-based evaluation for several reasons: (i) Even in speech models trained on transcriptions, it is challenging to predict the precise interval of a word \cite{sanabria2021}. (ii) Our evaluation approach avoids an extra post-processing step, leading to a more straightforward and fairer comparison between the localisation methods. Otherwise, for three of the proposed methods (Grad-CAM, score aggregation, attention; discussed in Section~\ref{sec:localisation_method}), we would need to devise a method for grouping the most promising segments into a larger segment. (iii) A similar evaluation approach is also encountered in the computer vision literature, especially in papers proposing saliency models \cite{bylinskii2016,Zhang2018,rebuffi2020}.

\section{Localisation methods}
\label{sec:localisation_method}
In Chapter~\ref{chap:problem_formulation_setup} we detailed how we train and evaluate our VGS models for keyword detection and keyword spotting.\ We specifically considered four different convolutional-based architectures (summarised here in Table~\ref{tbl:architectures_localisation_chp4}). The core part of these architectures is adopted from an existing VGS model~\cite{kamper2017a, kamper2019b} which we modify in this section for keyword localisation. 
In particular, we assume we are given an audio network trained for the task of keyword detection (Section~\ref{sec:keyword_detection_spotting_VGS}), and we present four classes of methods that can augment the network's output with localisation scores.
\subsection{Grad-CAM}
\label{sec:grad-cam}

Gradient-based class activation map (Grad-CAM)~\cite{selvaraju2017} is a saliency-based method that can be applied to any convolutional neural network architecture to determine which parts of the network input most contribute to a particular output decision.
Intuitively, it works by determining how vital each filter in a convolutional layer is to a particular output prediction (a specified keyword in our case).
Localisation scores are then calculated for an input utterance by weighing the filter activations with their importance for the keyword under consideration.
We use the scores computed after the last convolutional layer as localisation scores.

Grad-CAM first determines how important each filter in the output of the last convolutional layer is to the word $w$. More formally, let us call the  output from the last one-dimensional convolutional layer $\hb_t$ where each $\hb_t \in \mathbb{R}^{\emb}$ is a $\emb$-dimensional vector  with $\emb$ the number of filters.
We first determine the ``importance" of the $k^{\textrm{th}}$ filter to the word $w$:
\begin{equation}
	\gamma_{k, w} = \frac{1}{\temp} \sum^{\temp}_{t = 1} \frac{\partial \hat{y}_w}{\partial h_{t, k}}.
\end{equation}
This value indicates how closely the $k^{\textrm{th}}$ convolutional filter pays attention to word $w$, based on the trained model weights.
The output activation $\hb_t$ at time step $t$ is then weighted by its importance $\mathbf\gamma_w$ towards word $w$, giving the localisation scores. We are only interested in changes that would result in a higher score for $w$, so we take the $\mathrm{ReLU}$, resulting in:
\begin{equation}
	\alpha_{w, t} = \mathrm{ReLU} \left[ \sum_{k = 1}^{\emb} \gamma_{k, w} h_{t, k} \right].
\end{equation}

\subsubsection*{Relation to prior work} 
For context, we briefly give some background on the development of Grad-CAM.
Class activation maps (CAM) is proposed in \cite{zhou2016} for object localisation. Their network consists of convolutional layers, a global average pooling performed on the feature maps and fed directly into (softmax) classification layer. The importance of the image regions in discriminating an object is then identified by projecting back the weights of the output layer on to the convolutional feature maps. The weighted sum of the feature maps of the last convolutional layer yields the class activation map. 
Grad-CAM \cite{selvaraju2017} is an attempt to generalise CAM beyond fully-convolutional-based models. The motivation for the extension, as mentioned in the Grad-CAM paper, is to avoid trading off model complexity and performance (associated with the removal of fully-connected layers) for easier explanation of the model as is the case in CAM. 
They use ReLU as a threshold to capture only the feature maps that have positive influence on the class of interest.

\subsection{Score aggregation}
\label{subsubsec:score-agg}
\begin{figure}[t]
	\centering
	\includegraphics[width=1.0\columnwidth]{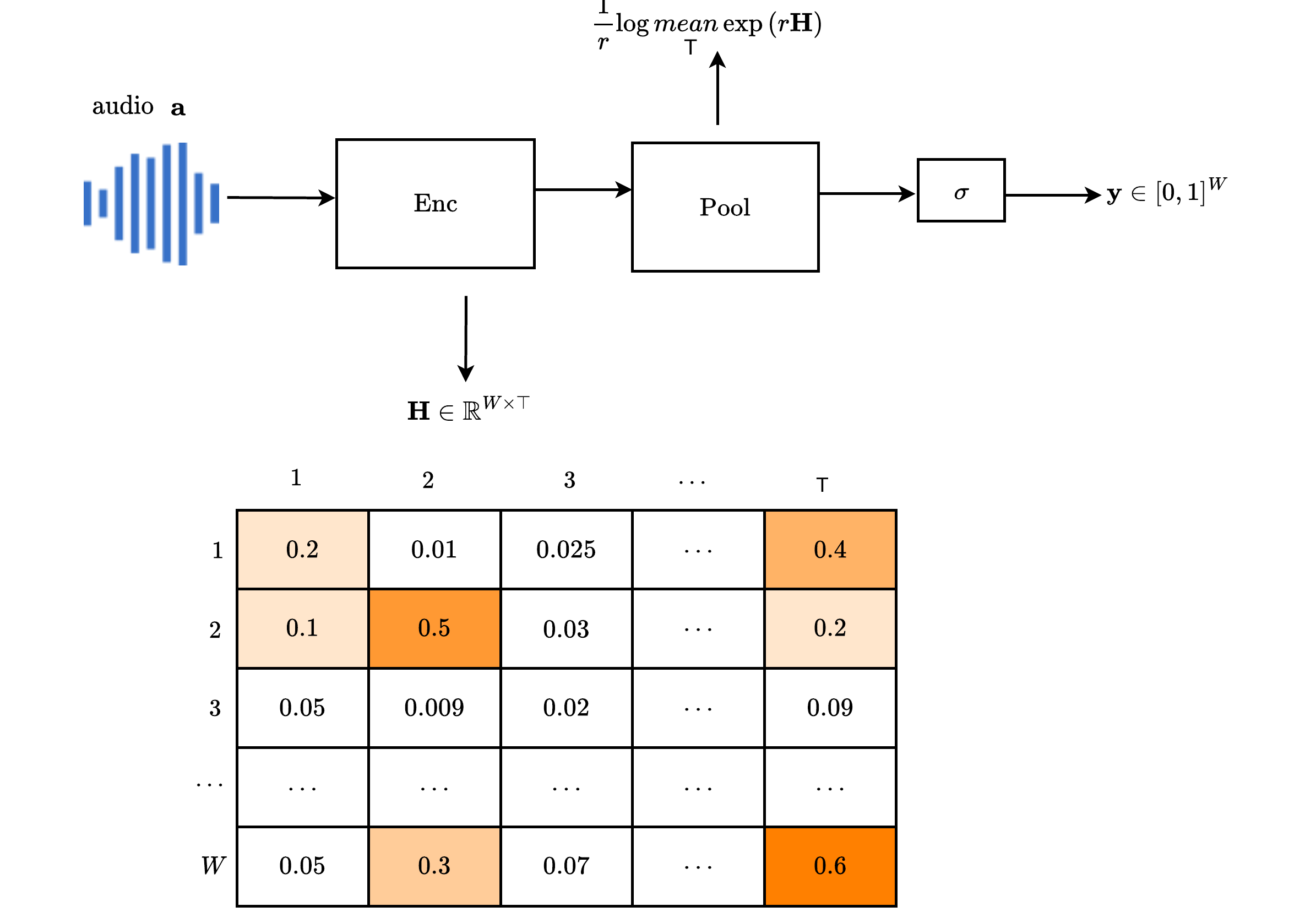}
	\captionof{figure}{An illustration of the score aggregation localisation method. The orange region represents how the output of the encoder ($\mathrm{Enc}$) is unpacked to give a score $h_{w,t}$ for each frame $t$ and each word $w$.
		In this example, the proposed location for the 
		keywords corresponding to indices $2$ and $\vocab$ would respectively be 
		frames $2$ and $\temp$, which achieve the highest scores in the depicted 
		orange region.}
	\label{fig:psc_architecture}
\end{figure}
The core part of the score aggregation method is its use of the log-mean-exp pooling function which we described in Section~\ref{subsec:exp_setup}. This is a soft pooling method that summarises the output from a speech encoder into a fixed-dimensional vector. In this section we give details about how we use the function to facilitate keyword localisation. The score aggregation method was originally designed in~\cite{palaz2016} to enable a model trained only for word classification and locate a predicted word.
It does so by explicitly making the architecture location-aware.	
The final convolutional layer must have the same number of filters as the final output of the model. By aggregating the output of this final convolutional layer in a specific way (described below; and depicted in Figure~\ref{fig:psc_architecture}), we can 
backtrace from the output prediction to determine which of the filters
at which time position causes that particular output prediction. 

More formally, the score aggregation method interprets the audio embeddings $\Hb = \left[\hb_1, \dots, \hb_\temp\right]$ as localisation scores $\alpha$, that is
$\alpha_{w, t} = h_{w, t}$. This method relies on two assumptions:
\begin{itemize}
	\item[1.] The embedding dimension $\emb$ is equal to the number of keywords in the vocabulary $\vocab$.
	\item[2.] The pooling layer is the final layer to preserve the semantics of the activations as scores. Hence, the method cannot be applied to an arbitrary network. Concretely, it needs to be used with the PSC architecture described in Section~\ref{subsec:exp_setup} which has no classifier layer following the pooling layer. The PSC name comes from the author names of the original paper~\cite{palaz2016}.
\end{itemize}
To aggregate the frame-level scores into an utterance-level detection score (across the $\temp$ steps), we use the log-mean-exp function
followed by the sigmoid function $\sigma$ (which ensures that the values are in $[0, 1]$ and can be interpreted as probabilities):
\begin{equation}
	\pool(\Hb) = \sigma\left(\frac{1}{r} \log \mean_{\temp} \exp \left(r \Hb\right)\right).
	\label{eq:log-mean-exp}
\end{equation}
Here $r$ is a positive number that controls the type of pooling:
when $r \to 0$ we obtain average pooling,
when $r \to \infty$ we obtain max pooling. By controlling the hyperparameter, we can therefore get something in between these two extremes. According to~\cite{palaz2016}, this intermediate aggregation operation drives the weights of frames which have similar scores close to each other during training, resulting
in better localisation performance.
\subsubsection{Relation to prior work}
Palaz \etal~\cite{palaz2016} proposed this score aggregation method to jointly locate and classify words using only BoW supervision (see Section~\ref{subsec:training}).
In this dissertation, we instead train the score aggregation method using visual supervision.
For image convolutional neural networks, a related technique is CAM~\cite{zhou2016} (discussed in Section~\ref{sec:grad-cam}),
which decomposes the classification score of a class as a sum of spatial scores from the last feature map in the convolutional neural network.
This relies on the assumption that the last feature map is passed through a global average pooling layer and a linear classifier.
This approach is related to Grad-CAM since
both methods compute spatial scores by pooling the convolutional feature maps across the channel axis.
The main difference is
that CAM aggregates using the linear classifier's weights
while Grad-CAM uses gradients of the output with respect to the feature map.

\subsection{Attention}
\label{ssec:attention}
\begin{figure}[!t]
	\centering
	\includegraphics[width=\columnwidth]{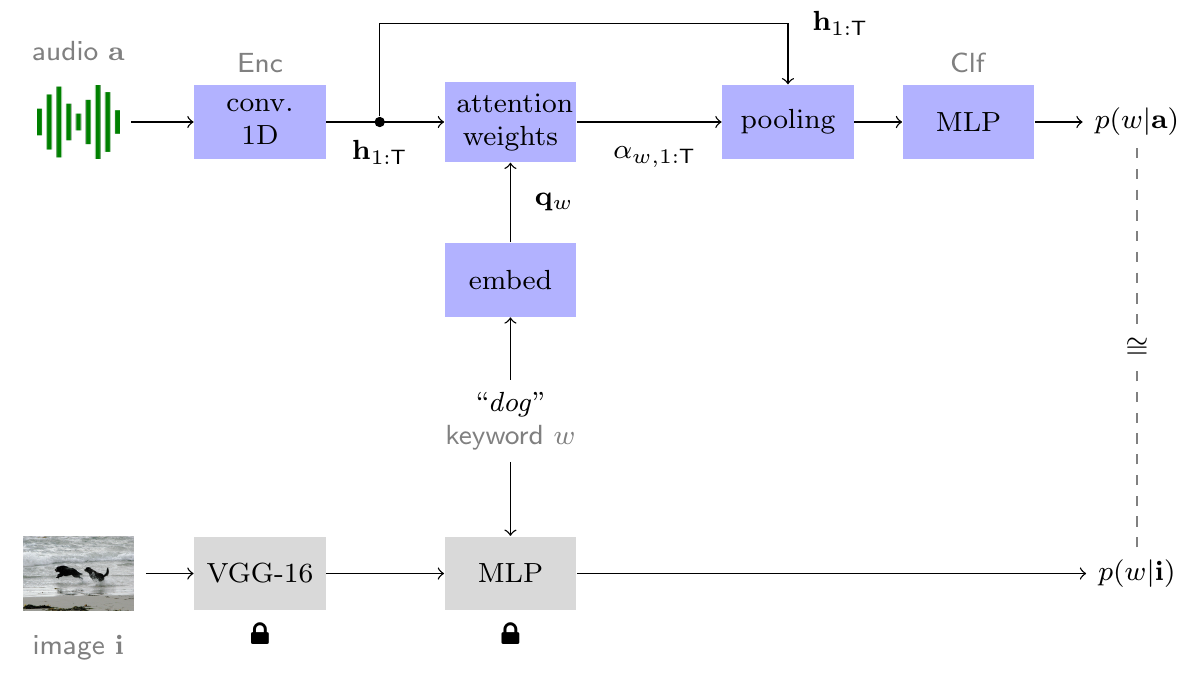}
	\caption{%
		Our attention-based architecture for keyword localisation in speech.
		The speech network (in blue) first passes the audio input $\ab$ through a convolutional encoder.
		The resulting features $\hb_{1:\mathsf{T}}$ are then pooled using an attention mechanism based on the embedding $\mathbf{q}_w$ of the query keyword $w$.
		The previous step yields a vector that finally goes through a multi-layer perceptron (MLP) to output the probability $p(w|\ab)$ of the keyword occurring in the audio.
		The visual network (in grey) embeds the corresponding image $\ib$ into a vector using the VGG-16 network and computes the probability $p(w|\ib)$ of the keyword occurring in the image
		using a keyword specific MLP.
		Note that the visual network is used only at training time (for supervising the speech branch) and its weights are fixed; at test time we only use the speech branch at the top.
	}
	\label{fig:attention_mechanism}
\end{figure}

Attention has become a standard mechanism to allow a neural network to automatically weigh different parts of its input when producing some (intermediate) output~\cite{graves2014,luong2015}. We described earlier (see Section~\ref{subsec:exp_setup}) how we use the attention approach for keyword detection. We now give full details about how we use the mechanism to perform keyword localisation in speech. 
Intuitively, a given textual keyword is mapped to a learned query embedding, which is then used to weigh the relevance of intermediate convolutional filters. Time steps that are weighed highest (i.e.,\ are attended to most) are likely to be most indicative of the location of a particular keyword.
The method is illustrated in Figure~\ref{fig:attention_mechanism}.

Concretely, we start by embedding the given keyword $w$ into a $\emb$-dimensional embedding $\mathbf{q}_w$ (query) using an index-based lookup table. 
The attention module then computes a score $e_{w, t}$ for each time step $t$ by using a dot product to measure the similarity between each time step along the convolutional features $\mathbf{h}_t$ and the query embedding,
$e_{w, t} = \mathbf{q}_w^{\intercal} \mathbf{h}_t$,
followed by a softmax operation which converts the similarity score $e_{w, t}$ to attention weights:
\begin{equation}
	\alpha_{w, t} = \frac{\exp \{ e_{w, t} \} }{\sum_{{t' = 1}}^{\temp}\exp \{ e_{w, t'} \}}.
	\label{eq:attention-weights}
\end{equation}
The attention weights are used for keyword localisation (as localisation scores). The attention weights can also be used for keyword detection (as intermediary features)---as mentioned in Section~\ref{subsec:exp_setup}---by pooling the features $\mathbf{h}_t$ at every time step $t$ into a keyword-specific context vector:
\begin{equation}
	\pool(\Hb)_w = \sum_{t = 1}^{\temp} \alpha_{w, t} \mathbf{h}_t
\end{equation}
and computing the score for query $w$ by passing the context vector through the classifier $\clf$,
which is typically implemented as a series of fully-connected layers (a multi-layer perceptron).
\subsubsection{Relation to prior work}
Our approach is very heavily inspired by Tamer \etal~\cite{tamer2020}, who performed keyword search in sign language
using an attention-based graph-convolutional network over skeleton joints.
We adapted their architecture to fit our keyword localisation task in speech.
The attention mechanism was also used for the general case of multiple instance learning~\cite{ilse2018}.
This task assumes the classification of bags of instances:
a bag is positive if it contains at least one positive instance.
The challenge lies in the fact that the labels are given only at the level of the bag, while the instance labels are unknown.
The idea is to pool the instances using attention and use the corresponding weights as an indication of whether an instance is positive.
As a practical use case, Ilse \etal \cite{ilse2018} applied this technique to highlight cancerous regions in weakly-labelled images.

\subsection{Input masking}
\label{subsubsec:input-masking}
In this section we introduce two new techniques that have never been considered for keyword localisation before.
Rather than looking at intermediate activations within a network, the methods modify a network's input at specific locations and then observe the impact on the output detection probabilities.
For instance, in the \textit{masked-in} approach described formally below, we pass only a part of an input utterance through the model; if this increases the output probability of a particular word, that could be an indication that the word occurs in the segment.
In the \textit{masked-out} approach, we instead pass an entire utterance through a network but mask out a part of the input; if this decreases the output probability of a particular word, that could be an indication that the word occurs in the masked-out segment.
Both methods have the advantage that they can be applied to any type of detection model. 
However, depending on how segments in an utterance are masked in or out, they can be computationally expensive since we might need to make several passes through the model for a single utterance.
\begin{figure}[bt]
	\centering
	\includegraphics[width=1.0\linewidth]{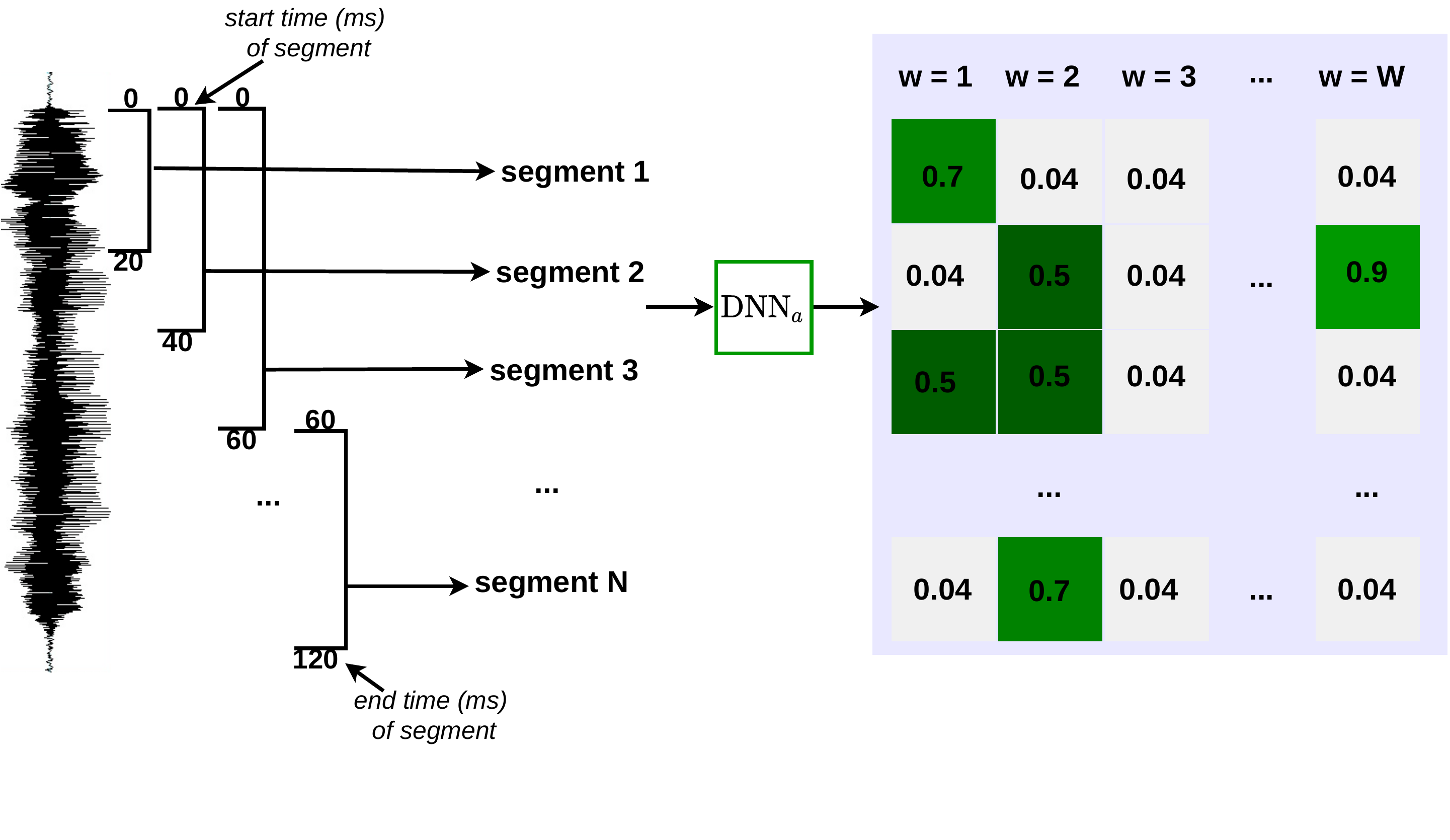}
	\caption{Our masked-in approach to keyword localisation in speech. An input utterance is split into overlapping segments of varying durations. Each segment is then fed into the network ($\dnna$) to determine the output detection probability of a particular word. The segment corresponding to the maximum output detection probability of the word is a possible location of the word. For instance, the most probable segment containing the word with index $w=1$ is segment 1, with $\alpha_{1, 1} = 0.7$.
	}
	\label{fig:masking_localisation_approach}
\end{figure}

More formally, the 
\textit{masked-in} approach, as depicted in Figure~\ref{fig:masking_localisation_approach}, treats the keyword localisation task as a keyword detection task.
Specifically, we split a test utterance into segments of minimum duration of 200 ms and maximum duration of 600 ms, with consecutive segments overlapping by 30 ms. We roughly tuned the minimum and maximum duration hyperparameters on the development data.

Each segment is zero-padded back to the actual length of the utterance.
We then feed each padded segment into the trained model to compute a detection score for a keyword. This process yields the following localisation scores:
\begin{equation}
	\alpha_{w, t} = \sigma\left(\dnna(\ab \odot \mathbf{m}_t)_w\right),
\end{equation}
where $\mathbf{m}_t$ is a binary mask with ones to the entries corresponding to the $t$-th segment and zeros otherwise,
and $\odot$ denotes element-wise multiplication.
The segment with the highest detection score for the keyword is treated as the location of the keyword.

The \textit{masked-out} approach is the direct opposite, where we occlude (by replacing it with zeros) each segment from the utterance to compute the detection score.
If the masked-out segment contains the keyword, we expect a detection score that is close to $0$.
In order to select the location $\tau$ based on the maximum (Equation~\ref{eq:location-max}), we use the complement of the occurrence probabilities as localisation scores:
\begin{equation}
	\alpha_{w, t} = 1 - \sigma\left(\dnna(\ab \odot (1 - \mathbf{m}_t))_w\right).
\end{equation}

\subsubsection{Relation to prior work}

The masked-in method operates similarly to the sliding window approach often used for object detection in images~\cite{viola2001}.
An important distinction is that in our case the network that is repeatedly applied to each window is trained in a very weak setting with noisy supervision labels and without any location information.
The masked-out approach is related to the work of Zeiler and Fergus~\cite{zeiler2014}; they 
investigated whether convolutional neural network-based vision models are really
identifying the location of an object in an image or just finding
the surrounding context. 
We adapted their method of systematically perturbing different portions of the input image, but apply this approach to speech here. In the speech domain,
Harwath and Glass~\cite{harwath2017} proposed a method for connecting word-like acoustic units of an utterance to semantically relevant image regions. 
They did this by extracting segments from the utterance and then used their  multimodal network to associate them with an appropriate subregion of the image. This is different from our setting because rather than matching speech segments to relevant image regions, our method matches speech segments to textual labels.
Using a model that is more similar to ours, Kamper \etal~\cite{kamper2019a} investigated the task of retrieving utterances relevant to a given spoken query.
They did this by comparing the query embedding to the embedding of each of the utterance sub-segments using cosine distance, and then treated the minimum cosine distance as the score for the relevance of the query to the utterance. 
Here, we use their approach of splitting each utterance into overlapping segments varying from some minimum duration to some maximum duration.
But here the query is a written keyword (not a spoken query) and, much more importantly, \cite{kamper2019a} did not evaluate localisation performance.

\subsection{Architectures and implementation details}
\label{ssec:implementation_details}
As mentioned, we used the VGS architectures described in Chapter~\ref{chap:problem_formulation_setup} as our base. A summary of these architectures and the corresponding localisation methods tied to them is presented in Table~\ref{tbl:architectures_localisation_chp4}. Some localisation methods are tied to particular architectures, for example, score aggregation is directly incorporated into the PSC~\cite{palaz2016} architecture and loss which uses log-mean-exp as its global pooling function, and CNN as its audio encoder. The attention localisation method uses a classifier that outputs a single detection score for whether a given keyword occurs.
Other localisation methods can be applied irrespective of the architecture, for example, masking-based localisation can, in principle be used with any architecture. All models are implemented as already described in Section~\ref{subsec:exp_setup}.

\section{Results: Keyword localisation}
\label{sec:main_findings}
\subsection{Quantitative analysis}
\label{ssec:quantitative_results}
Our main results are presented in Tables~\ref{tbl:oracle_localisation_evaluation}, \ref{tbl:actual_localisation_evaluation}, and~\ref{tbl:spotting_localisation_evaluation}.
Table~\ref{tbl:oracle_localisation_evaluation} shows the performance of both VGS 
models and BoW-supervised models on the oracle localisation task.  Table~\ref{tbl:actual_localisation_evaluation} shows performance on the actual keyword localisation task, and Table~\ref{tbl:spotting_localisation_evaluation} shows performance on the keyword spotting localisation task.

Our main research goal is to investigate to what extent keyword localisation is possible with a VGS model.
Focussing therefore on the results in the ``Visual'' columns in
Tables~\ref{tbl:oracle_localisation_evaluation}, \ref{tbl:actual_localisation_evaluation}, and~\ref{tbl:spotting_localisation_evaluation}, we see that modest performance is achievable with visually supervised models.
In the oracle setting (Table~\ref{tbl:oracle_localisation_evaluation}), where the system knows that a keyword is present and then asked to predict where it occurs, accuracy is above 57\% using the best approach.
When localisation is performed after detection, the best $F_1$ score is at 25.2\% (Table~\ref{tbl:actual_localisation_evaluation}), while the best keyword spotting localisation $P@10$ is at 32.1\% (Table~\ref{tbl:spotting_localisation_evaluation}).
Although these scores are modest, these keyword localisation results are achieved by models that do not receive any explicit textual supervision.

In what follows, we try to quantify the drop in performance due to the first detection or ranking pass, we consider the effect of architectural choices, and break down the performance for individual keywords.
\begin{table*}[!t]
	\captionof{table}{
		Results on test data
		for the oracle keyword localisation task (when assuming perfect detection).
		We report results (in percentage) in terms of accuracy.
		We consider both visual and idealised bag-of-words supervision.
	} 
	\label{tbl:oracle_localisation_evaluation}
	\centering
	\renewcommand{\arraystretch}{1.1}
		\begin{tabularx}{1.0\textwidth}{@{}llCC@{}} 
			\toprule
			Architecture & Localisation & Visual & Bag-of-words \\
			\midrule
			\textbf{PSC} & score agg. & 13.7 & 61.5 \\[3pt]
			\textbf{CNN-Pool} & Grad-CAM & 12.7 & 13.3 \\[3pt]
			\textbf{CNN-Attend} & attention & 46.0 & 73.7 \\[3pt]
			\textbf{CNN-Attend} & masked-in & \bf 57.3 & \bf 87.5 \\[3pt]
			\textbf{CNN-Attend} & masked-out & 25.1 & 28.2 \\[3pt]
			\textbf{CNN-PoolAttend} & attention& 16.5 & 33.6 \\[2pt]
			\textbf{CNN-PoolAttend} & masked-in & 41.6 & 57.3 \\[3pt]
			\textbf{CNN-PoolAttend} & masked-out & 22.3 & 28.7 \\
			\bottomrule
		\end{tabularx}
\end{table*}

\begin{table*}[!t]
	\captionof{table}{
		Results on test data
		for the actual keyword localisation task.
		We report results (in percentage) in terms of $F_1$.
		We consider both visual and idealised bag-of-words supervision. We use a threshold of $\theta = 0.5$ to decide whether a keyword occurs in the utterance.
	} 
	\label{tbl:actual_localisation_evaluation}
	\centering
	\renewcommand{\arraystretch}{1.1}
		\begin{tabularx}{1.0\textwidth}{@{}llCC@{}} 
			\toprule
			Architecture & Localisation & Visual & Bag-of-words \\
			\midrule
			\textbf{PSC} & score agg. & 7.4 & 68.1 \\[3pt]
			\textbf{CNN-Pool} & Grad-CAM & 8.1 & 17.4 \\[3pt]
			\textbf{CNN-Attend} & attention & \bf 25.2 & 72.1 \\[3pt]
			\textbf{CNN-Attend} & masked-in & 24.9 & \bf 79.8 \\[3pt]
			\textbf{CNN-Attend} & masked-out & 12.9 & 57.8 \\[3pt]
			\textbf{CNN-PoolAttend} & attention& 8.2 & 39.0 \\[2pt]
			\textbf{CNN-PoolAttend} & masked-in & 18.0 & 55.3 \\[3pt]
			\textbf{CNN-PoolAttend} & masked-out & 10.4 & 48.1 \\
			\bottomrule
		\end{tabularx}
\end{table*}
\begin{table*}[!t]
	\captionof{table}{
		Results on test data for the keyword spotting localisation task.
		We report results (in percentage) in terms of precision at rank 10 ($P@10$).
		We consider both visual and idealised bag-of-words supervision.
	} 
	\label{tbl:spotting_localisation_evaluation}
	\centering
	\renewcommand{\arraystretch}{1.1}
		\begin{tabularx}{1.0\textwidth}{@{}llCC@{}} 
			\toprule
			Architecture & Localisation & Visual & Bag-of-words \\
			\midrule
			\textbf{PSC} & score agg. & 8.2 & 62.2 \\[3pt]
			\textbf{CNN-Pool} & Grad-CAM & 4.2 & 15.2 \\[3pt]
			\textbf{CNN-Attend} & attention & \bf 32.1 & 79.7 \\[3pt]
			\textbf{CNN-Attend} & masked-in & 21.9 & \bf 86.6 \\[3pt]
			\textbf{CNN-Attend} & masked-out & 5.1 & 7.9 \\[3pt]
			\textbf{CNN-PoolAttend} & attention& 8.1 & 42.7 \\[2pt]
			\textbf{CNN-PoolAttend} & masked-in & 9.7 & 51.8 \\[3pt]
			\textbf{CNN-PoolAttend} & masked-out & 5.5 & 11.9 \\
			\bottomrule
		\end{tabularx}
\end{table*}
\subsubsection*{Which architectures and localisation methods perform best when trained as VGS models?}
Overall we see that the CNN-Attend and CNN-PoolAttend architectures outperform the PSC and CNN-Pool architectures and that the masked-in and attention localisation methods outperform the other localisation approaches.
This is the first time mask-based localisation has been incorporated with VGS modelling. We see in oracle localisation that this approach outperforms its closest competitor by more than 10\% absolute in accuracy.
In contrast to the Grad-CAM and score aggregation localisation methods which are tied to a particular model structure, the mask-based localisation approaches have the benefit that they can be applied to any model architecture.
When using BoW supervision, the best performance is always achieved with the CNN-Attend model employing masked-in prediction, while when using visual supervision, attention-based localisation works best on two of the three tasks. 
We speculate that Grad-CAM performs poorly (Tables~\ref{tbl:oracle_localisation_evaluation}, \ref{tbl:actual_localisation_evaluation},~\ref{tbl:spotting_localisation_evaluation}, row 2) across all localisation tasks, regardless of the form of supervision, because it tries to identify the parts of the input that will cause a large change in a particular output unit. In single-label multi-class classification, for which Grad-CAM was developed, a higher probability for a particular output implies lower probabilities for others. But this is not the case for multi-label classification, as used here, and the gradients of multiple words could affect the output, which would be captured in the gradients. While the score aggregation method performs relatively well across all localisation tasks when bag-of-words supervision is available during training, we observe that the performance drops 
when we move to visual supervision (Tables~\ref{tbl:oracle_localisation_evaluation}, \ref{tbl:actual_localisation_evaluation},~\ref{tbl:spotting_localisation_evaluation}, row 1).

\subsubsection*{What penalty do we pay for using visual supervision instead of BoW supervision?}
This is an important question since BoW supervision can be seen as a setting where we have access to a perfect visual tagger:
BoW labels are still weak (they do not contain any location information), but all the words in the bag are related to the image (according to a caption provided by a human).
By comparing the ``Visual'' and ``Bag-of-words'' columns in Tables~\ref{tbl:oracle_localisation_evaluation}, \ref{tbl:actual_localisation_evaluation}, and~\ref{tbl:spotting_localisation_evaluation}, we see that there is a large drop in all cases in performance when moving from idealised supervision to visual supervision. For instance, for actual localisation, the best approach drops from around 80\% to around 25\% in $F_1$.
Improving visual tagging could therefore lead to substantial improvements for keyword localisation using VGS models.
However, it should be noted that although the BoW gives an upper bound for VGS performance, this could be a high upper bound: a perfect visual tagger might detect objects that humans do not describe. In other cases, it might not have visual support for a keyword that might occur in an utterance (for example, \textit{camera} in ``a girl is posing in front of the camera").

As a sub-research question, we note that improving keyword localisation for BoW-supervised models is in itself a useful endeavour. 
BoW labels can be easier to obtain than transcriptions in many cases, see for example~\cite{van2022}.
Localisation with BoW supervision was also the main goal of the original PSC study~\cite{palaz2016}.
Comparing the BoW models, we see that our CNN-Attend model with masked-in localisation gives a substantial improvement over the PSC approach on all three tasks, with absolute improvements in accuracy and $F_1$ scores of more than 10\% absolute in all three cases.
\begin{table}[!t]
	\centering
	\caption{{
			Actual keyword localisation precision, recall and $F_1$ scores for VGS models.
			Keyword detection scores are given in parentheses; since actual localisation involves first doing detection and only then locating words, the detection scores serve as upper bounds on localisation performance.
	}}
	\label{tbl:actual_upperbound}
	\begin{tabular}{rllccc}
		\toprule
		& Architecture                       & Localisation               & $P$    & $R$    & $F_1$  \\
		\midrule
		\multirow{2}{*}{\ii{3}} & \multirow{2}{*}{CNN-Attend}        & \multirow{2}{*}{attention} & \bf 28.8   & \bf 22.4   & \bf 25.2   \\
		&                                    &                            & (38.9) & (28.2) & (32.7) \\[3pt]
		\multirow{2}{*}{\ii{4}} & \multirow{2}{*}{{CNN-Attend}}      & \multirow{2}{*}{masked-in} & 28.3   & 22.2   & 24.9   \\
		&                                    &                            & (38.9) & (28.2) & (32.7) \\[3pt]
		\multirow{2}{*}{\ii{7}} & \multirow{2}{*}{{CNN-Pool-Attend}} & \multirow{2}{*}{masked-in} & 21.9   & 15.3   & 18.0   \\
		&                                    &                            & (35.1) & (22.6) & (27.5) \\
		\bottomrule
	\end{tabular}
\end{table}

\subsubsection*{Limits on localisation performance from first-pass detection}
When doing actual keyword localisation (Table~\ref{tbl:actual_localisation_evaluation}), we are limited by the first detection pass which only considers keywords above a threshold ($\theta = 0.5$).
Since we cannot locate words that are not correctly detected, the detection scores are upper bounds on the localisation performance that can be achieved.

To quantify the limits on localisation performance for a subset of the VGS models, Table~\ref{tbl:actual_upperbound} gives the recall, precision, and $F_1$ scores that were achieved together with the keyword detection scores (given in parentheses).
(One way to see the scores in parentheses is in analogy to the oracle results in Table~\ref{tbl:oracle_localisation_evaluation}: for oracle localisation, we assumed that detection was perfect and only considered localisation; now we consider localisation to be perfect and consider detection performance.)
We see in all cases that the first detection pass impairs localisation. For recall, we are within roughly 6\% of the upper-bound due to detection errors, while for precision, we are roughly within 10\%.
Future work could consider an approach where detection and localisation are performed using different models tailored to these distinct tasks.

\subsubsection*{Limits on localisation from first-pass keyword spotting}
In a similar way, the keyword spotting localisation results ( Table~\ref{tbl:spotting_localisation_evaluation}) are limited by the first ranking pass as performed in standard keyword spotting.
I.e., it is impossible to correctly locate a keyword if the first pass incorrectly ranked an utterance at the top, but the utterance does not actually contain the given keyword.
Here Table~\ref{tbl:spotting_upperbound} shows the achieved $P@10$ localisation scores together with the upper-bound keyword spotting $P@10$ (in parentheses) for a subset of the VGS models.
We also include the $P@N$ scores here: achieved (and upper-bound).
We again see that localisation performance is limited: even for the best approach (CNN-Attend with attention) only about 10\% and 5\% of $P@10$ and $P@N$ localisation mistakes are not the cause of ranking errors.
Improvements in first-pass keyword spotting could therefore again potentially improve subsequent localisation.

\begin{table}[!t]
	\centering
	\caption{{
			Keyword spotting localisation scores in terms of $P@10$ and $P@N$ for VGS models.
			The corresponding keyword spotting scores (ignoring localisation) are given in parentheses; these are upper bounds on localisation performance.
	}}
	\label{tbl:spotting_upperbound}
	\begin{tabularx}{1\linewidth}{@{}rllCC}
		\toprule
		& Architecture & Localisation & $P@10$ & $P@N$ \\
		\midrule
		\ii{3} & {CNN-Attend}      & attention & \bf 32.1 (43.7) & \bf 24.9 (30.1) \\
		\ii{4} & {CNN-Attend}      & masked-in & 21.9 (43.7) & 19.5 (30.1) \\
		\ii{7} & {CNN-Pool-Attend} & masked-in & \hphantom{0}9.7 (31.6) & 10.4 (24.4)  \\
		\bottomrule
	\end{tabularx}
\end{table}

\subsubsection*{Effect of intermediate max pooling on localisation performance}
Another way to analyse the different architectures and localisation methods is by looking at how their choice of encoder structure affects their localisation capabilities. Each architecture uses either CNN or CNN-Pool. In the CNN-Pool encoder, each convolutional layer operation (except the last one) is followed by a max pooling operation. The CNN encoder type doesn't perform such intermediate pooling operations. Intuitively, the max pooling operation calculates and returns the most prominent feature of its input by downsampling along the time axis. While such operation has been shown to help with better word detection~\cite{kamper2017a}, it hurts localisation performance in both the VGS and BoW supervision settings. We demonstrate this by comparing localisation scores of our two best localisation methods for the two encoder types.
As shown in Tables~\ref{tbl:oracle_localisation_evaluation}, \ref{tbl:actual_localisation_evaluation}, and~\ref{tbl:spotting_localisation_evaluation}, %
all architectures with the CNN encoder type that uses the attention and masked-in localisation methods outperforms the ones with the CNN-Pool encoder type on the three tasks. This behaviour is expected since the max pooling operation employed by the CNN-Pool makes it more challenging to backtrace each output prediction to its most relevant filter.
\begin{figure}
	\centering
	\includegraphics[width=0.95\linewidth]{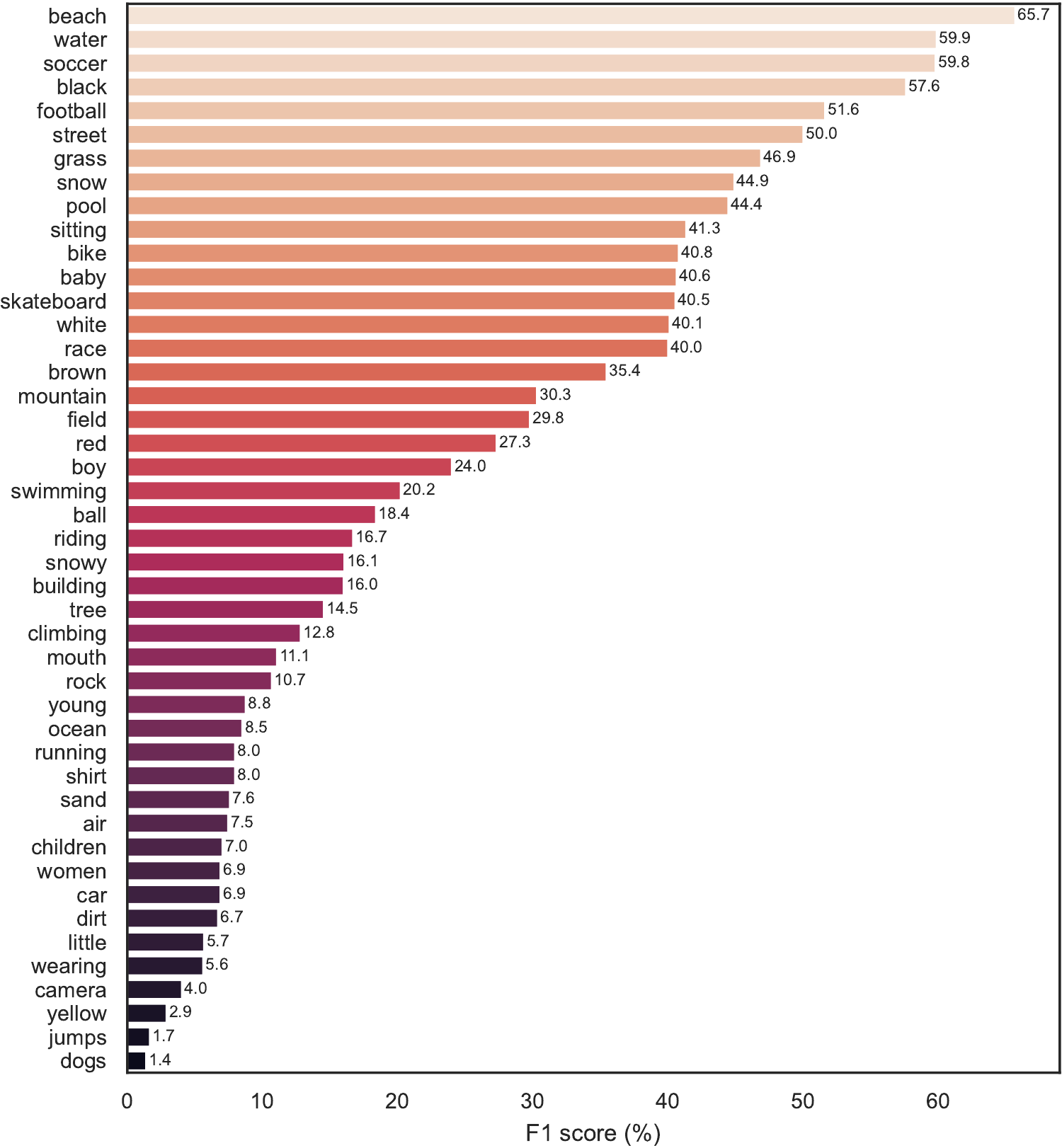}
	\caption{%
		Actual localisation performance measured in terms of $F_1$ score reported per keyword.
		We include only those keywords (45 out of 67) for which at least one utterance was detected, that is, the detection score is greater than the threshold $\theta$ of 0.5).
		The results are reported for the CNN-Attend architecture and the masked-in localisation method.
	}
	\label{fig:keyword-evaluation-f1}
\end{figure}
\subsubsection*{Per keyword performance}
We investigate the localisation performance for each individual keyword in the vocabulary.
We report the results in the actual keyword localisation setup, that is,
for each keyword $w$, we first select the utterances that have a detection score over the threshold $\theta$ and
then check if the localisation is correct by verifying whether the location $\tau_w$ of the maximum localisation score falls into a segment corresponding to the query keyword $w$.
The results are presented in Figure \ref{fig:keyword-evaluation-f1}.

We report performance for 45 out of the total 67 keywords in the vocabulary, since for the remaining 22 keywords no utterance had a detection score over the imposed threshold $\theta$ of 0.5;
when no utterance is detected we cannot localise it---the precision is not defined, while the recall is 0.
We observe that the actual localisation performance varies strongly across the keywords in the vocabulary (from 1.4\% for \textit{dogs} to 65.7\% for \textit{beach}) suggesting that the methods are very keyword-dependent.
Our assumption is that the keyword performance is influenced by how well each word is visually grounded.

\begin{figure}[!t]
	\centering
	\includegraphics[width=0.95\linewidth]{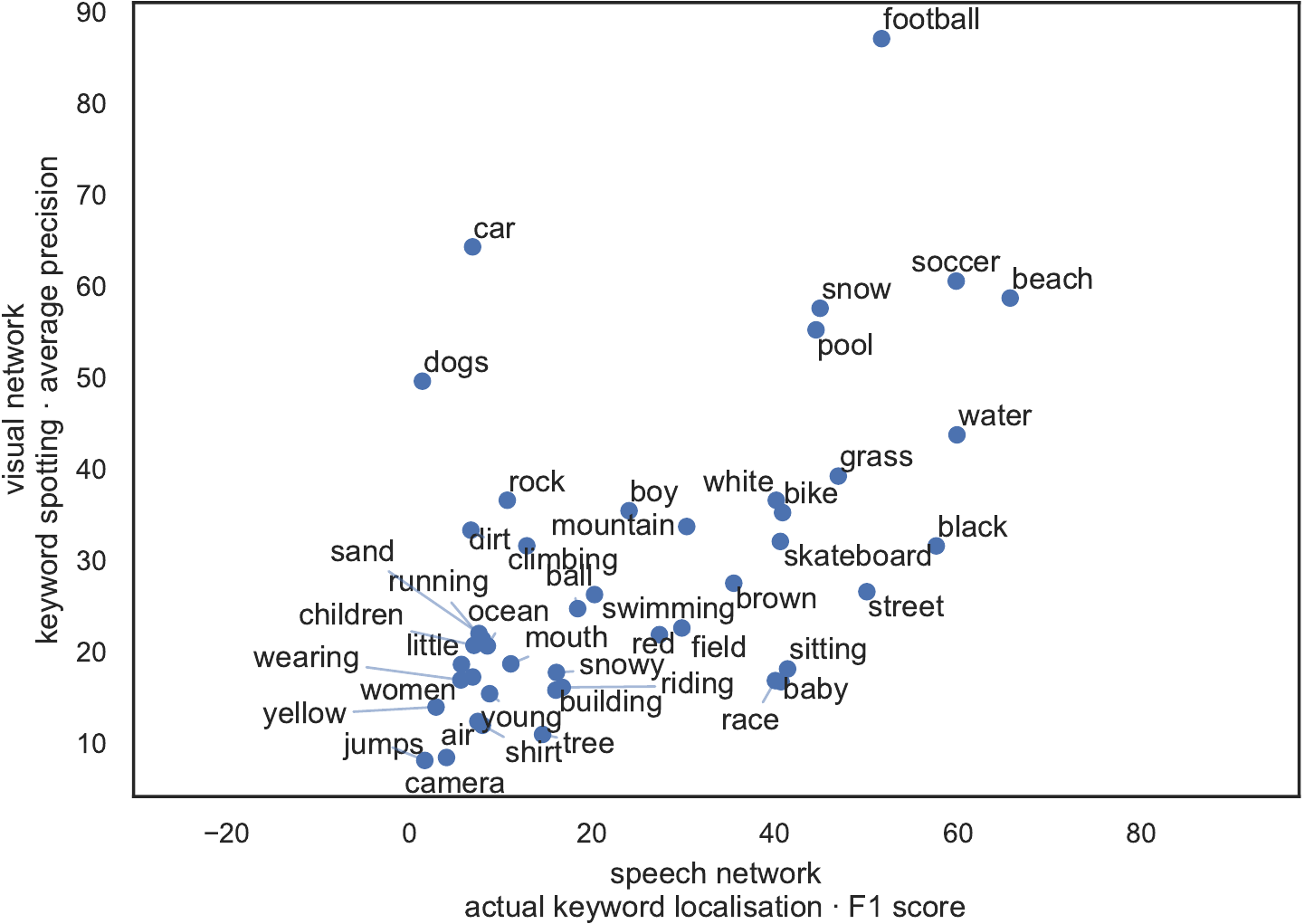}
	\caption{%
			Localisation performance ($F_1$) of the speech network versus spotting performance (average precision) of the visual network for each of the 44 keywords that were included in Figure \ref{fig:keyword-evaluation-f1}.
			The results are reported for the CNN-Attend architecture and the masked-in localisation method.
		}
	\label{fig:keyword-evaluation-loc-vs-vis}
\end{figure}

To test this hypothesis we look at how well the localisation performance correlates with the performance of the visual network; see Figure \ref{fig:keyword-evaluation-loc-vs-vis}.
We observe that indeed there is a dependency between the two systems:
keywords like \textit{camera}, \textit{air}, and \textit{wearing} understandably have poor visual grounding, which impacts the speech localisation capability;
while keywords like \textit{beach}, \textit{soccer}, \textit{snow} are easily spotted by the visual network and also by the speech-based localisation network.
However, there are some outliers: the keywords \textit{football}, \textit{car}, \textit{dogs} while well visually grounded, suffer when localised.
We investigate these cases in Section~\ref{ssec:qualitative-results} and observe that they are caused by co-occurrences of the keywords with other words: \textit{football}--\textit{player}, \textit{car}--\textit{race}, \textit{dogs}--\textit{three} (see Table \ref{tbl:most-common-confusions}).
Words that often occur together in the same utterance are very challenging to be discriminated by any localisation method that does not have access to additional information.

\begin{figure}[t]
	\includegraphics[scale=0.5]{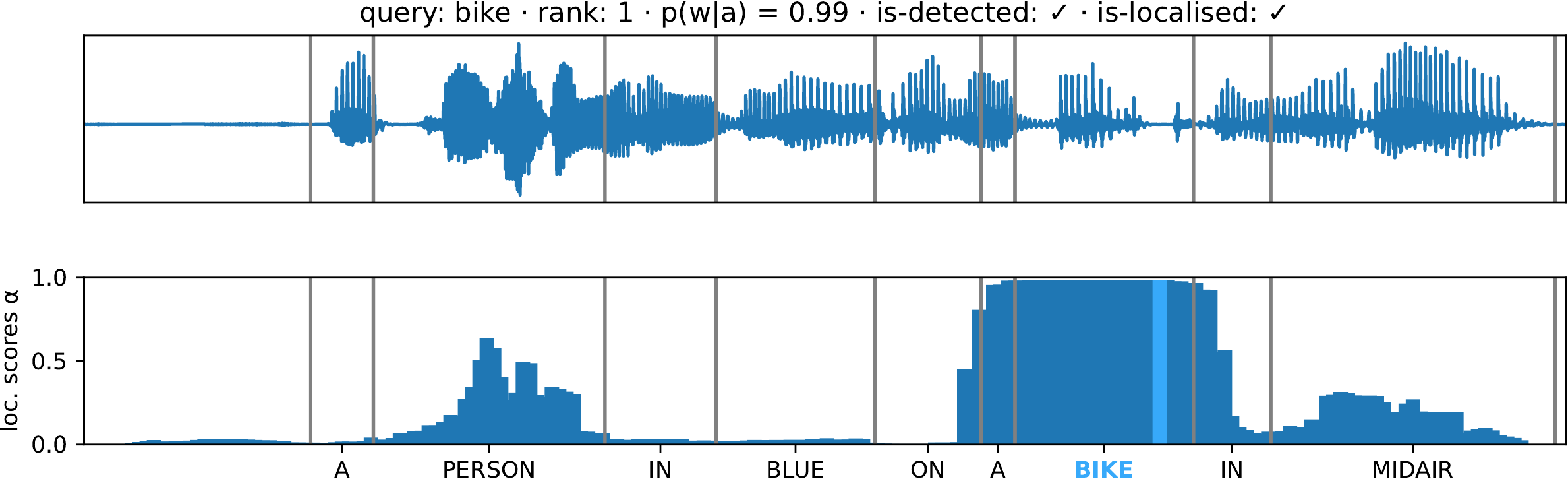} \\[20pt]
	\includegraphics[scale=0.5]{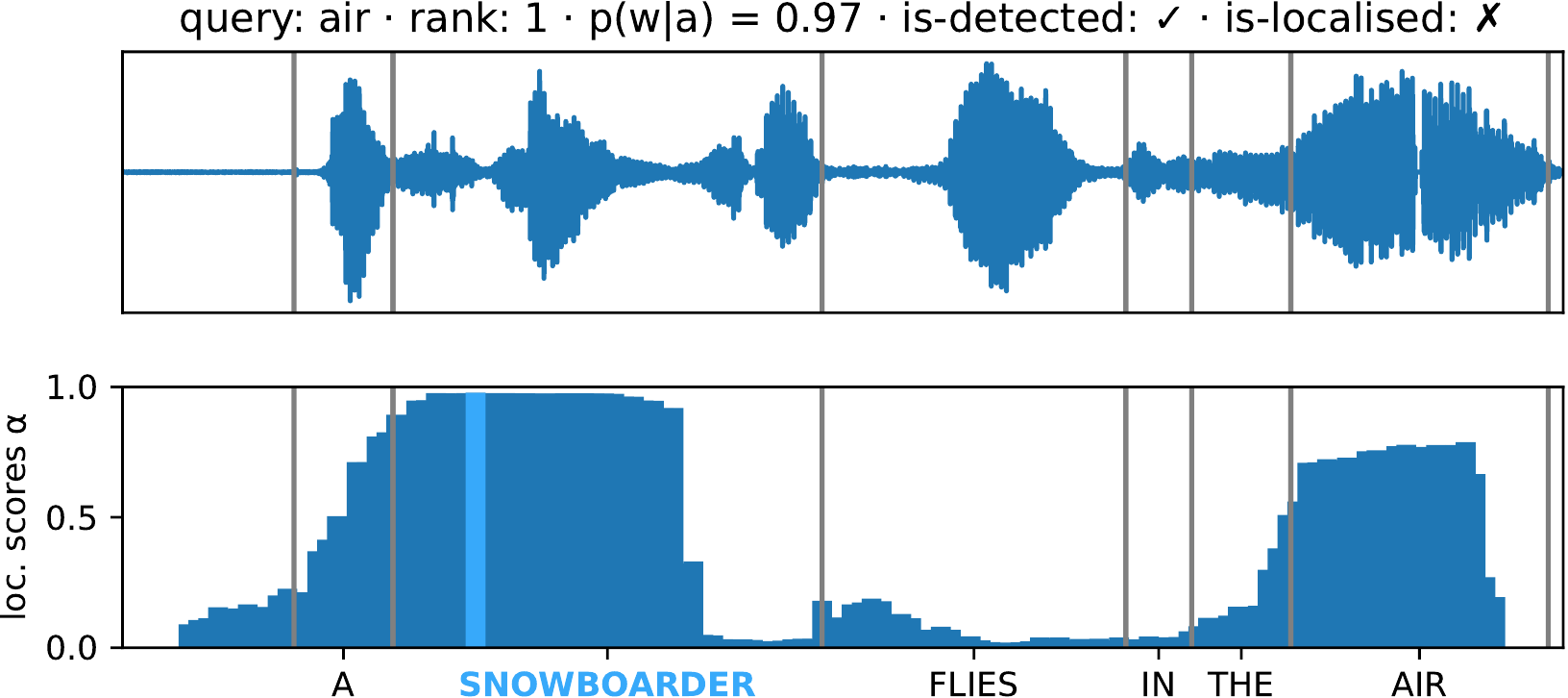} \\[20pt]
	\includegraphics[scale=0.5]{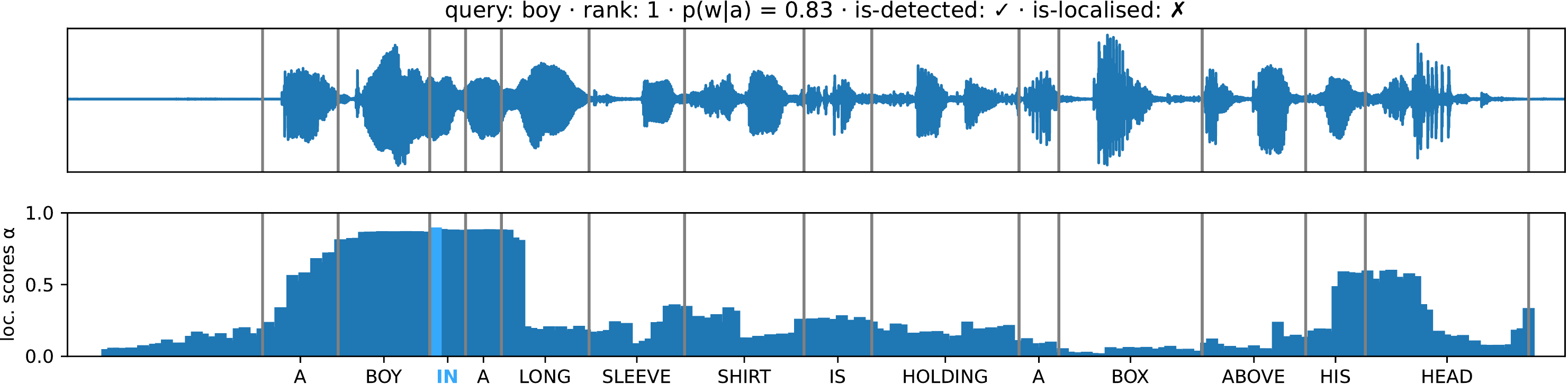} \\[20pt]
	\includegraphics[scale=0.4]{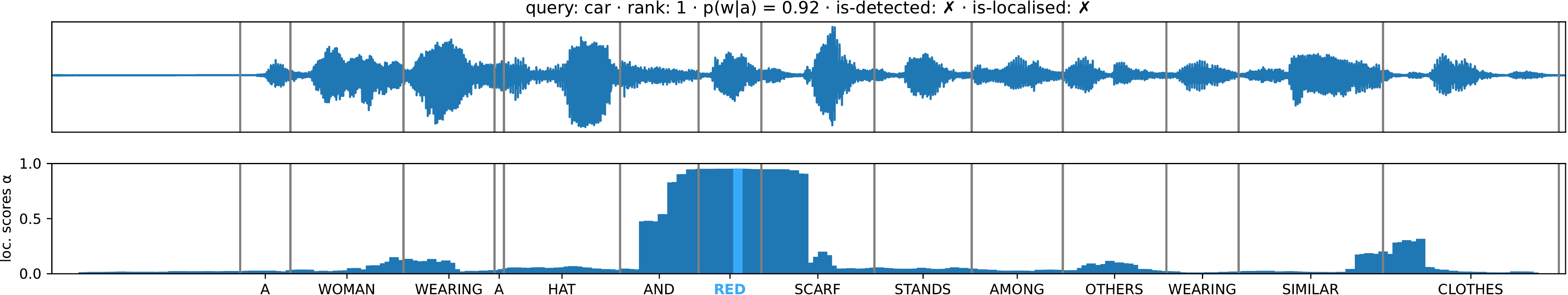} \\[20pt]
	\caption{%
		Examples of predictions for four query keywords (\textit{bike}, \textit{air}, \textit{car}, \textit{boy}) using the CNN-Attend architecture and the masked-in localisation method.
		In each figure we show the input audio signal (top subplot) and the output localisation scores (bottom subplot).
		We also provide the detection score $p(w|\ab)$, whether each keyword was correctly detected or localised, the transcription of the utterance.
		The vertical grey lines denote the word boundaries and the word coloured in light blue shows the top scoring word.\label{fig:qualitative-samples}
	}
\end{figure}

\subsection{Qualitative results}
\label{ssec:qualitative-results}

In this subsection we report concrete examples obtained by manually inspecting the samples and the network's predictions.
\subsubsection*{Qualitative samples}
In Figure \ref{fig:qualitative-samples} we show the top-ranked utterances (based on their detection scores) and their corresponding localisation scores for four keywords:
\textit{bike}, \textit{air}, \textit{car}, \textit{boy}.
For \textit{bike}, we see that both detection and localisation are correct,
both achieving high scores close to the maximum value of 1.
In the case of the second example where the query keyword is \textit{air},
the top-ranked utterance does contain the query (hence the detection is correct),
but the word \textit{snowboarder} produces a higher localisation score than \textit{air}, yielding an incorrect localisation.
As we will see in the next set of experiments, \textit{air} is often confused with \textit{snowboarder} due to the fact they frequently co-occur.
The third example highlights another common confusion---\textit{red} is often predicted when the query is \textit{car}---%
in this case probably exacerbated by the phonetic similarity between \textit{red scarf} and the more common phrase \textit{red car}.
The final example shows a case of a narrowly missed localisation: while the word \textit{boy} achieves a large score, the maximum value is obtained at the boundary with the word \textit{in}.
This incorrect localisation could be due to the inherent imprecision of the forced alignment procedure, which is used to assign location information to the words in the transcription.

\subsubsection*{Most common confusions}
For each keyword in the vocabulary, we find which words in the audio utterance they are most commonly confused with.
To generate the results, we take the following steps.
(1) Given a keyword, we sort the utterances in decreasing order of their detection (per-utterance) score.
(2) For the each of the top twenty utterances, we choose the predicted word as the word that contains the maximum of the localisation scores $\alpha_{w,1}, \dots, \alpha_{w,\temp}$.
(3) We report the top five most commonly predicted words together with their counts.
Table \ref{tbl:most-common-confusions} shows the results for a subset of the keywords.

\begin{table}
		\centering
		\caption{%
			The most commonly located words for a subset of sixteen query keywords.
			For each keyword, we consider the top 20 utterances based on the detection scores and in each utterance we find the predicted word based on the localisation scores.
			We report the most common 5 located words.
			We use the CNN-Attend architecture and the masked-in localisation method.
			The 
			`--' symbol denotes spoken noise or words that are missing from the phonetic dictionary of the forced aligner.
		}
		\label{tbl:most-common-confusions}
		\begin{tabular}{@{}llrlrlrlrlr}
			\toprule
			Query & \multicolumn{10}{l}{Top 5 located words and their counts} \\
			\midrule
			air        & snowboarder & 11 & bike        & 2 & --            & 2 & snowboard   & 1 & skiing      & 1 \\
			ball       & soccer      &  9 & ball        & 5 & player        & 2 & basketball  & 2 & colorful    & 1 \\
			black      & black       & 18 & dog         & 2 &               &   &             &   &             &   \\
			car        & race        &  7 & red         & 4 & racetrack     & 2 & raises      & 2 & car         & 2 \\
			dogs       & three       &  8 & two         & 6 & dog           & 2 & puppy       & 1 & dogs        & 1 \\
			face       & climbing    &  7 & rock        & 5 & goes          & 1 & rocks       & 1 & blonde      & 1 \\
			football   & football    & 15 & player      & 4 & players       & 1 &             &   &             &   \\
			ocean      & surfer      & 11 & surfboard   & 4 & wave          & 3 & surfing     & 1 & surface     & 1 \\
			orange     & red         &  9 & wears       & 1 & a             & 1 & retriever   & 1 & guitar      & 1 \\
			pink       & girl        &  8 & young       & 2 & girls         & 2 & dressed     & 1 & one         & 1 \\
			pool       & pool        & 18 & swimming    & 1 & swimmer       & 1 &             &   &             &   \\
			soccer     & soccer      & 20 &             &   &               &   &             &   &             &   \\
			swimming   & pool        & 16 & swimming    & 3 & swim          & 1 &             &   &             &   \\
			tree       & tree        &  9 & trees       & 7 & training      & 1 & climbs      & 1 & --          & 1 \\
			yellow     & yellow      & 17 & bikes       & 1 & bike          & 1 & race        & 1 &             &   \\
			\bottomrule
		\end{tabular}
\end{table}

We observe good matches for some keywords (\textit{black}, \textit{pool}, \textit{soccer}, \textit{tree}),
while others are confused with semantically related words:
\textit{air} $\to$ \textit{snowboarder}; \textit{ocean} $\to$ \textit{surfer}; \textit{ball} $\to$ \textit{soccer}; \textit{swimming} $\to$ \textit{pool}; \textit{face} $\to$ \textit{climbing}, \textit{rock}.
The performance for \textit{car} is surprising: given the fact that \textit{car} has visual support, we would have expected better localisation, instead the word is often confused with \textit{race} and \textit{red}.
Interestingly, the keyword \textit{dogs} is sometimes associated with numerals (\textit{two} and \textit{three});
we believe this happens because we distinguish between singular and plural words (the word \textit{dog} is different from the word \textit{dogs}).
For some keywords representing colour (\textit{black}, \textit{white}, \textit{yellow}) we obtain strong results,
while for others there are more semantic confusions: \textit{pink} $\to$ \textit{girl}; \textit{orange} $\to$ \textit{red}.

\section{Summary}

In this chapter we considered different keyword localisation approaches using VGS models that learn from images and unlabelled spoken captions.
We specifically used a VGS method that tags training images with soft textual labels as targets for a speech network that can then detect the presence of a keyword in an utterance.
We equipped this type of VGS model with localisation capabilities, investigating four localisation approaches.
Our best approach relied on masking: only a portion of the spoken input is passed through the model, and by looking at the change in output probability, a decision can be made regarding the presence of a keyword within the unmasked region of the utterance.
By using a sliding window with a changing width, the best location for a keyword can be identified.
This approach gave a 57\% localisation accuracy (compared to the 46\% achieved with the attention method) in an oracle setting where we assume the system knows that a keyword occurs in an utterance and needs to predict its location and an $F_1$ of 25\% when detection is first performed. However, input masking has a higher computational cost than the attention method.

We showed that the first detection or ranking pass limits the system's performance.
Better performance could be possible by improving the visual tagger since per-keyword performance is strongly correlated with the performance of the visual tagger on that keyword.
We have also shown that many incorrect localisations are caused by predicting semantically related words—seeing what benefit this holds is another avenue to explore in future work (see Section~\ref{sec:conclusion_future_work}).

Up to this point, we have considered a moderately sized set of speech-image pairs where the unlabelled spoken captions are given in English.
In order to apply the proposed approach in a real low-resource setting, we need to investigate how performance is impacted when less data is used.
In an extreme low-resource setting where a language is unwritten, there is also the question of how a written keyword could be used to query the model.
We address this in the next chapter where we look at cross-lingual keyword localisation: where a keyword is given in one (well-resourced) language and then needs to be located in speech from another low-resource language~\cite{kamper2018}.
We conduct these experiments on a real low-resource language, rather than artificially treating English as low-resource.

%% file: yfacc.tex
\chapter{Cross-lingual keyword localisation in a real low-resource setting}
\label{chap:cross_lingual_keyword_localisation}
In the previous chapters we have presented approaches for visually grounded keyword detection and localisation. At the core of approaches presented are existing models~\cite{kamper2017a, kamper2019b} which we augment with four different localisation methods. As mentioned in Chapter~\ref{chap:introduction}, such methods could allow linguists to process and investigate languages that exist only in spoken form. However, the lack of a truly low-resource VGS dataset makes it difficult to correctly evaluate the performance of a VGS system in this scenario---the analyses of the system carried out up to this point are in an artificial setting where English is treated as a low-resource language.

In this chapter we address in full our second research question (Section~\ref{sec:research_questions}): Can we do visually grounded keyword localisation cross-lingually in a real low-resource setting? We move towards a more realistic setting by introducing a multilingual and multimodal dataset that enables training and evaluating of VGS models in a real low-resource setting. Concretely, we collect and release a new single-speaker dataset of audio captions for 6k Flickr8k images in \yoruba (Section~\ref{sec:yfacc})---a real low-resource language spoken in Nigeria. We train an attention-based VGS model where images are automatically tagged with English visual labels and paired with \yoruba utterances (Section~\ref{sec:cross_lingual_tasks}). This enables cross-lingual keyword localisation: a written English query is detected and located in \yoruba speech. The evaluation protocol used to assess the performance of the models on cross-lingual keyword localisation is described in Section~\ref{sec:eval_cross_lingual_localisation}. We present the architecture and implementation details in Section~\ref{sec:architecture_implementation_details_cross_lingual}. Finally in Section~\ref{sec:exp_results}, we present the quantitative and qualitative analyses of our visually grounded cross-lingual system.
\begin{tcolorbox}[width=\linewidth, colback=white!95!black, boxrule=0.5pt]
	\small
	\textit{This study was reported in IEEE SLT:}\\
	K. Olaleye, D. Oneață, and H. Kamper, ``YFACC: A \yoruba speech--image dataset for cross-lingual keyword localisation through visual grounding,'' \textit{IEEE Spoken Language Technology (SLT)}, 2023.
	
\end{tcolorbox}

\section{Relation to previous work}
\label{sec:previous_work_chapter5}
The goal of this section is to relate work done in this chapter to prior work in the literature. In particular, the task attempted and enabled by the new \yoruba speech--image dataset (introduced in Section~\ref{sec:yfacc})---cross-lingual keyword localisation---is a new step from the two previous variants explored in the literature: (i) monolingual keyword localisation (Chapter~\ref{chap:methods-for-keyword-localisation}, ~\cite{olaleye2022}); and (ii) cross-lingual keyword detection~\cite{kamper2018}.

As defined in Chapter~\ref{chap:introduction}, models that learn from images and their spoken captions are known as VGS models \cite{driesen2010, synnaeve2014b, harwath2015, harwath2016,  harwath2017, harwath2018a, harwath2018b, eloff2019, harwath2019a, harwath2019b, nortje2020}.
While previous work on VGS has shown promise for performing a number of downstream tasks in low-resource conditions, very few studies validate the approach in a realistic setting: most papers use unlabelled English speech or data from another high-resource language~\cite{harwath2018b, ohishi2020a} to simulate a low-resource setting.

\begin{figure}[!b]
	\centering
	\includegraphics[width=0.90\columnwidth]{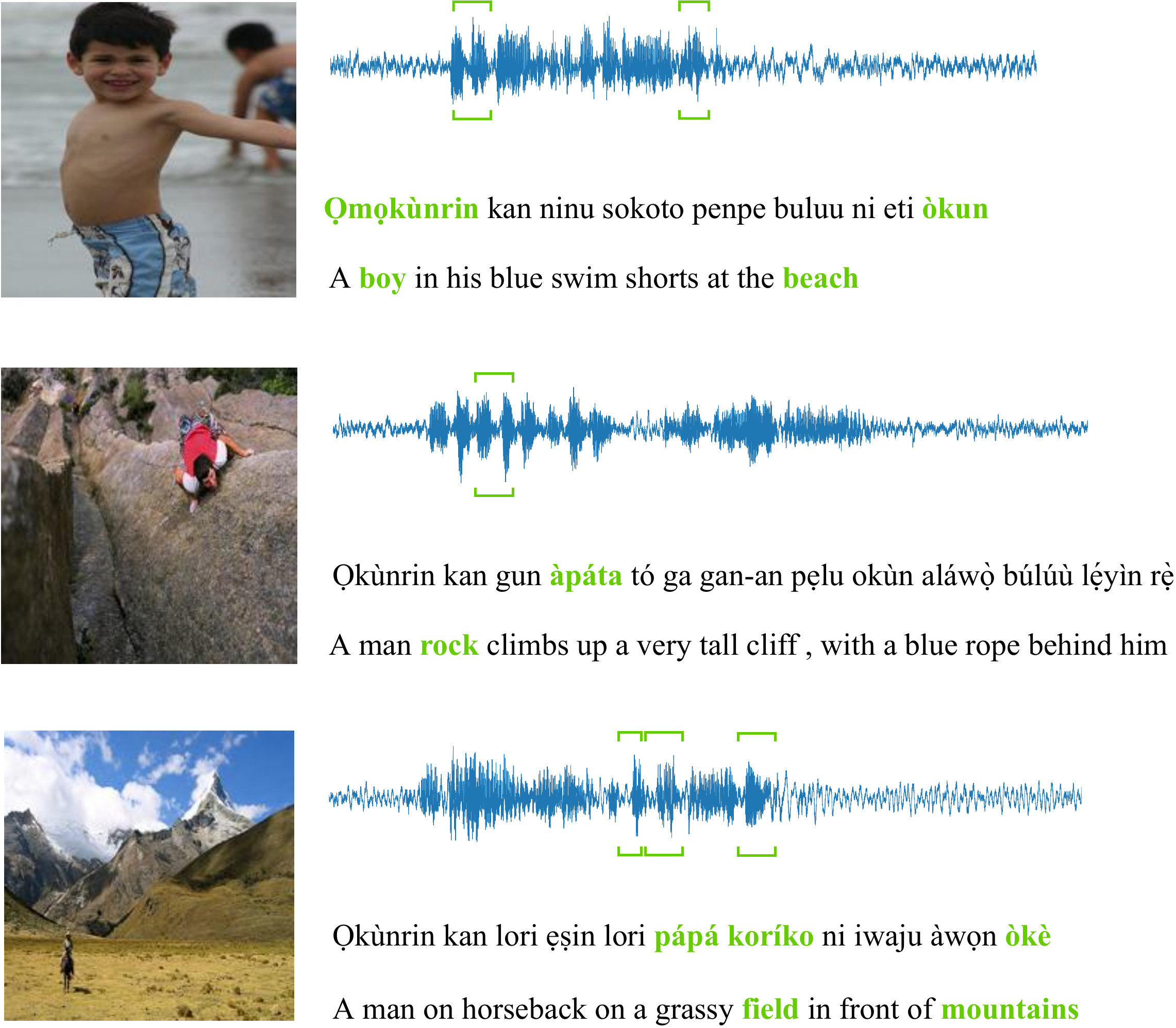}
	\caption{%
		Samples from the new
		\yoruba Flickr Audio Caption Corpus (YFACC):
		images from the Flickr8k dataset (left) are paired with \yoruba audio captions (right),
		which are based on the original English captions. 
		For a subset of keywords (green) we provide temporal alignments for evaluation.
	}
	\label{fig:sample_datasets}
\end{figure}

To
address this shortcoming in VGS research,
we introduce a new speech--image dataset in a low-resource language, \yoruba.
The \yoruba language is one of the three official languages in Nigeria, and even though 
it is spoken natively by more than 44M speakers,
there are only a handful of high-quality datasets available in this language
\cite{vanniekerk2014a,gutkin2020,meyer2022,ogayo2022}.
Our dataset is built by augmenting the Flickr8k dataset~\cite{rashtchian2010,hodosh2015} with spoken captions: from a single \yoruba speaker, we record 6k \yoruba utterances corresponding to translations of the original English image captions.
We also provide manual alignments with start and end timestamps for 67 selected keywords in each of the 500 test utterance.
We call the resulting dataset the \yoruba Flickr Audio Captions Corpus (YFACC).
Examples of samples from YFACC are shown in Figure~\ref{fig:sample_datasets}.
A major advantage of extending Flickr8k is that it allows reusing existing annotations (English recordings \cite{harwath2015}, translations in other languages \cite{elliott2016,li2016}), which could be used to construct 
richer and potentially more interesting tasks in future work.

Prior work has considered cross-lingual keyword spotting: retrieving utterances, in one language, containing
a given keyword in another language. Specifically, Kamper and Roth~\cite{kamper2018} investigated whether visual grounding can be used for cross-lingual keyword spotting. The authors claimed that such a cross-lingual keyword spotting system could enable searching through speech in a low-resource language using text queries in a high-resource language. To validate their claim, they used English speech (artificially treated as low-resource) with German queries (high-resource language). In particular, they used a German visual tagger to provide keyword labels for each training image, and then trained a neural network to map English speech to German keywords without seeing parallel speech-transcriptions or translation. The authors further showed that most erroneous retrievals contain equivalent or semantically relevant keywords. In this chapter, we directly extend their work to include keyword localisation. Furthermore, we consider a real low-resource setting rather than artificially treating English as a low-resource language.

To our knowledge, we are the first to attempt cross-lingual keyword localisation. While more challenging, this task is also potentially more impactful in developing applications for linguists.
Our approach applies the model architecture described in Section~\ref{subsec:exp_setup} (which used existing architectures~\cite{kamper2017a, kamper2019b} as their base), and the localisation method described in Section~\ref{ssec:attention} to the cross-lingual setting:
images are automatically tagged with English visual labels, which serve as targets for an attention-based model that takes \yoruba speech as input.
This enables the \yoruba speech model to be queried with English text.

\section{YFACC: The Yor\`ub\'a Flickr Audio Caption Corpus}
\label{sec:yfacc}

We introduce the new \yoruba Flickr Audio Caption Corpus (YFACC).
The YFACC dataset extends the Flickr8k image--text dataset~\cite{rashtchian2010,hodosh2015} to \yoruba with three modalities:
\yoruba translations of 6k of its captions;
corresponding spoken recordings of these translations;
and temporal alignments of 67 \yoruba keywords for a subset of 500 of the captions.

\subsection{Data collection}
We
selected a random set of 6k images from the Flickr8k dataset
and for each image we picked
one out of the
five associated captions.
The resulting English captions were manually translated to \yoruba by two native speakers.
Partial diacritics were added to aid the recording process.
All translations were recorded from
a single male native speaker.
A primary motivation for collecting speech from only a single speaker is that this is representative of a real language documentation setting, where a linguist will only have access to a small number of speakers.
Recordings were made with a BOYA BY-M1DM dual omni-directional lavalier microphone using
a sampling rate of 48 kHz and two channels.

The 6k utterances were split into train (5k), development (500) and test (500) sets.
For the 500 test utterance, a native speaker produced
temporal alignments for a set of 67 \yoruba keywords corresponding to the 67 English keywords 
presented in Table~\ref{tbl:vocabulary_list}.
These keywords were originally presented in~\cite{kamper2019b}, and were selected such that they are visually groundable and provide high inter-annotator agreement,
making them suitable to evaluate the localisation algorithm. We prepare the alignments with \texttt{Praat}~\cite{boersma2001}. 
Alignments include the start time, end time, and duration of each of the 67 keywords. Table~\ref{tbl:yoruba_corpus_overview_quantitative} presents details about the quantitative characteristics of YFACC and how they compare to other existing \yoruba datasets (more details on these other datasets in Section~\ref{sec:relationship}).
Samples from YFACC are shown in Figure \ref{fig:sample_datasets}.

\subsection{Cross-linguistic differences}
\label{ssec:yfacc-cross-differences}

\yoruba is a very different language from English both linguistically and culturally.
In what follows we discuss some of the specific aspects
as they relate to
YFACC.

\subsubsection*{Tone marks}
In \yoruba, pronunciation and writing involves using three types of tone marks \cite{wood1879}:
\textit{d\`o}, a falling tone (grave accent);
\textit{re}, a flat tone (no accent);
\textit{m\'i}, a rising tone (acute accent). 
These marks are applied to the top of the seven \yoruba vowels (\textit{a, e, \d{e}, i, o, \d{o}, u}), and the four nasal vowels (\textit{\d{e}n, in, \d{o}n, un}).
The tone marks are key to understanding \yoruba, since without them many words become ambiguous.
E.g.\ 
\textit{okun} can stand for \textit{\`okun} (beach), \textit{ok\`un} (rope) or \textit{okun} (strength),
and \textit{\d{o}k\d{o}} can stand for \textit{\d{o}k\d{\`o}} (car), \textit{\d{o}k\d{\'o}} (hoe) or \textit{\d{o}k\d{o}} (husband). 
In practice including all the marks when writing is a time consuming activity.
So for our dataset we provide diacritics
only for the 67 keywords,
since only these are used for evaluation.
However, in the future it might be possible to generate the marks for the remaining words using an automated approach~\cite{orife2020}.

\subsubsection*{Keywords across the two languages}
Even if YFACC is built by translating the Flickr8k
dataset,
a keyword that appears in the English caption is not guaranteed to appear in the corresponding \yoruba translation, and vice versa.
For example, 
\textit{A blonde woman appears to wait for a ride} is translated to \yoruba as \textit{Ob\`inrin bilondi nd\'ur\'o de \d{o}k\d{\`o}}
(which back-translates to \textit{A blonde woman \underline{stands} for a \underline{car}}), making the keywords \textit{stands} and \textit{car} appear in the \yoruba caption.
We quantify the common appearances of the keywords across the two languages by computing the Cohen's Kappa ($\kappa$) correlation coefficient. Each sample in the test set is represented by two sentences: one in English and a corresponding one in \yoruba. Given an English word and a \yoruba one, we verify if each of them appears in the corresponding sentences; for example:\vspace{12.6pt}
\\
\begin{tabular}{ccc}
	id & blue & \`okun \\
	s01 & 1 & 0 \\
	s02 & 0 & 1 \\
	s03 & 1 & 1 \\
	... &... &...
\vspace{12.6pt}	
\end{tabular}
\\
(with 1 denoting presence and 0 denoting absence).

\begin{figure}
	\centering
	\includegraphics[width=1.0\textwidth]{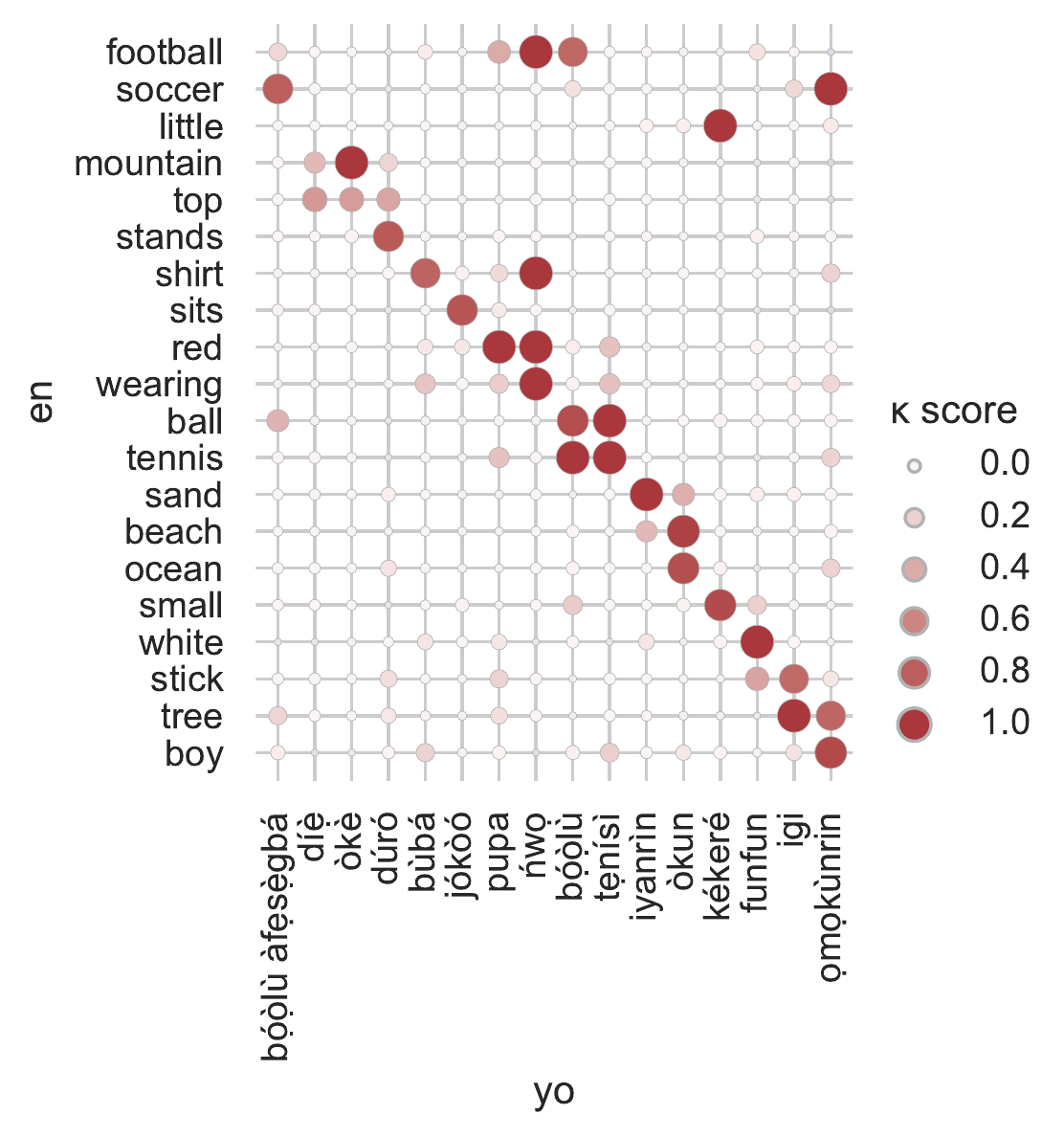}
	\caption{%
		Co-occurrences of English--\yoruba keyword pairs on the 500 test utterances in YFACC computed as normalised Cohen's $\kappa$ score.
		For readability, we pick keywords 
		from the 67-word vocabulary.
	}
	\label{fig:keyword-correlation}
\end{figure}

We then take the two vectors of appearances, the one corresponding to the English word, $[1, 0, 1, \ldots]$, and the one corresponding to the \yoruba word, $[0, 1, 1, \ldots]$, and compute their agreement using a normalized variant of the Cohen’s Kappa score~\cite{cohenkappa2022} (described below). Intuitively, the larger this score the more likely it is that the pair of words occurs in the same set of samples as shown in Figure \ref{fig:keyword-correlation}.

The Cohen's Kappa coefficient ($\kappa$) is a metric that measures the agreement of two binary vectors by correcting for random agreement. But the $\kappa$ score has an upper bound based on the difference in marginal probabilities (in our case, the number of occurrences of a word in the test sentences): if one word (in one of the languages) appears more often than the other (in the other language), then the maximum achievable score is less than one. Since we compare across words and languages (implying different number of occurrences), we normalise the usual coefficient by the maximum achievable value, resulting in what we refer to as the normalised Cohen $\kappa$ score:

\begin{equation}
	\kappa_\text{norm} = \frac{\kappa}{\kappa_\text{max}}
\end{equation}
where
\begin{equation}
	\kappa = \frac{p_o - p_e}{1 - p_e}
\end{equation}
and
\begin{equation}
	\kappa_\text{max} = \frac{p_\text{max} - p_e}{1 - p_e}
\end{equation}
with
\begin{itemize}
	\item $p_o$ being the relative observed agreement
	\item $p_e$ being the probability of agreement by chance
	\item $p_\text{max}$ being the maximum achievable probability.
\end{itemize}

Intuitively, if
	an English and a \yoruba word tend to co-occur (i.e., appear in the corresponding English--\yoruba translation samples, as is the case with the \textit{ride}--\textit{\d{o}k\d{\`o}} pair in the previous example),
	that pair will yield a large score.

Figure~\ref{fig:keyword-correlation} presents the plot of the correlation between the English-\yoruba pair on the 500 test utterances in YFACC. We observe a strong correlation along the diagonal of the plot, suggesting that English keywords do correlate with their translations, but there are also some exceptions.
Some \yoruba words have multiple meanings, so they co-occur with more than a single English word,
e.g.\ \textit{\`okun}, which means both \textit{beach} and \textit{ocean}.
Conversely, some English words have multiple meanings:
\textit{football} is not just the sport, but also the \textit{ball};
\textit{little} refers not only to size, but also to age (\textit{little boy}; \textit{\d{o}m\d{o}k\`unrin k\'eker\'e}).
Finally, some correlations appear just because words co-occur together in both languages: \textit{\'nw\d{o}} (\textit{wearing}) correlates with \textit{shirt}, \textit{red}, and \textit{football}. 

\subsubsection*{Cultural differences}
Starting from an English dataset may induce a mismatch regarding the cultural aspects, as discussed in \cite{hershovich2022}.
Most of the 67 selected keywords (Table~\ref{tbl:vocabulary_list}) in Flickr8k pertain to general cross-cultural concepts (such as \textit{children}, \textit{young}, \textit{swimming}),
but there are some of the English words that have no indigenous \yoruba translation.
E.g.\ words like  \textit{jacket} or \textit{tennis}
are usually translated based on their English pronunciation by adding extra marks to the English word: \textit{j\'ak\d{\`e}t\`i} and  \textit{t\d{e}n\'is\`i}.
Perhaps the only keyword that presents little interest to the typical Nigerian is the English word \textit{skateboard}.

\subsection{Relation to Flickr8k and other \yoruba datasets}
\label{sec:relationship}
In Section~\ref{sec:dataset} we gave details about the Flickr8k dataset, a bimodal (image and text) dataset. We also gave details about the Flickr Audio Caption Corpus (FACC), introduced by Harwath and Glass \cite{harwath2015}, which is a previous extension of the Flickr8k dataset consisting of 36 hours of non-silent
English speech, for each of the text captions in the Flickr8k dataset. As mentioned in the introduction of this section, YFACC extended the Flickr8k dataset with \yoruba translations of 6k of its captions, spoken utterances in \yoruba, and temporal alignments of 67 \yoruba keywords for a subset of 500 of the Flickr8k captions.

Compared to previous \yoruba datasets,
ours is different in that
it has a richer set of modalities, as shown in Table \ref{tbl:yoruba_corpus_overview_modalities}.
By extending Flickr8k we are able to reuse its existing annotations and build a multimodal and multilingual dataset, which, similarly to FACC, allows for a wide variety of tasks to be defined.
In this chapter we specifically consider the task of visually-grounded cross-lingual keyword localisation.
Example tasks that can be considered in the future include speech-to-speech machine translation or grounding visual concept in images from \yoruba speech.
The fact that the recordings are done by a single speaker and at high quality (48 kHz), also enables the future development of text-to-speech technology for \yoruba.

\begin{table*}
	\centering
	\caption{%
		A comparison of the open-source \yoruba speech corpora in terms of the available modalities.
		The new 
		YFACC datasets provides alignments for a subset of 67 keywords.
	}
	\begin{tabular}{@{}l@{}lcccccc@{}}
		\toprule
		&  & \multicolumn{5}{c}{Modalities}                                         &  \\
		\cmidrule(lr){3-7}
		Dataset                                   &  & Speech & Text (yo) & Text (en) & Images & Alignments    \\
		\midrule
		van Niekerk et al. \cite{vanniekerk2014a} &  & \cmark & \cmark    &           &        &              \\
		Gutkin et al. \cite{gutkin2020}           &  & \cmark & \cmark    &           &        &              \\
		Meyer et al. \cite{meyer2022}             &  & \cmark & \cmark    &           &        &              \\
		Ogayo et al. \cite{ogayo2022}             &  & \cmark & \cmark    &           &        &              \\
		YFACC (ours)                              &  & \cmark & \cmark    & \cmark    & \cmark & partial      \\
		\bottomrule
	\end{tabular}
	\label{tbl:yoruba_corpus_overview_modalities}
\end{table*}

\begin{table*}
	\centering
	\caption{%
		A comparison of the open-source \yoruba speech corpora in terms of the quantitative characteristics. 
	}
	\begin{tabular}{@{}l@{}lcrrrc@{}}
		\toprule
		&  & \multicolumn{1}{c}{Speakers}         & \multicolumn{1}{c}{Duration}        & \multicolumn{1}{c}{Utterances}       & \multicolumn{1}{c}{Sampling} \\
		Dataset                                   &  & \multicolumn{1}{c}{\mylabel{number}} & \multicolumn{1}{c}{\mylabel{hours}} & \multicolumn{1}{c}{\mylabel{number}} & \multicolumn{1}{c}{\mylabel{kHz}} \\
		\midrule
		van Niekerk et al. \cite{vanniekerk2014a} &  &  33 & 2.75 & 4316 & 16 \\
		Gutkin et al. \cite{gutkin2020}           &  & 36  & 4.00 & 3583 & 48 \\
		Meyer et al. \cite{meyer2022}             &  &  & 33.30 & 10228 & 48 \\
		Ogayo et al. \cite{ogayo2022}             &  &  & 18.04 & 10978 & \\
		YFACC (ours)                              &  & 1 & 13.3 & 6000 & 48 \\
		\bottomrule
	\end{tabular}
	\label{tbl:yoruba_corpus_overview_quantitative}
\end{table*}

\paragraph{\yoruba speech corpora.} Up until very recently, there were only two open-source \yoruba speech datasets:
	the Lagos-NWU \yoruba Speech Corpus \cite{vanniekerk2014a}, consisting of approximately 2.75 hours recorded by 17 male and 16 female speakers at 16 kHz,
	and the corpus of Gutkin \etal~\cite{gutkin2020}, consisting of roughly 
	4 hours of 48 kHz recordings from 36 male and female volunteers. 
	Concurrent work to ours \cite{meyer2022,ogayo2022}, acknowledging the importance of creating datasets for African languages, has released more sizeable datasets geared towards building text-to-speech systems;
	Table~\ref{tbl:yoruba_corpus_overview_quantitative} gives their statistics.

\section{Evaluation of cross-lingual keyword localisation}
\label{sec:eval_cross_lingual_localisation}
We have already discussed in detail the evaluation procedure used to assess the performance of our monolingual models on keyword detection (see Section~\ref{subsec:keyword_detection_evaluation}), oracle keyword localisation (Section~\ref{ssec:oracle_keyword_localisation_measure}), and actual keyword localisation (Section~\ref{ssec:actual_keyword_localisation_measure}). In this section, we will only briefly explain how we adapt the evaluation procedures for the cross-lingual models considered in this chapter.

The main task we consider in this chapter is cross-lingual (English) keyword localisation in (\yoruba) speech. 
We choose precision as our main metric, which is computed as the ratio of true positive samples to retrieved samples. A sample is \textit{retrieved} if the detection score for the given English keyword is greater than a threshold $\theta$.
A sample is a \textit{true positive} if:
(1)~it is retrieved and 
(2)~the given English keyword is correctly localised 
in the \yoruba utterance,
i.e., the predicted location falls within the interval of the corresponding spoken \yoruba keyword.
We referred to this metric as (cross-lingual) actual keyword localisation.
For all experiments, we set the value of $\theta$ to $0.5$, which gave best performance on the FACC development set. 

We also quantify the upper bound on this task by looking at two additional (simpler) tasks: (cross-lingual) keyword detection and (cross-lingual) oracle keyword localisation. Both tasks and their evaluation procedures are exactly as described in Sections~\ref{subsec:keyword_detection_evaluation} and~\ref{ssec:oracle_keyword_localisation_measure}, respectively, except that here we consider the cross-lingual setting. An additional information we provided in Section~\ref{ssec:oracle_keyword_localisation_measure} is that in the oracle keyword localisation setting, we are interested in assessing the localisation ability of the model without considering detection.

\section{Cross-lingual keyword localisation with a VGS model}
\label{sec:cross_lingual_tasks}
\begin{figure}[t]
	\centering
	\includegraphics[width=0.99\columnwidth]{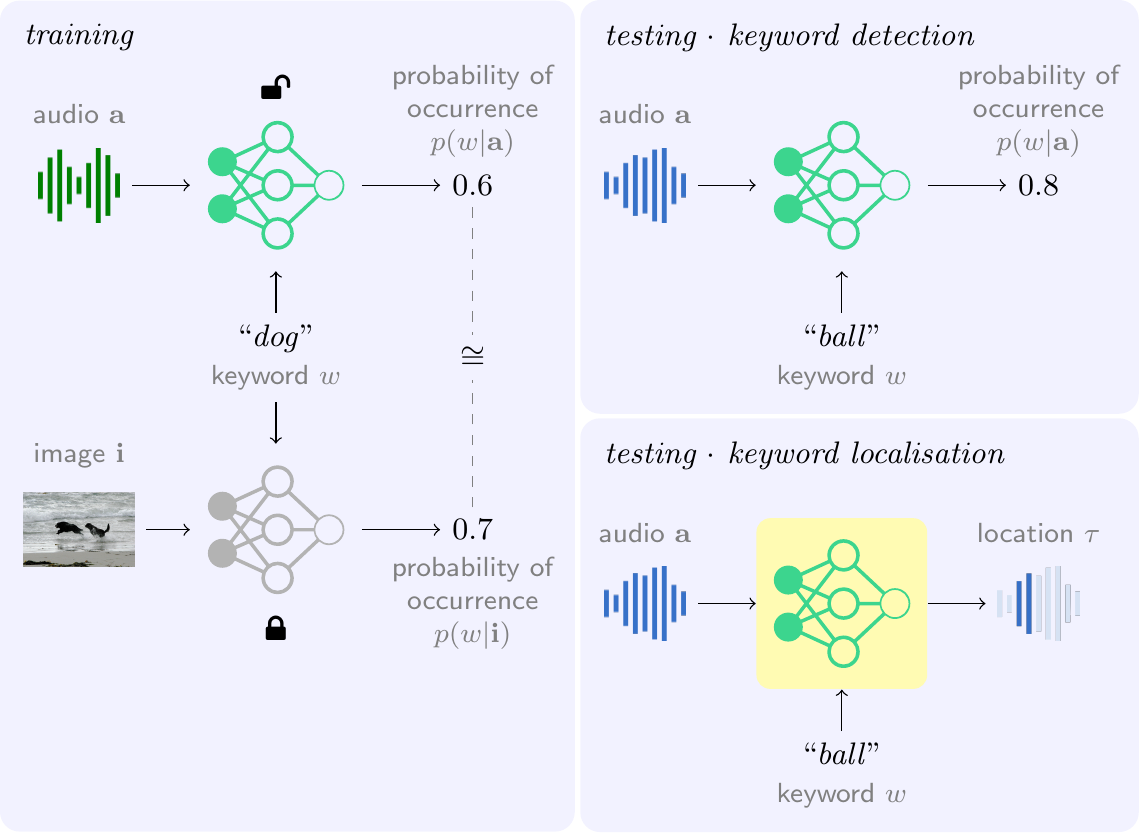}
	\caption{%
		Cross-lingual keyword localisation with a visually-grounded speech model.
		\textit{Left:} At train time, the speech model is trained to predict whether an English keyword occurs in a \yoruba audio utterance based on the visual supervision of a pretrained and frozen (English) image model.
		\textit{Right:} At test time, given an audio utterance in \yoruba, the model outputs the probability of an English keyword occurring in the utterance.
		If this probability is over a predefined threshold $\theta$, then the model additionally locates the keyword in the utterance based on the attention weights in the network.
	}
	\label{fig:overview_real}
\end{figure}
The methodological approach used in this chapter is the same as the one already described in Chapters~\ref{chap:problem_formulation_setup} and~\ref{chap:methods-for-keyword-localisation}---here we only briefly describe how we adapt it for cross-lingual task. 

We specifically consider
the task of cross-lingual keyword localisation:
given an English keyword and a spoken utterance in \yoruba,
we want to automatically find the location in the audio of the \yoruba word that corresponds to the given English keyword.
If the keyword does not occur, the model should indicate that the keyword was not detected.
Crucially, we assume that the only source of supervision comes from a visual channel.
Concretely
in our case we assume we have access to a set of images associated with corresponding audio captions (YFACC).
In order to be able use this weak source of supervision, we follow 
an almost identical approach to the monolingual VGS-based localisation method described in Section~\ref{sec:localisation_method}, which we adapt to the cross-lingual setting.

\subsection{Training: Visual supervision}

Our model consists of an audio network that processes \yoruba
speech and outputs the probability that a given English keyword occurs (anywhere) in the utterance.
This audio model is trained by cross-modal supervision:
it learns to reproduce the predictions of a pretrained image network on the associated image for a predefined set of visual categories.
The weights of the audio network are optimised through the binary cross-entropy loss between the image and audio predictions;
the loss is averaged over the keywords in the vocabulary.
The categories are defined on English keywords (such as \textit{dog}, \textit{child}, \textit{ball}), but they also correspond to \yoruba words which occur in the audio,
with the visual channel linking the two languages.
See Figure~\ref{fig:overview_real} (left) for an 
illustration of the training process.

\subsection{Testing: Cross-lingual localisation}

At test time, given a \yoruba utterance---no images are needed for inference---our model outputs the probability of occurrence of a given English keyword from the visual vocabulary.
When this probability is over a certain threshold $\theta$, we assume that the keyword occurs in the utterance.
Once a keyword is detected, we also want to
locate the corresponding \yoruba word.
To obtain the time location of the \yoruba word, we leverage the extra structure of our model, namely its attention weights over the input utterance.
We pick as the most probable location of the input keyword the time stamp with the highest
weight.
See Figure~\ref{fig:overview_real} (right) for an illustration of the testing phase.

\section{Architecture and implemetation details}
\label{sec:architecture_implementation_details_cross_lingual}

\subsection{Datasets}
In our experiments, apart
from the new
YFACC dataset (Section~\ref{sec:yfacc}) we also use the previous English FACC dataset (Section~\ref{subsec:FACC})~\cite{harwath2015} for the monolingual English VGS systems which we compare to.

We experiment with two variants of FACC that differ in the amount of training data:
\config{en-30k}, which contains all the original 30k training samples, and
\config{en-5k}, which contains a subset of 5k training samples to match the size of
YFACC.
The test data used for models trained on FACC are the 500 English utterances corresponding to the YFACC test set.
\subsection{Architecture: Attention network}
We use the architecture CNN-Attend as described in Sections~\ref{subsec:exp_setup} and~\ref{ssec:implementation_details}, but carry out a hyperparameter search over the learning rate, activations functions and size of last layer.
This search is done based on
$F_1$ keyword detection score on the development set of FACC.
Compared to the original architecture in Section~\ref{subsec:exp_setup}, the only change is increasing the last layer's size from 4096 to 8192; this improved development set $F_1$
from 14.4\% to 19.4\%.

The audio network consists of four subcomponents: an audio encoder, a keyword encoder, an attention layer and a classification network.
The audio encoder---a convolutional network in our case---embeds the \yoruba audio to a sequence of feature vectors,
while the keyword encoder assigns an embedding vector to the English keyword using a learnable lookup table.
The audio features are pooled temporally into a single vector based on the keyword embedding using the attention layer.
Finally, the aggregated feature vector is then projected to a classification score using a small multi-layer perceptron network.
As mentioned in the previous subsection, we use the location of the maximum attention scores as the location prediction of the query keyword.
This architecture is inspired by~\cite{tamer2020},
an attention-based graph-convolutional network for visual keyword spotting in sign language videos.

\subsection{Implementation details}
We follow an implementation similar to the one used for the experiments described in Sections~\ref{subsec:exp_setup} and~\ref{ssec:implementation_details}. Specifically, we downsample YFACC to 16 kHz to match FACC. All utterances are
encoded as mel-frequency cepstral coefficients,
which are further augmented at train time using SpecAugment~\cite{park2019}.
The model's visual targets are computed with a VGG-16 network~\cite{simonyan2014} trained on MSCOCO~\cite{lin2014} for 67 keywords as described in \cite{kamper2019b}.
All models are implemented as already described in Section~\ref{subsec:exp_setup}.

\section{Results: Cross-lingual keyword localisation}
\label{sec:exp_results}

\begin{table*}[!t]
	\centering
	\caption{
		Results of visually-grounded speech models
		on a cross-lingual task (querying English in \yoruba audio) on the proposed YFACC dataset (first three rows) and
		on a monolingual task (querying English in English audio) on the FACC dataset (last two rows).
		The models are evaluated for the main task of actual keyword localisation.
		The experiments are repeated three times by changing the random seed and we report the mean performance and its standard deviation.
	}
	\begin{tabular}{@{}lcc@{\hspace{1.0\tabcolsep}}c@{\hspace{1.0\tabcolsep}}ccrccr}
		\toprule
		&  &                    &  &                    &  &                                          &       &  & \multicolumn{1}{c}{Actual}\\
		&  & Keyword            &  & Audio              &  & \multicolumn{1}{c}{Training}             & Initialisation &  & \multicolumn{1}{c}{localisation} \\
		&  & \mylabel{language} &  & \mylabel{language} &  & \multicolumn{1}{c}{\mylabel{num. utts.}} &       &  & \multicolumn{1}{c}{\mylabel{precision}} \\
		\midrule
		\config{random}     &  & \multicolumn{6}{l}{Random baseline}                                              &  &           $0.1 \pm 0.0$ \\
		\config{yo-5k}      &  & \multicolumn{1}{l}{English} & $\rightarrow$ & \yoruba &  & 5k  & random          &  &          $16.0 \pm 1.6$ \\
		\config{yo-5k-init} &  & \multicolumn{1}{l}{English} & $\rightarrow$ & \yoruba &  & 5k  & \config{en-30k} &  & $\mathbf{22.8} \pm 4.3$ \\
		\midrule
		\config{en-5k}      &  & \multicolumn{1}{l}{English} & $\rightarrow$ & English &  & 5k  & random          &  &          $21.3 \pm 4.4$ \\
		\config{en-30k}     &  & \multicolumn{1}{l}{English} & $\rightarrow$ & English &  & 30k & random          &  & $\mathbf{26.9} \pm 1.8$ \\
		\bottomrule
	\end{tabular}
	\label{tbl:main_results_actual_real}
\end{table*}

\begin{table*}
	\centering
	\caption{
		Results of visually-grounded speech models
		on a cross-lingual task (querying English in \yoruba audio) on the proposed YFACC dataset (first three rows) and
		on a monolingual task (querying English in English audio) on the FACC dataset (last two rows).
		The models are evaluated for an easier task (oracle localisation), which provides an upper bound on the actual localisation performance.
		The experiments are repeated three times by changing the random seed and we report the mean performance and its standard deviation.
	}
	\begin{tabular}{@{}lcc@{\hspace{1.0\tabcolsep}}c@{\hspace{1.0\tabcolsep}}ccrccr}
		\toprule
		&  &                    &  &                    &  &                                          &       &  & \multicolumn{1}{c}{Oracle} \\
		&  & Keyword            &  & Audio              &  & \multicolumn{1}{c}{Training}             & Initialisation &  & \multicolumn{1}{c}{localisation} \\
		&  & \mylabel{language} &  & \mylabel{language} &  & \multicolumn{1}{c}{\mylabel{num. utts.}} &       &  &  \multicolumn{1}{c}{\mylabel{accuracy}} \\
		\midrule
		\config{random}     &  & \multicolumn{6}{l}{Random baseline}                                              &  &    $5.3 \pm 0.9$ \\
		\config{yo-5k}      &  & \multicolumn{1}{l}{English} & $\rightarrow$ & \yoruba &  & 5k  & random          &  &    $23.1 \pm 3.0$  \\
		\config{yo-5k-init} &  & \multicolumn{1}{l}{English} & $\rightarrow$ & \yoruba &  & 5k  & \config{en-30k} &  &  $\mathbf{35.0} \pm 2.5$ \\
		\midrule
		\config{en-5k}      &  & \multicolumn{1}{l}{English} & $\rightarrow$ & English &  & 5k  & random          &  &  $27.3 \pm 2.8$  \\
		\config{en-30k}     &  & \multicolumn{1}{l}{English} & $\rightarrow$ & English &  & 30k & random          &  & $\mathbf{44.0} \pm 0.9$ \\
		\bottomrule
	\end{tabular}
	\label{tbl:main_results_oracle_real}
\end{table*}
\begin{table*}
	\centering
	\caption{
		Results of visually-grounded speech models
		on a cross-lingual task (querying English in \yoruba audio) on the proposed YFACC dataset (first three rows) and
		on a monolingual task (querying English in English audio) on the FACC dataset (last two rows).
		The models are evaluated for an easier task (keyword detection), which provides an upper bound on the actual localisation performance.
		The experiments are repeated three times by changing the random seed and we report the mean performance and its standard deviation.
	}
	\begin{tabular}{@{}lcc@{\hspace{1.0\tabcolsep}}c@{\hspace{1.0\tabcolsep}}ccrccr}
		\toprule
		&  &                    &  &                    &  &                                          &       &  &  \multicolumn{1}{c}{Keyword}   \\
		&  & Keyword            &  & Audio              &  & \multicolumn{1}{c}{Training}             & Initialisation &  &  \multicolumn{1}{c}{detection} \\
		&  & \mylabel{language} &  & \mylabel{language} &  & \multicolumn{1}{c}{\mylabel{num. utts.}} &       &  &  \multicolumn{1}{c}{\mylabel{precision}} \\
		\midrule
		\config{random}     &  & \multicolumn{6}{l}{Random baseline}                                              &  &           $2.4 \pm 0.2$ \\
		\config{yo-5k}      &  & \multicolumn{1}{l}{English} & $\rightarrow$ & \yoruba &  & 5k  & random          &  &           $26.7 \pm 6.4$ \\
		\config{yo-5k-init} &  & \multicolumn{1}{l}{English} & $\rightarrow$ & \yoruba &  & 5k  & \config{en-30k} &  &  $\mathbf{33.3} \pm 7.1$ \\
		\midrule
		\config{en-5k}      &  & \multicolumn{1}{l}{English} & $\rightarrow$ & English &  & 5k  & random          &  &          $29.5 \pm 5.0$ \\
		\config{en-30k}     &  & \multicolumn{1}{l}{English} & $\rightarrow$ & English &  & 30k & random          &  &  $\mathbf{39.2} \pm 1.7$ \\
		\bottomrule
	\end{tabular}
	\label{tbl:main_results_detection_real}
\end{table*}

\subsection{Quantitative results}
Our goal is to
investigate to what extent cross-lingual keyword localisation is possible with a VGS 
model trained in an extreme low-resource setting.
We specifically consider the case where we have 
only 13.3 hours of untranscribed \yoruba speech with corresponding images.
English--\yoruba cross-lingual results are given at the top of Tables~\ref{tbl:main_results_actual_real}, ~\ref{tbl:main_results_oracle_real}, and~\ref{tbl:main_results_detection_real}.

Table~\ref{tbl:main_results_actual_real} shows the performance of the models on our main task, actual keyword localisation task. The cross-lingual model \config{yo-5k} obtains a performance of about $16.0$\%.
Although this 
appears modest when viewed in isolation, it is much better than the performance of a random model ($0.1$\%).
More
importantly, the result 
falls within the range of results ($21.3\pm4.4$\%)
that are obtained in a monolingual setting (querying English keywords in English speech) with a model trained on equal number of English utterances (\config{en-5k}).
It can be argued that the cross-lingual setting is more challenging because the visual keywords are more English-centric and do not account for the cultural and linguistic differences to \yoruba (Section~\ref{ssec:yfacc-cross-differences}).
On the other hand, YFACC is easier in the sense it is a single-speaker dataset, whereas the FACC dataset involves multiple speakers.
Next we consider aspects that can help improve performance.

\subsubsection*{Amount of training data}
While for the cross-lingual setting we are limited to the 5k training utterances available in 
YFACC, 
we can extrapolate the performance on more data based on corresponding monolingual setting using
FACC.
In Table~\ref{tbl:main_results_actual_real} we see that training with six times more data,
unsurprisingly, yields considerable improvements, from $21.3$\% (\config{en-5k}) to $26.9$\% (\config{en-30k}).
However, while increasing the training size is the most straightforward way to improve performance, it might also be the most challenging in practice.

\subsubsection*{Speech representations}
Training a model from scratch in low-resource conditions is challenging.
A common practice is to start from a pretrained model and to then fine-tune it 
on the downstream task.
Here we do this by initialising
the training on \yoruba from a VGS model trained on all the image--English audio pairs in FACC.
Such English pairs
are arguably more readily available than
\yoruba utterances.  We observe a strong boost in performance---from $16.0$\% (\config{yo-5k}) to $22.8$\% (\config{yo-5k-init}) shown in Table~\ref{tbl:main_results_actual_real}.

\subsubsection*{Decoupling the tasks}
Keyword localisation involves two steps:
(1) detecting whether the query keyword appears (somewhere) in the utterance, and, if so,
(2) localising the keyword in the given utterance.
In order to understand which component of 
our approach
requires
improvements, it is instructive to independently evaluate each of these steps.
To do that, we assume that one of the steps is perfect, resulting in
two new tasks:
oracle localisation (assumes that detection is perfect) and keyword detection (assumes that localisation is perfect).
Whilst not directly comparable, the results for the two tasks (Table~\ref{tbl:main_results_oracle_real} for oracle keyword localisation and Table~\ref{tbl:main_results_detection_real}) for keyword detection) indicate that both components can be further improved.
Also note
that the previous conclusions in terms of relative performance between the models still hold 
on these
new tasks.

\subsubsection*{Individual keyword performance}
The performance varies widely across the 67 keywords in the vocabulary.
We observe that there are keywords with good performance (better than the average), such as 
\textit{brown} (\textit{b\'ur\'a\`un}; $100$\%),
\textit{bike} (\textit{k\d{è}k\d{\'e}}; $94.1$\%) or
\textit{grass} (\textit{kor\'iko} $90.9$\%).
But there are many others on which the model struggles.
The reason for the latter include poor visual grounding (e.g., \textit{camera}, \textit{wearing}, \textit{large}) and confusion between semantically related concepts (\textit{riding} often retrieves \textit{k\d{\`e}k\d{\'e}}, i.e. \textit{bicycle}; \textit{swimming} retrieves \textit{od\`o ad\'ag\'un}, i.e. \textit{pool}).
Paying close attention to the particularities of the individual keywords can serve as inspiration for
future improvements. 
\subsubsection*{Cross-lingual visual teacher performance}
\begin{figure}[t]
	\centering
	\includegraphics[width=0.99\columnwidth]{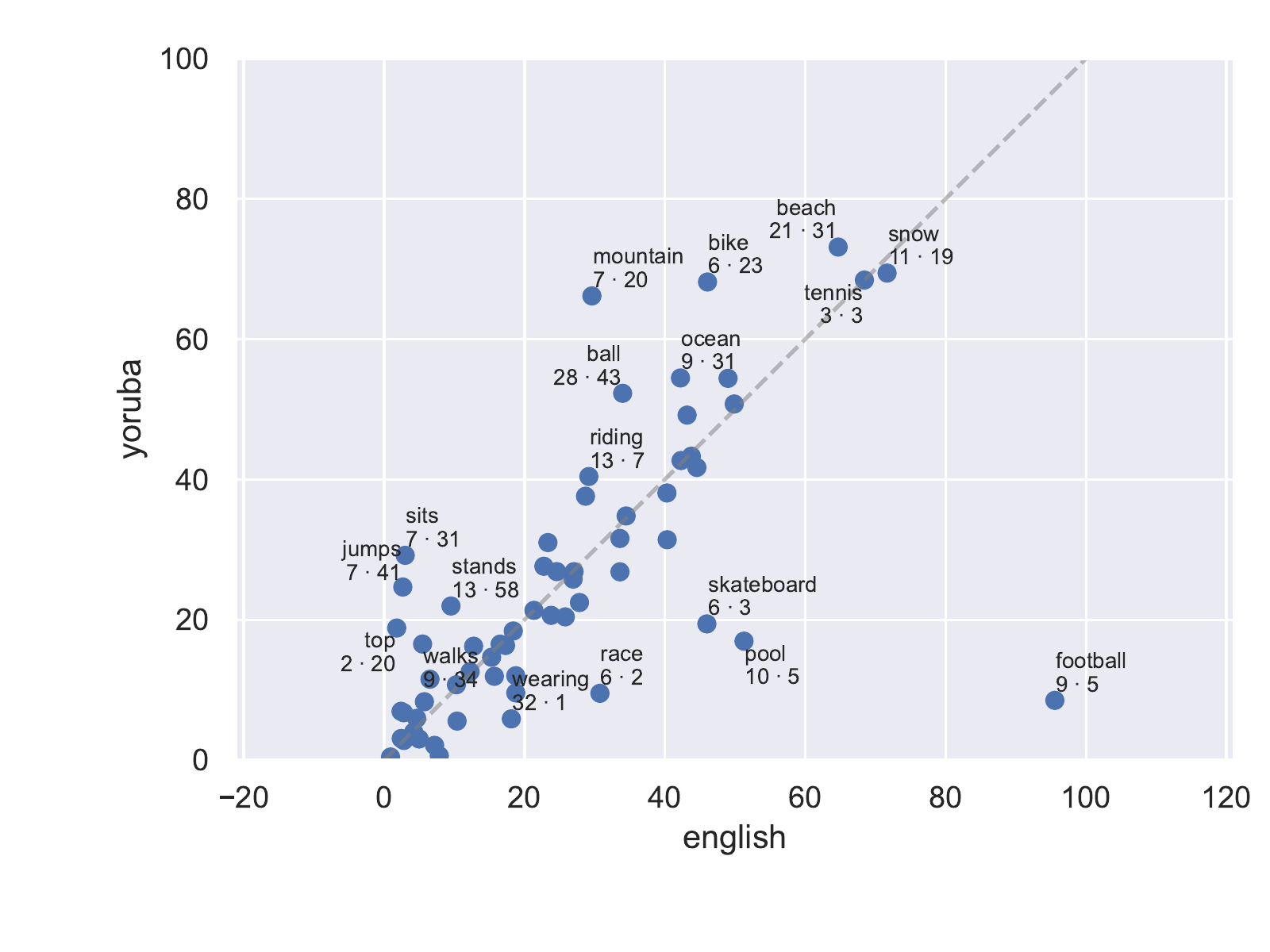}
	\caption{%
		The mean average precision of the visual teacher on the English and Yorùbá Flickr test sets.
	}
	\label{fig:cross-lingual-visual-teacher}
\end{figure}
\begin{figure}[!t]
	\centering
	\includegraphics[width=0.99\linewidth]{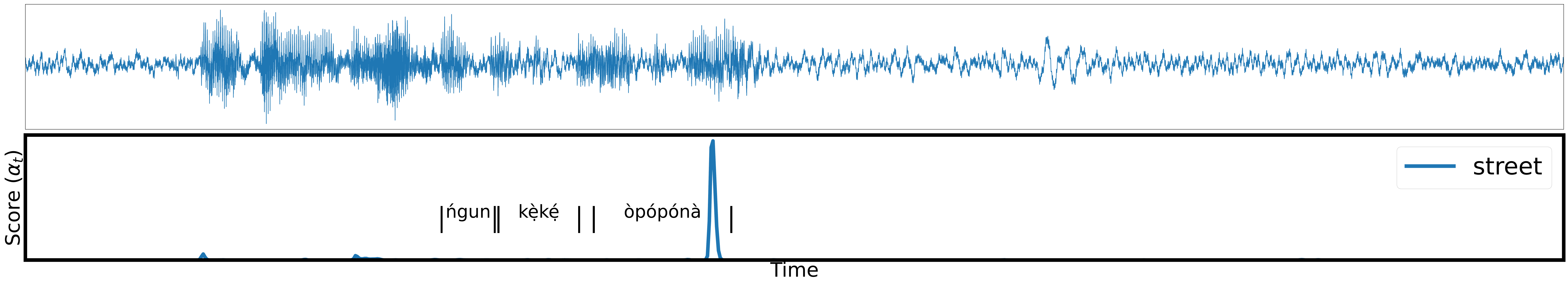} \\[-3pt]
	{\footnotesize (a)} \\[5pt]
	\includegraphics[width=0.99\linewidth]{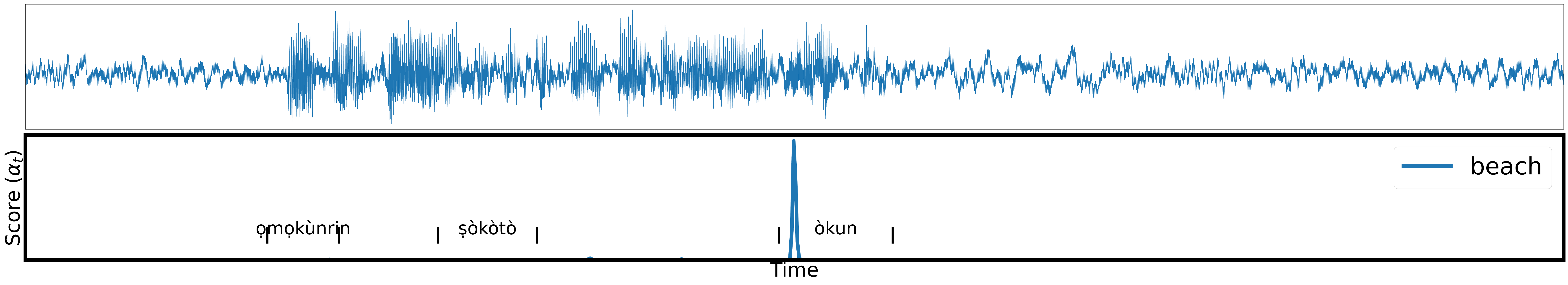} \\[-3pt]
	{\footnotesize (b)}
	\includegraphics[width=0.99\linewidth]{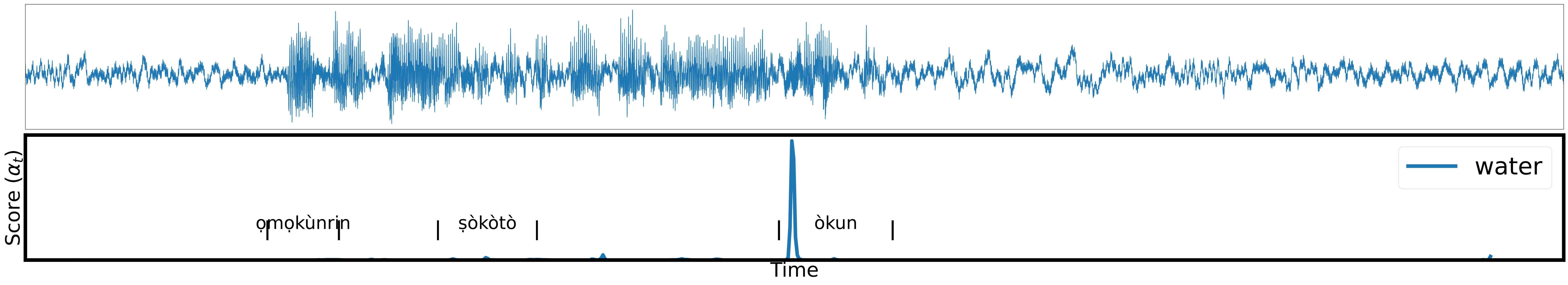} \\[-3pt]
	{\footnotesize (c)}
	\includegraphics[width=0.99\linewidth]{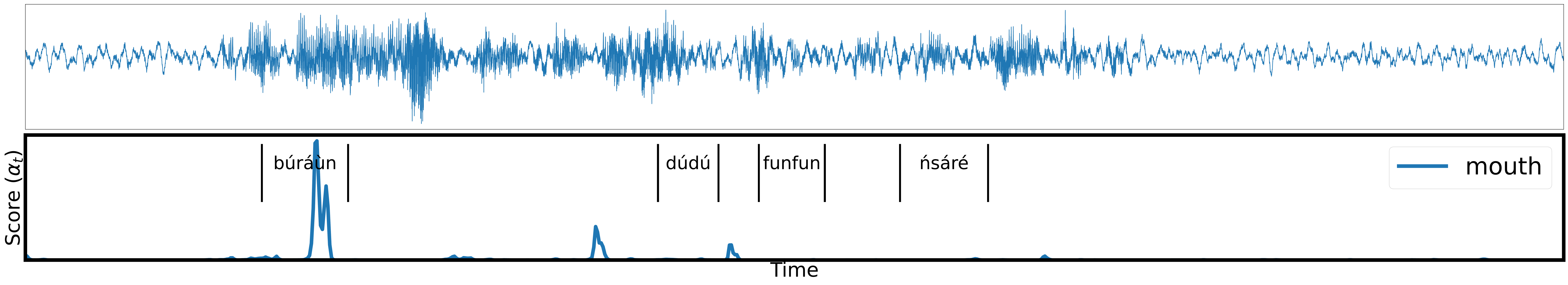} \\[-3pt]
	{\footnotesize (d)} 
	\vspace*{-5pt}
	\caption{Examples of cross-lingual localisation with the \config{yo-5k} VGS model. The English query keyword is shown on the right.}
	\label{fig:qualitative_eval}
\end{figure}
We compute the performance of the visual teacher on the English and Yorùbá Flickr test sets. The images are the same in both cases (so the visual teacher’s predictions are the same), but because of the languages characteristics, the captions may use different keywords, so naturally we will have differences in performance. We report the mean average precision and show the results in Figure~\ref{fig:cross-lingual-visual-teacher}. (The figure also shows the number of occurrences for each keyword in the two test sets, the original English test set and the Yorùbá test set, respectively.)

Some interesting observations to be made:
\begin{itemize}
	\item We notice that “football” has very good performance in English, but poor performance in Yorùbá. This behaviour might be explained by the fact that the English “football” and “soccer” keywords are translated using a single Yorùbá keyword, but they are visually distinct; the visual teacher is trained with  semantics of the English keyword.
	\item However, while “beach” and “ocean” also have a single Yorùbá correspondent, the two performances are much closer together; this happens because the two words are correlated visually—they usually co-occur in images.
	\item Words like “jumps”, “sits”, “stands” obtain a better performance in Yorùbá than in English, because they are more frequent in the Yorùbá captions (maybe because some other English words are mapped to them?), so their base (random) performance is higher.
\end{itemize}

\subsection{Qualitative results}
To better understand the failure modes of the cross-lingual model, we 
qualitatively assess the results.
Figure~\ref{fig:qualitative_eval}(a) and (b) show correct localisations of English keywords in \yoruba utterances:
the model fires within the frames containing the spoken word \textit{\`op\'op\'on\`a} when prompted with the English keyword \textit{street} (\ref{fig:qualitative_eval}a) and
fires within the frames of \textit{\`okun} when prompted with the keyword \textit{ocean}.
Figures~\ref{fig:qualitative_eval}(c) and (d) show
failure cases. 
Figure~\ref{fig:qualitative_eval}(c) shows that the model also sometimes localises a semantically related word in the utterance: the spoken \yoruba word \textit{\`okun} (\textit{ocean}) is assigned the highest localisation score when prompted with the English keyword \textit{water}.
Figure~\ref{fig:qualitative_eval}(d) is 
an outright failure.

\section{Summary}
 This chapter introduced a new single-speaker dataset of \yoruba spoken captions for Flickr8k images. 
 To showcase what is possible with such a small dataset (13.3 hours, 6k speech--image pairs) in a real low-resource setting, we used the data for cross-lingual keyword localisation using a VGS model, trained without any transcribed audio.
 
Apart from the inherent challenges in the cross-lingual retrieval task,
 the English--\yoruba pair is also a challenging combination, as the two languages belong to different families (Indo-European and Nigerian-Congo) and they are associated with very different cultures.
 Through systematic experimentation, we quantified the impact of this language mismatch: we analysed
 how well the selected keywords match across languages,
 we compared to a monolingual English VGS model trained on similar amounts of data to YFACC,
 and we transferred English speech representations to \yoruba (resulting in improved retrieval performance).
 
 A major advantage of extending Flickr8k is that it allows reusing existing annotations (English recordings \cite{harwath2015}, translations in other languages \cite{elliott2016,li2016}), which could help with constructing richer and potentially more interesting tasks in future work.
 	
 To summarise, our contributions in this chapter include:
 \begin{itemize}
 	\item The release of a new speech--image dataset in \yoruba.
 	\item A baseline system for cross-lingual keyword detection and localisation.
 	\item A multi-faceted cross-lingual analysis of systems.
 \end{itemize}
 We hope that YFACC
 will stimulate research in the use of VGS models for solving speech tasks in real low-resource languages.

%% file: conclusion.tex
\graphicspath{{conclusion/fig/}}

\chapter{Summary and conclusion}
\label{chap:conclusion}
 Visually grounded speech modelling (VGS) can be used for enabling speech applications in low-resource settings where transcribed speech data is not available. This dissertation investigated keyword localisation in speech---finding where in an utterance a given written keyword occurs---using VGS models trained in a real low-resource linguistic environment. Below, we first provide direct answers to the research questions tackled throughout this dissertation (Section~\ref{sec:conclusion_questions_answers}). We then summarise and reflect on the directions and approaches taken to arrive at our answers to the questions, and present our concrete contributions to VGS modelling (Section~\ref{sec:conclusion_main_findings}). We propose recommendations for future work on the topic of the dissertation (Section~\ref{sec:conclusion_future_work}). Finally, we give a concise conclusion of the dissertation, focussing on key findings and take-aways (Section~\ref{sec:conclusion_conclusion}).
 \section{Research questions and proposed answers}
 \label{sec:conclusion_questions_answers}
 We start by briefly stating the main findings for our two main research questions.
 \paragraph{Research question 1.} Is keyword localisation possible with VGS models? Previous work had shown that VGS models can be used for tasks such as cross-modal retrieval, keyword detection and keyword spotting, but keyword localisation was not explored. One application of visually grounded keyword localisation is in the documentation of endangered languages~\cite{ferrand2020}. A linguist can employ such tools to rapidly locate the speech segments containing a query keyword in a collection of utterances; the only prerequisite is a collection of image-speech pairs from the target language to train the model. Based on a quantitative and qualitative analyses of keyword localisation performance of VGS model in an artificial low-resource setting, we conclude that limited keyword localisation is possible using VGS model trained in a setting where English was used to simulate a low-resource linguistic environment. Furthermore, as in many other VGS studies, we found that many of the incorrect keyword localisations were due to semantic confusions, for example, locating the word ``backstroke" in an utterance for the query keyword ``swimming".
\paragraph{Research question 2.} In a real low-resource setting, can we do visually grounded keyword localisation cross-lingually?
Most previous VGS studies used datasets where images were paired with speech in English (or another well-resourced language). English is therefore often used as a proxy for a low-resource language, making it difficult to accurately assess VGS system performance in a real low-resource setting. We addressed this limitation by collecting a new speech--image dataset in \yoruba, and then trained a VGS model in a real low-resource setting to perform cross-lingual keyword detection and localisation. Based on quantitative and qualitative analyses of the VGS systems on the cross-lingual keyword localisation task, it can be concluded that the VGS system achieved a modest performance on a limited number of keywords in the system's vocabulary. Furthermore, it can be concluded that the representations learned by a cross-lingual VGS model in this low-resource setting can be enriched through initialising from a model pretrained on a well-resourced language. Qualitative analysis of the cross-lingual models also showed a behaviour similar to the first research question, where the models made semantic mistakes. One question is whether models should be penalised for making these types of mistakes when measuring their performance. Localising a semantically related keyword could be a useful application in itself. For example, in delivering humanitarian service to flood victims, localising the keyword ``\`e\'eb\`i" (the \yoruba word for ``vomit") in a \yoruba utterance when prompted with the English word ``cholera" could still give the relief worker an idea of what treatment to administer.

\section{Findings}
 In this section we summarise and reflect on the approaches taken to tackle the above research questions, focussing on briefly recapping the VGS methodology that we used as a starting point, summarising the four localisation methods that we proposed as an extension of this VGS methodology, and finally reflecting on the findings. We considered two different settings throughout the dissertation: the artificial low-resource setting which is the setting used to tackle the first research question, and the real low-resource setting used to tackle the second research question.
\label{sec:conclusion_main_findings}
\subsection{The artificial low-resource regime}
Tackling the first research question within the artificial low-resource context---where English is used to simulate a low-resource linguistic environment---brought about a number of gains. Firstly, it allowed us to compare and directly extend previous work. Secondly, it allowed us to first investigate whether a VGS system trained on an existing VGS dataset---specifically the existing Flickr Audio Caption Corpus (FACC)---can perform keyword localisation. Through this investigation we could also estimate how much data resources would be required to carry out the study in a real low-resource linguistic environment---where VGS systems are actually needed. Thirdly, simulating a low-resource setting as opposed to jumping straight into a real low-resource setting might also allow for faster prototyping and exploration of methodologies that might be better suited for a specific downstream task. For example, a computational linguist attempting to build a system for detecting and localising a keyword in an endangered language would ideally first need to identify the potentially top-performing methodology that would yield the best result.

The specific methodology best suited for answering our first research question is based on the work of Kamper~\etal~\cite{kamper2017a, kamper2019b}, who showed that we could use an external image tagger with a fixed vocabulary to obtain soft text labels as targets for a speech network that maps speech to keyword labels. We used this methodology as a starting point, not only in tackling the first research question, but also for the second research question. As a contribution, we proposed four localisation methods. The proposed localisation methods used the intrinsic ability of the VGS models to localise keywords.
In the first method, we adapted the Gradient-weighted class activation mapping (Grad-CAM)~\cite{selvaraju2017}:~a saliency-based method popularly used in the vision community for determining which parts of an input image contributes most to a particular output prediction. In the second, we adapted a score aggregation method~\cite{palaz2016} to use visual supervision. In the third, we asked whether attention (with an architecture inspired by~\cite{tamer2020}) is beneficial for visually grounded keyword localisation. Finally, we proposed mask-based methods, which used masking at different locations to measure the response score predicted by the trained model on the partial inputs; significant variations in the output suggest the presence of a keyword. The mask-based method, like Grad-CAM, is adaptable to any architecture.

Although Grad-CAM is adaptable to any convolutional neural network architecture, it matched poorly with the specific multi-label loss used to train the VGS model. The results suggested a mismatch between saliency-based localisation and the multi-label loss used, with a superior detection model
performing poorly in localisation. This finding indicated that better localisation should be possible given a mechanism better aligned to the model and multi-label classification loss.
The absolute performance of score aggregation was low. Nevertheless, this method showed some signals for keyword localisation and performed better than Grad-CAM. We found that the attention method provided a significant gain over existing keyword localisation methods. However, scores are still far from models trained with the unordered bag-of-word supervision (obtained transcriptions), which could be considered an idealised VGS model. Furthermore, we showed that the visually trained models are penalised for locating semantically related words, for example, locating the word ``backstroke" for the query keyword ``swimming".
In the oracle setting, where the system assumes that a keyword is present and then directly predicts where it occurs, the mask-based approach gave an oracle performance of 57.0\% compared to the 46.0\% achieved with the attention-based method. However, the masked-based approach is computationally more expensive than the attention localisation method.

We performed detailed analyses on the different aspects of the methodologies employed within the artificial low-resource setting. We summarise our findings below.
\paragraph{Comparing visual supervision to BoW supervision.} 
We investigated the BoW setting, which we considered as a setting where we have access to a perfect visual tagger. The analysis allowed us to estimate the penalty we pay using visual supervision from an offline pretrained vision system.
In all cases, we saw a significant drop in performance when we moved from BoW supervision to visual supervision. We underscored that although BoW gave an upper bound for VGS performance, it could be a high upper bound since a perfect visual tagger might detect objects that humans do not describe.
We concluded that improving visual tagging could substantially improve keyword localisation using VGS models.

\paragraph{Per-keyword performance.}We investigated the localisation performance for each individual keyword in the system vocabulary. 
We observed that the performance of the methods is very keyword-dependent and verified that how well a keyword is localised is influenced by how well each word is visually grounded. Moreover, for cases where visually groundable keywords are still poorly localised, we found that co-occurrences of the keywords with other words caused such poor localisation. We concluded that it is challenging to discriminate words that often occur together in the same utterance with any localisation method that does not have access to additional information.

\paragraph{Effect of intermediate max pooling on localisation performance.}We also analysed the different architectures and localisation methods by looking at how their choice of encoder structure affects their localisation abilities. We found that in the architectures where a max pooling operation followed each convolutional layer operation (except the last one), localisation performance dropped. We expected such behaviour since the max pooling operation makes it more challenging to backtrace each output prediction to its relevant filter. Although a previous study~\cite{olaleye2020} showed that the max pooling operation helped with better keyword detection, we concluded that keyword localisation models should avoid it.

\paragraph{Decoupling keyword localisation and detection.}The keyword localisation task investigated in this dissertation involved two steps: (i) checking whether the query keyword appears (somewhere) in the utterance and, if so, (ii) finding where the keyword appeared. To quantify the limit these coupled steps imposed on the models, we assumed that one of the steps is perfect. This resulted in two localisation tasks: oracle localisation (assumed that detection is perfect) and keyword detection (assumed that localisation is perfect). We observed in all cases that the first detection pass impaired localisation. We concluded that decoupling these tasks could result in improved performance in future work. 
\subsection{The real low-resource regime} 
We tackled the second research question in a real low-resource setting. We started by collecting a new VGS dataset in a real low-resource language, \yoruba, and then demonstrated how cross-lingual keyword detection and localisation can be performed
with a VGS model trained on this data. The \yoruba Flickr Audio Caption Corpus (YFACC) dataset that we collected contains spoken captions for 6k Flickr images produced by a single speaker in \yoruba---a real low-resource language spoken in Nigeria. We used YFACC to train a VGS system for cross-lingual keyword detection and localisation: given an English text query, detect whether the query occurs in \yoruba speech, and if it is detected, localise where in the utterance the query occurs. 

To design this VGS system, we used as starting point the approach of Kamper and Roth~\cite{kamper2018} which simulated a low-resource setting by using two well-resourced languages. In particular, the authors used German soft text labels---generated by an offline visual tagger---to train a speech network that takes in English speech and produces German text labels. 
We instead tagged images associated with corresponding \yoruba speech with English labels using an offline (English) image tagger. This allowed us to train an attention-based VGS model to perform cross-lingual keyword detection and localisation
with an English keyword and \yoruba speech. 
The attention architecture we used is similar to the one employed in the artificial setting. However, instead of embedding English utterances to a sequence of feature vectors, we embedded \yoruba utterances.

The cross-lingual model trained on 5k \yoruba utterances obtained a precision of 16.0\% on an actual keyword localisation task (detecting whether  keyword occurs and then localising the keyword). This result is modest when viewed in isolation, but falls
within the range of results ($21.3 \pm 4.4\%$) obtained with a monolingual model trained on 5k English utterances in the artificial setting. We argued, on the one hand, that the cross-lingual setting is more challenging because the visual keywords are more English-centric and do not account for the cultural and linguistic differences in \yoruba. On the other hand, YFACC is easier in the sense it is a single-speaker dataset, whereas the Flickr Audio Caption Corpus (FACC)---the English dataset that we used---involves multiple speakers.

We performed a detailed analysis of the cross-lingual keyword tasks focussing on the impact of the amount of training data, speech representations, individual keyword performance, and task decoupling. Furthermore, we gave qualitative interpretations of the cross-lingual attention-based VGS model. We summarise our findings below.

\paragraph{Amount of training data.}
For the cross-lingual setting, we were limited to 5k training utterances available in YFACC. To know how the model would perform if trained on more data, we extrapolated the performance on more data based on the artificial setting using FACC. We saw that training improved considerably with six times more data. However, we concluded that scaling up training data in a real low-resource setting might not be viable in practice.

\paragraph{Speech representations.}
We faced two constraints in the real low-resource setting: (i)~training a model from scratch in low-resource conditions is challenging, partly due to lack of an adequate amount of data, and (ii) collecting more data in such a setting might be impossible. Hence,
we followed a common practice of mitigating these constraints by starting from a pretrained model and then fine-tuning it on the downstream task. We specifically initialised the training on \yoruba from a VGS model trained on all image--English audio pairs in the FACC dataset---a bigger dataset.
This yielded a substantial boost in performance.

\paragraph{Individual keyword performance.}
The performance of the cross-lingual model varied widely across the 67 keywords in the system's vocabulary.
We observed that there are keywords with good precision (better than the average) on the actual keyword localisation task.
However, there were many others on which the model struggled.
The reason for the latter includes poor visual grounding (for example \textit{camera}, \textit{wearing}, \textit{large}) and confusion between semantically related concepts (for example \textit{riding} often retrieved \textit{k\d{\`e}k\d{\'e}}, i.e. \textit{bicycle}; \textit{swimming} retrieved \textit{od\`o ad\'ag\'un}, i.e. \textit{pool}). 
We concluded that paying close attention to the particularities of the individual keywords can serve as inspiration for
future improvements.

\subsection{Research contributions}
This dissertation aimed to answer two research questions: (i) Is keyword localisation possible with VGS models? (ii) In a real low-resource setting, can we do visually grounded keyword localisation cross-lingually?
To address these questions we started with an artificial low-resource setting where the English language, consider to be a well-resourced language, is treated as a low resource language. We then proceeded to a real low-resource scene where we investigated VGS models in a real low-resource linguistic environment.
We summarise the contributions that resulted from tackling these research questions below. 
\paragraph{Research contribution 1.} We developed a new VGS model for keyword detection and keyword spotting using attention, and performed a thorough comparison of this approach to existing VGS-based methods. 
\paragraph{Research contribution 2.} We extended the VGS models for keyword localisation by proposing four localisation methods.
\paragraph{Research contribution 3.} We performed a detailed quantitative and qualitative analysis revealing the limits of the extended VGS models, showcasing the success and failure modes. This is carried out in a setting where English VGS data is artificially considered as a low-resource language.
\paragraph{Research contribution 4.} We released a new multimodal, multilingual dataset which enables VGS modelling in a real low-resource setting resembling language documentation setting. 
\paragraph{Research contribution 5.} We built a baseline system for cross-lingual keyword detection and keyword localisation
in a real low-resource setting. The baseline system was trained  on 5k \yoruba utterances paired with English soft text labels, producing a speech network that detects and localise an English keyword in \yoruba speech.
\paragraph{Research contribution 6.} We performed multi-faceted analyses of the cross-lingual VGS models above. The analyses helped to better understand how performance on the keyword localisation task varies across the 67 keywords in the system's vocabulary, and helped to interpret the failure modes of the cross-lingual model.
\section{Future work}
\label{sec:conclusion_future_work}
\paragraph{On the choice of modalities.}
The datasets considered in this dissertation are distinct from video datasets because they do not include the video modality---only still images. Extending the approaches developed in this dissertation to deal with video could be an interesting direction for future work.
\paragraph{On the choice of the visual tagger.} We considered a well-established off-the-shelf visual tagger to extract soft labels from the image modality. There might be more sophisticated visual taggers better suited for the keyword spotting and localisation tasks that we consider in this dissertation.
\paragraph{Moving to an open vocabulary.} Our VGS models are trained to detect and localise a predefined set of keywords. Hence it is not clear how the models will behave in an open vocabulary context where the models are not explicitly trained to recognise predefined keywords. There are some very recent work in this direction, for instance~\cite{nortje2023} and~\cite{shih2023}.
\paragraph{Going beyond point-based localisation.} We considered a point-based localisation task, where a keyword is considered correctly localised if its predicted location falls within the ground truth alignments. This is sufficient for the application context described earlier, but it would also be interesting to investigate predicting the entire duration of a keyword in an utterance, i.e. predicting the start and end points of a given keyword. There are also some very recent work in this direction~\cite{peng2022}.
\paragraph{Assessing VGS systems on semantic keyword localisation.} We qualitatively showed that many incorrect localisations are caused by localising semantically related words.
Future work can examine cases where the models are evaluated not just on their ability to propose the location of the exact keyword of interest but also on semantically related keywords. To assess this in a real low-resource setting, one option would be to extend our YFACC dataset with semantic annotations.
\section{Conclusion}
\label{sec:conclusion_conclusion}
This dissertation made contributions in three of the main areas of focus in VGS modelling: (i)~methodology, (ii) datasets and (iii) analysis. 

In terms of methodology (i), we extended existing methods for VGS-based keyword detection, by giving them the ability to also locate a given written keyword. We specifically used a VGS method that tags training images with soft textual labels as targets for a speech network that can then detect the presence of a keyword in an utterance. We equipped this type of VGS model with localisation capabilities, investigating four approaches. Our best approach relied on input masking: an approach that masks the input signal at
different locations and measures the difference in the output unit for a particular keyword.
In an artificial low-resource setting where we treated English VGS data as low-resource, this approach gave a 57.0\% localisation accuracy on an oracle keyword localisation task in which we assumed the system knew that a keyword occurs in an utterance and only needed to predict its location. The attention localisation approach---a method that pools features over the temporal axis to produce attention weights as localisation scores---came second, with an oracle localisation accuracy of 46.0\%. In a real low-resource setting, we used this attention localisation method, but instead of embedding English utterances to a sequence of feature vectors, we embedded \yoruba utterances. The attention-based localisation method achieved an actual keyword localisation (detecting whether a keyword occurs and then localising the keyword) precision score of 16.0\% with a model trained from scratch and 22.8\% with a model initialised from a model trained on a larger dataset.
In terms of analysis of VGS models (iii), we systematically compared four different keyword localisation methods; provided detailed experiments revealing the impact of our choice of architecture and performance at the individual keyword level. Furthermore, we provided qualitative analyses showcasing the success and failure modes of the VGS models on keyword detection and localisation.

In terms of datasets (ii), we introduced a new single-speaker dataset of \yoruba spoken captions for Flickr images, called YFACC. 
To showcase what is possible with such a small dataset (13.3 hours, 6k speech--image pairs) in a real low-resource setting, we used the data for cross-lingual keyword localisation using a VGS model trained without any transcribed audio.

We believe that our new dataset together with our findings presented here will stimulate more research in the use of VGS models for real low-resource languages.